\documentclass[conference]{IEEEtran}
\usepackage{times}

\usepackage[numbers]{natbib}
\usepackage{multicol}
\usepackage[bookmarks=true,hidelinks]{hyperref}
\usepackage{subcaption}

\usepackage{xspace}
\usepackage{amsmath,amssymb}
\usepackage{makecell}
\usepackage{url}
\usepackage{float}
\usepackage{placeins}
\usepackage{multirow}
\usepackage{graphicx}
\usepackage{booktabs}
\usepackage{tabularx}
\usepackage{wrapfig}
\usepackage[table]{xcolor}
\usepackage{array}
\usepackage[percent]{overpic}
\usepackage{tikz}
\newcommand{\dashedleg}{\protect\tikz[baseline=-0.5ex]{\protect\draw[blue, dashed, line width=1.2pt] (0,0) -- (1.5em,0);}}
\newcommand{\solidleg}{\protect\tikz[baseline=-0.5ex]{\protect\draw[red, line width=1.2pt] (0,0) -- (1.5em,0);}}
\definecolor{colB}{RGB}{240,200,200}
\definecolor{colA}{RGB}{200,235,210}
\newcolumntype{B}{>{\columncolor{colB}}c}
\newcolumntype{A}{>{\columncolor{colA}}c}

\usetikzlibrary{shapes.geometric, arrows.meta, positioning, fit, backgrounds, calc}

\begin{document}

\newcommand{\name}{NeuralActuator\xspace}
\newcommand{\dbname}{NAD\xspace}

\IEEEoverridecommandlockouts
\title{\vspace*{-3mm}\name: Neural Actuation Modeling for Robot Dynamics and External Force Perception\vspace*{-3mm}}

\newcommand{\affmark}[1]{\textsuperscript{\normalfont #1}}

\vspace{-2mm}
\author{\authorblockN{Zhiyang Dou\affmark{1},
John U. Onyemelukwe\affmark{1}\textcolor{black}{\affmark{*}},
Hangxing Zhang\affmark{1}\textcolor{black}{\affmark{*}},
Heng Zhang\affmark{1},
Minghao Guo\affmark{1},
Yunsheng Tian\affmark{1},\\
Michal Piotr Lipiec\affmark{1},
Joshua Jacob\affmark{1},
Chao Liu\affmark{1},
Peter Yichen Chen\affmark{1},
Yuri Ivanov\affmark{2,$\dagger$} and
Wojciech Matusik\affmark{1}}
\vspace{3mm}
\authorblockA{\affmark{1}MIT \quad \affmark{2}Amazon Robotics}
\thanks{$^{*}$These authors contributed equally. Both were research assistants at MIT CDFG.}
\thanks{$^{\dagger}$This work is unrelated to the author's position at Amazon.}}

\maketitle
\vspace*{-14mm}
\begin{abstract}

Differentiable simulators have advanced policy learning and model-based control across diverse robotic tasks.
To date, actuator dynamics remain underexplored and can be a major source of sim-to-real error, especially on low-cost platforms where the linear current-to-joint-torque approximation $\tau = K_t I$ becomes unreliable under commanded-target tracking because of friction, hysteresis, backlash, and thermal effects.
Beyond forward dynamics, accurate actuator models also support force perception and integrated force/position control in manipulation tasks.
We present \name, a neural actuator model that jointly predicts (i) a torque surrogate for trajectory propagation on low-cost servo platforms, (ii) external forces together with a contact-probability gate for sensorless force perception, and (iii) a motor-condition score for the supervised joint, distinguishing normal from mechanically restricted operation.
We introduce a twin-arm teleoperation system that collects robot states and actuator telemetry alongside external-force labels, yielding the Neural Actuation Dataset~(\dbname).
The torque-surrogate head is trained through differentiable simulation from pose trajectories without ground-truth joint-torque measurements.
A Transformer-based architecture captures temporal dependencies while enabling efficient real-time inference.
We validate \name across three platforms---a 5-DoF OpenManipulator-X, a 6-DoF SO-101 from LeRobot, and a 7-DoF Franka Emika Panda---spanning three actuator families and costs from approximately \$500 to more than \$30{,}000.
The low-cost platforms support physically plausible dynamics and enable force evaluation, while the offline Franka experiment provides an additional payload-force-estimation benchmark. We also demonstrate motor-condition estimation and improved behavior-cloning performance when \name is used as a pretrained module.
We release the dataset, code, and hardware configurations on the \href{https://frank-zy-dou.github.io/projects/NeuralActuator/index.html}{\textcolor[RGB]{163,31,52}{project page}}.

\end{abstract}

\IEEEpeerreviewmaketitle

\section{Introduction}

\begin{figure}[t]
    \centering
    \vspace{-9mm}
    \begin{overpic}[width=0.82\linewidth]{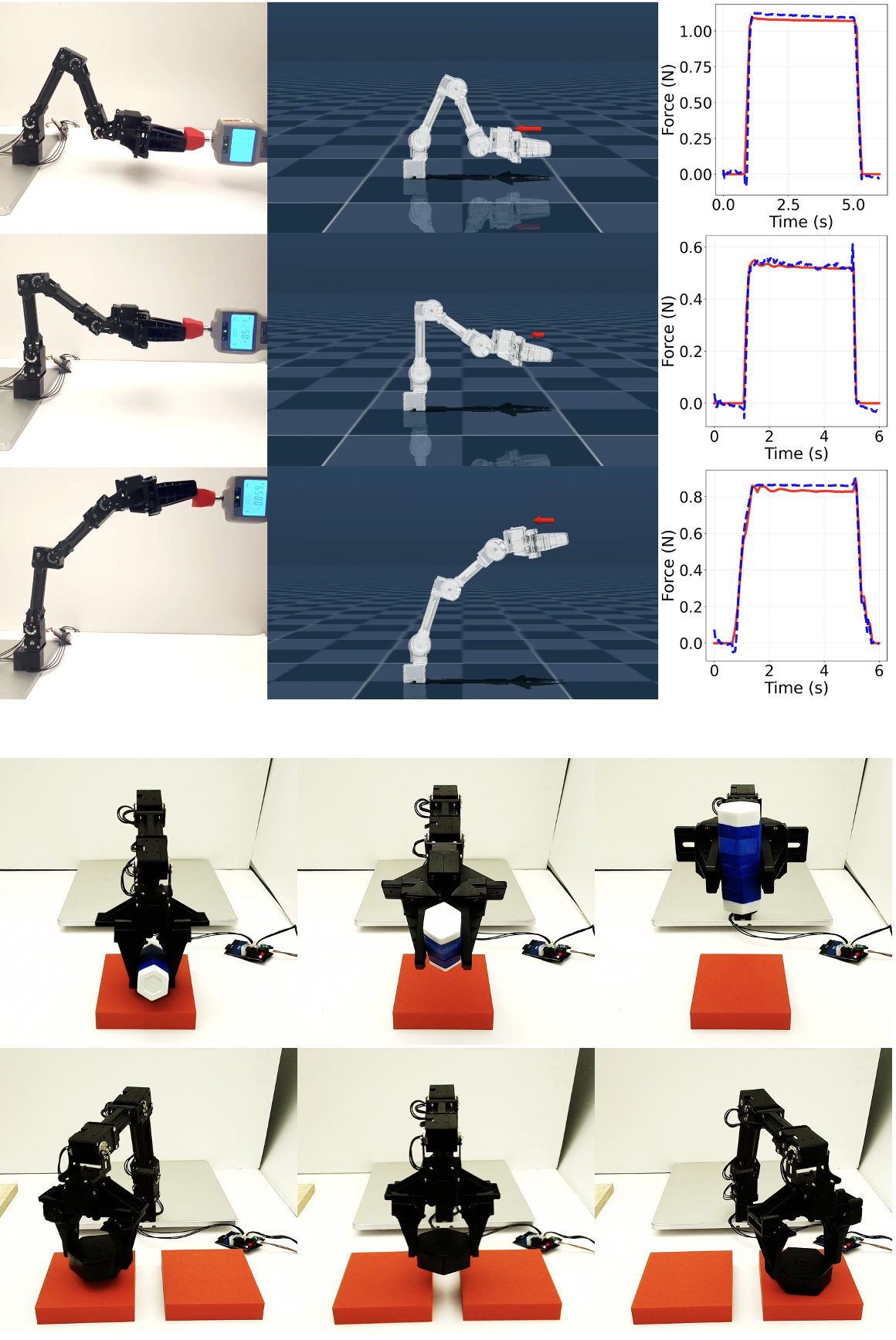}
        \put(32,44.5){(a)}
        \put(32,-2){(b)}
    \end{overpic}
    \vspace{1mm}
    \caption{\textbf{(a) Push--force gauge validation at three end-effector heights.} We show \name pushing a force gauge at \textit{low}, \textit{middle}, and \textit{high} end-effector heights (top to bottom). Left: real-robot execution; middle: corresponding simulated rollout; right: contact-axis force-magnitude profiles. The trace-level mean absolute errors~(MAEs) are 0.037\,N, 0.015\,N, and 0.028\,N, respectively. These values correspond to the three representative front-push sequences shown; Tab.~\ref{tab:gauge-pushing} reports aggregate componentwise force errors across positions, front/top contact directions, and rollout horizons. The \dashedleg\ shows the predicted magnitude, and the \solidleg\ shows the force-gauge ground truth. \textbf{(b) Estimated forces for downstream tasks.} \name provides online force feedback to behavior-cloning controllers for real-robot object lifting and holding (top) and pick-and-place (bottom) tasks.}
    \label{fig:teaser}
    \vspace{-8mm}
\end{figure}
Differentiable simulators~\cite{hu2019difftaichi, warp2022, brax2021github,howell2022dojo,authors2024genesis} enable efficient gradient-based optimization for robot control and robot learning, with substantial advances in modeling rigid-body dynamics and recent extensions to soft bodies and fluids.
However, actuator dynamics remain comparatively underexplored.
A common control approximation assumes that joint torque scales linearly with current, e.g., $\tau = K_t I$.
Although reasonable for well-calibrated industrial actuators, this approximation becomes unreliable on cost-effective, servo-driven platforms~\cite{cadene2024lerobot, interbotix_phantomx_reactor, tribotix_pincherx150, robotis_openmanipulator_x}.
These platforms often use mechanically commutated actuators (e.g., coreless brushed DC servo modules) whose hardware simplicity is accompanied by significant non-idealities: gearbox friction, hysteresis, backlash, current saturation, and thermal drift affect joint-side effort. Because the relevant internal friction and transmission states are inaccessible, these time-varying effects resist analytical modeling, resulting in poor effort tracking, unstable impedance behavior, and degraded sim-to-real transfer.

Beyond accurate modeling, a reliable actuator model also supports \emph{force perception without dedicated force/torque sensors at inference time}.
Recent studies demonstrate torque-based sensing for e-skin--like contact perception~\cite{iskandar2024intrinsic}, proprioceptive estimation in legged robots~\cite{fu2025unitac}, and force estimation and control in manipulation~\cite{kobayashi2025ilbit,hanai2023force,collins2023visual,zhi2025learning}.
Such torque-based methods, however, generally rely on high-quality effort feedback; when it is unavailable on low-cost platforms, an unreliable current--torque mapping undermines analogous force inference.
Because physical interaction is fundamentally mediated by force, contact-force prediction can support manipulation without precise geometric models of arbitrary or deformable objects.

We introduce \textbf{\name}, a data-driven actuation modeling method for low-cost robot platforms that captures actuator behavior and its coupling with robot dynamics.
\name is integrated with a differentiable simulator, enabling end-to-end training via gradient-based supervision from real-robot rollouts.
\name learns a history-dependent mapping from command and state histories together with actuator-side telemetry, including the platform-specific effort signal, voltage, and temperature where available, to torque surrogates, external forces, contact probabilities, and a motor-condition score that distinguishes normal from mechanically restricted operation on the supervised joint.
Prior actuator models use torque targets obtained either from joint-torque sensing~\cite{hwangbo2019learning} or from calibrated motor-current estimates~\cite{schwendeman2023improving}; both forms of supervision are difficult to obtain reliably on many low-cost servos.
Instead, we supervise the torque-surrogate head using pose trajectories: forward integration through our differentiable dynamics maps the predicted generalized inputs to future configurations, making pose an effective surrogate signal.
To capture nonlinear, time-varying actuator effects, \name adopts a Transformer architecture that processes recent state-and-command sequences through multi-head self-attention, modeling temporal dependencies and inter-joint coupling.
It outputs per-actuator torque surrogates, external forces, contact probabilities, and motor-condition scores in a multi-task manner.

To train and benchmark \name, we introduce a real-robot dataset capturing actuation, motion, and force signals.
We build a twin-arm leader--follower teleoperation system where an operator kinesthetically drives a leader arm to generate diverse trajectories, while the follower executes joint-space commands under closed-loop control.
For each trajectory, we log (i) robot states (joint positions, velocities, commanded targets), (ii) actuator telemetry (motor currents, temperature, supply voltage), and (iii) external-force labels.
The dataset spans free-space motions and contact-rich manipulation (pushing, pulling, lifting, known-weight loads), covering actuator effects induced by friction, load changes, and intermittent contacts.
We release it as the \textbf{Neural Actuation Dataset (NAD)}, providing synchronized supervision data for model training and benchmarking.

We conduct several experiments across three platforms---a cost-effective, servo-driven 5-DoF OpenManipulator-X (four revolute joints and a 1-DoF gripper)~\footnote{\url{https://emanual.robotis.com/docs/en/platform/openmanipulator_x/overview/}}, a 6-DoF SO-101 low-cost arm built on the LeRobot stack~\cite{cadene2024lerobot}, and a 7-DoF Franka Emika Panda industrial arm (Sec.~\ref{sec:cross_platform})---spanning three actuator families and costs from approximately \$500 to more than \$30{,}000.
We evaluate modeling fidelity on the low-cost arms, cross-platform external-force estimation including an offline Franka benchmark, online adaptation, throughput, and downstream real-robot control.
On held-out rollouts, \name yields low rollout errors (Sec.~\ref{sec:rollout_acc_no_load}), accurate external-force estimates (Sec.~\ref{sec:rollout_acc_with_force}), and accurate motor-condition classification (Sec.~\ref{sec:motor_condition}).
The evaluated implementation provides sub-millisecond GPU inference at the 60\,Hz control rate with a modest parameter footprint (Sec.~\ref{sec:runtime_performance}).
We also evaluate online adaptation using 12 newly collected trajectories (Sec.~\ref{sec:online_adaptation}).
For downstream real-robot control, \name provides online force feedback to behavior-cloning policies for object manipulation.
We show that learning actuator behavior together with external force perception improves task success rates (Sec.~\ref{sec:imitation_learning}).
Finally, we demonstrate that combining differentiable simulation with silhouette-based image-space supervision provides a visual training signal for \name (Sec.~\ref{sec:diff_rendering}).
We summarize our contributions below:
\begin{itemize}
    \item A data-driven, differentiable actuator model named \textbf{\name} that captures actuator behavior and its coupling with robot dynamics via neural torque-surrogate prediction, with additional outputs for external force prediction, contact detection, and motor-condition estimation---enabling force perception without dedicated force/torque sensors at inference time---and a training scheme that leverages differentiable physics without ground-truth joint-torque measurements.
    \item A twin-arm teleoperation system for efficient data collection on low-cost arms, time-synchronizing robot proprioception, commanded targets, and actuator-side telemetry.
    \item \textbf{Neural Actuation Dataset (\dbname)}: a real-robot dataset with time-aligned robot states, actuator telemetry, and end-effector external-force labels, enabling training and benchmarking for low-cost actuator modeling.
    \item Experiments across three platforms evaluate actuator-dynamics modeling, sensorless force estimation, motor-condition estimation, downstream manipulation, and computational efficiency.
\end{itemize}

\section{Related Work}

\subsection{Learning Neural-Augmented Simulation}
To close the sim-to-real gap, prior work has augmented analytical simulators with learned residual models; for example, the approach in~\cite{ajay2018augmenting} uses a stochastic RNN to capture residual dynamics and align simulation with reality.
Jiang et al.~\cite{jiang2022data} augment a rigid-body simulator with a data-driven contact model that predicts aggregate 3D contact impulses while retaining analytical contact constraints.
TossingBot~\cite{zeng2020tossingbot} further combines an analytic ballistic prior with a neural residual learned from visual features.
NeuralSim~\cite{heiden2021neuralsim} embeds neural modules in differentiable simulators to capture unmodeled dynamics and jointly optimizes them with physical parameters on real data; at the component level, Serifi et al.~\cite{serifi2023transformer} use a Transformer to predict residual state corrections for actuators and rigid bodies from simulated states and interaction forces.
Recently, \citet{xu2025neural} present NeRD, which learns robot-specific dynamics models for predicting future states for articulated rigid bodies under contact constraints.
Hwangbo et al.~\cite{hwangbo2019learning} train a supervised actuator network from physical-system data and use it in the simulation loop to model each joint of ANYmal; the network maps state histories and position targets to joint torques for simulation.
\citet{schwendeman2023improving} introduce a system identification method that combines a physics model with a neural network to learn residual dynamics, enabling domain-consistent dynamics modeling.
In their setting, joint torques are estimated from motor-current measurements.
Accordingly, these approaches use torque targets obtained either from joint-torque sensing~\cite{hwangbo2019learning} or from a calibrated current--torque estimate~\cite{schwendeman2023improving}; both forms of supervision are difficult to obtain reliably on many low-cost servo platforms.
In contrast, the torque-surrogate head in \name requires no direct generalized-effort labels: it is supervised through differentiable simulation using pose trajectories, without assuming reliable current--torque calibration.
Concurrently, \citet{fey2025bridging} learn residual corrective torques via reinforcement learning from tracking errors; \name instead predicts torque surrogates without assuming a fixed current--torque conversion.

\paragraph{Learning Actuator Models}
Existing work mostly targets permanent-magnet electrical machines, using neural models to estimate, control, or emulate electromagnetic torque from electrical states or simulation-derived features~\cite{yan2019pmsm_deeplearn_torque,liu2024rbfnn_dtc_torque,wang2025pinn_pmsm_torque,tahkola2020ann_torque_surrogate}.
These approaches concern electronically commutated machine models rather than the low-cost integrated coreless DC servo dynamics studied here.
Prior work also covers hydraulic actuators~\cite{kim2021force} and twisted-string actuators~\cite{kwon2024learning}.
Unlike these efforts, \name targets low-cost \emph{coreless DC} servo actuators, where friction, backlash, saturation, and thermal drift can make the effective current--torque relationship nonlinear and history-dependent.

\subsection{Force Estimation and Proprioceptive Sensing}
\paragraph{Torque- and effort-based contact inference}
A long-standing line of work reconstructs external joint torques from model discrepancies and uses them for collision detection and safe reaction~\cite{de2005sensorless,de2006collision,haddadin2008collision}.
Building on reliable effort feedback, recent systems further elevate this idea to richer ``intrinsic'' contact perception (e.g., e-skin--like touch without tactile skins)~\cite{iskandar2024intrinsic}, whole-robot touch sensing without dedicated tactile sensors~\cite{fu2025unitac}, and learning-based controllers that couple force and motion in manipulation and loco-manipulation~\cite{kobayashi2025ilbit,zhi2025learning}.
These approaches highlight the growing role of actuation as an implicit sensing channel, but they often rely on high-quality torque measurements or well-calibrated current--torque mappings.

\paragraph{External force estimation on robot arms}
For manipulators, a complementary body of work estimates external forces and contacts by explicitly modeling the robot dynamics and attributing residual terms to interaction wrenches.
Representative examples include virtual force sensing and contact localization along the robot structure~\cite{magrini2014estimation,manuelli2016localizing}, as well as observer and filtering formulations that improve robustness under friction and model uncertainty~\cite{hu2017contact,liu2021sensorless,han2021toward,linderoth2013robotic}.
Recently, learning has also been used to directly map joint-level proprioception and measured joint torques from sensors to end-effector wrenches in static or quasi-static regimes~\cite{osburg2022using}.
\citet{chen2025learning} leverage differentiable simulation to infer external object properties solely from proprioceptive signals.
Despite this progress, most proprioceptive force-estimation methods assume accurate torque feedback, while the setting of low-cost, servo-driven arms remains largely unexplored.
A summary is provided in Tab.~\ref{supp_tab:force_sensing_taxonomy} in the Appendix.

\section{Method}
\label{sec:method}

\begin{figure}[t]
\vspace{2mm}
  \centering
  \begin{overpic}[width=0.95\linewidth]{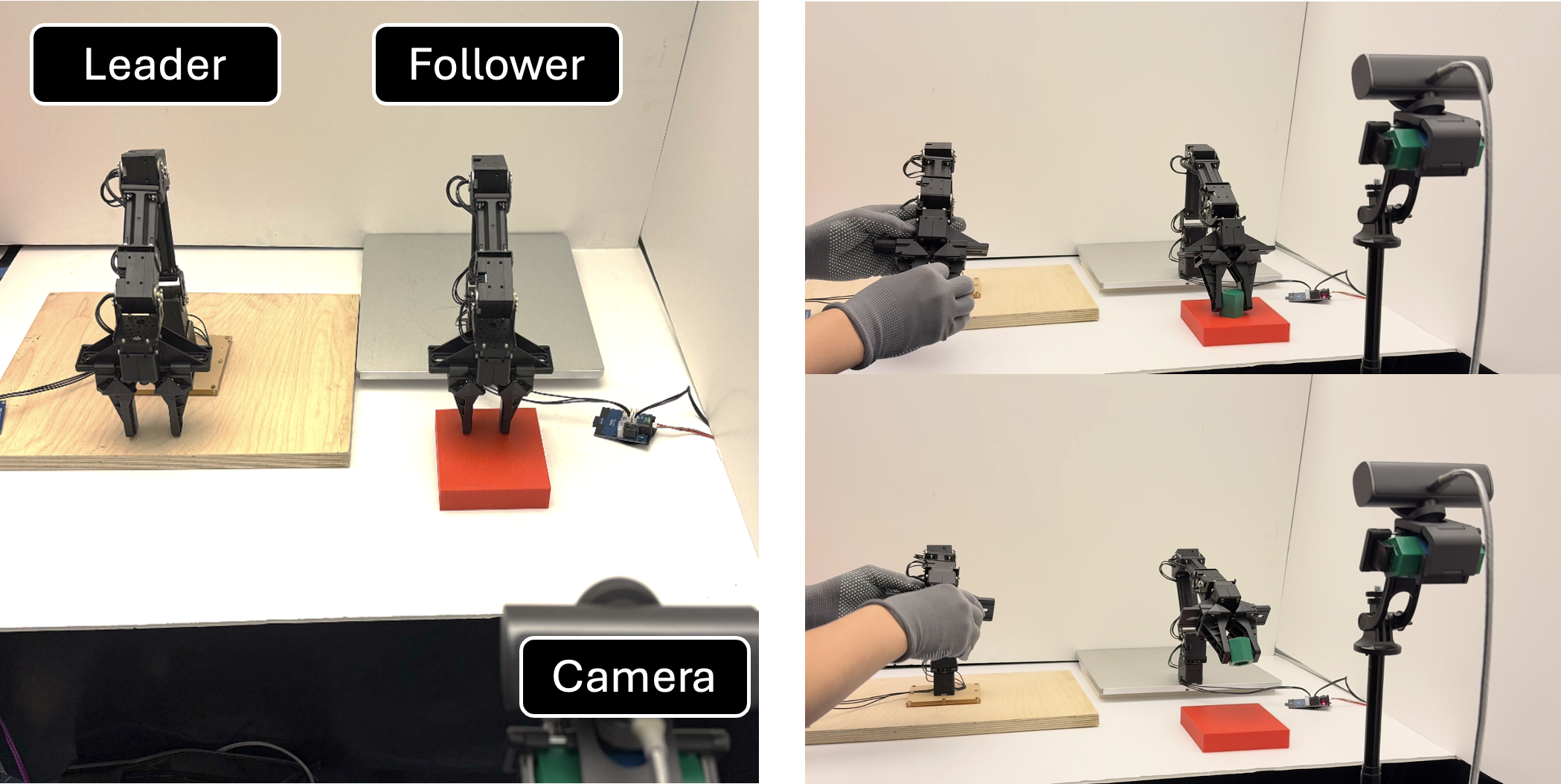}
    \put(23,-4.5){(a)}
    \put(73,-4.5){(b)}
  \end{overpic}
  \vspace{3mm}
\caption{\textbf{Neural Actuation Dataset (NAD) data collection.}
(a) Overview of the twin-arm leader--follower system with an external camera.
(b) An operator kinesthetically drives the leader to generate diverse motions; the follower mirrors these motions while manipulating a payload and recording synchronized actuator currents, robot states, and end-effector forces.}
  \label{fig:NAD_collection}
  \vspace{-7mm}
\end{figure}

\begin{figure*}[t]
\vspace{-4mm}
  \centering
\includegraphics[width=\linewidth]{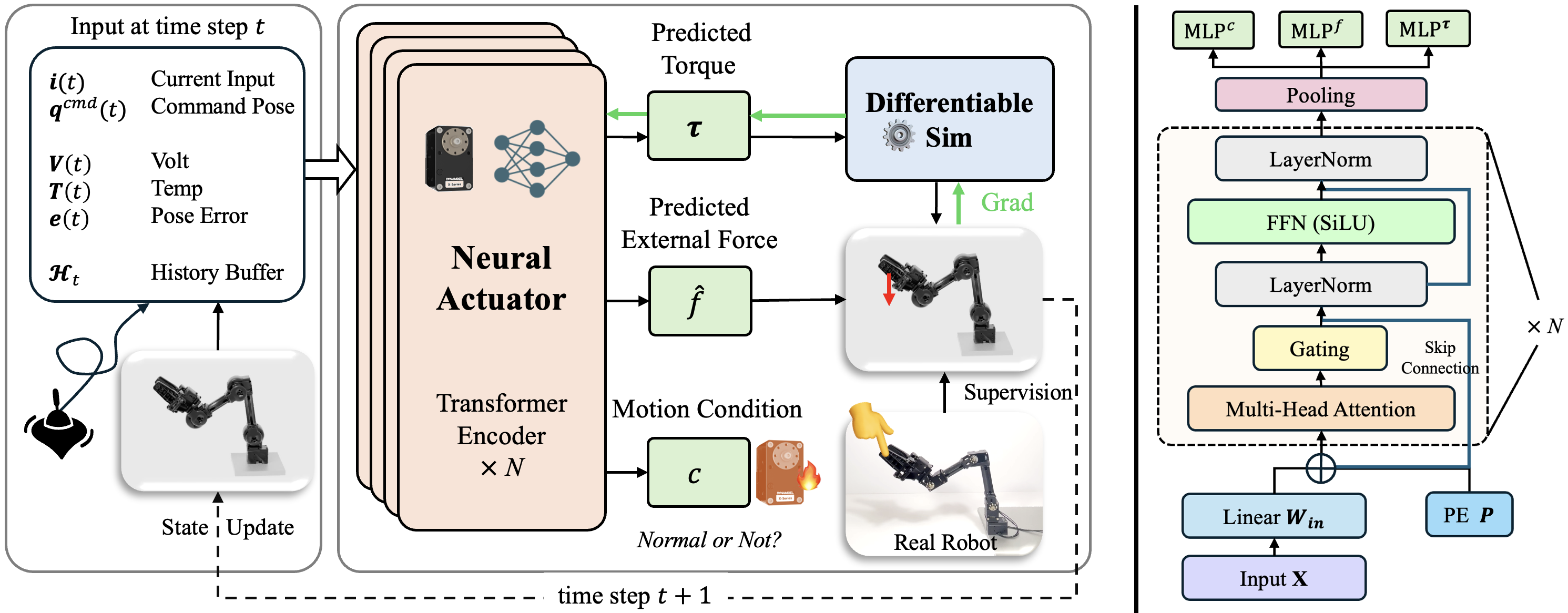}
  \vspace{-4mm}
  \caption{\small\textbf{\name\ pipeline.}
At each time step $t$, \name takes the commanded pose $q^{\mathrm{cmd}}(t)$, an effort-related actuator signal $u(t)$, actuator-side telemetry, and tracking feedback $e(t)$, together with a history buffer $\mathcal{H}_t$ that summarizes recent commands, states, and telemetry (e.g., joint poses and velocities).
A Transformer encoder ($\times N$ blocks) maps the sequence to four task heads:
(i) a torque surrogate $\boldsymbol{\tau}$,
(ii) a raw external-force estimate $\hat{\mathbf{f}}_{\mathrm{raw}}$,
(iii) a contact-probability gate $g$, and
(iv) a per-motor condition vector $\mathbf{c}$.
The gated force is $\hat{\mathbf{f}}=g\hat{\mathbf{f}}_{\mathrm{raw}}$.
The torque surrogate supplies the simulator's generalized control input; the simulator then advances the state to time $t{+}1$ and updates the history buffer. See Sec.~\ref{sec:manipulator_connection} for a discussion of the torque-surrogate and external-force outputs.
The torque-surrogate head is trained through \textit{Diff Sim} from real-robot state transitions. \textbf{Right:} High-level schematic of the Transformer and output heads. The implementation uses the pre-normalized gated-attention and GELU encoder blocks specified in the text; the condition branch represents per-motor condition scores, and the separately implemented contact gate is omitted for visual clarity.}
  \label{fig:pipeline}
  \vspace{-6mm}
\end{figure*}

\subsection{Problem Formulation}

Consider a robotic manipulator with $n_a$ arm joints and $n_m$ modeled actuator channels, including a gripper channel when applicable. For OpenManipulator-X, $n_a=4$ and $n_m=5$. Let $\tau_j^{\text{real}}$ denote the physical joint-side effort for actuator channel $j$ (torque for a revolute joint). It depends on the commanded position, effort-related actuator telemetry, state, and operating history:
\begin{equation}
\tau_j^{\text{real}}(t) = f_\tau(q_j^{\text{cmd}}(t), u_j(t), q_j(t), \dot{q}_j(t), V_j(t), T_j(t), \mathcal{H}_t)
\label{eq:real_torque}
\end{equation}
where $q_j^{\text{cmd}}(t)$ is the commanded (goal) position, $q_j(t)$ and $\dot{q}_j(t)$ are joint position and velocity, and $V_j(t)$ and $T_j(t)$ denote the supply voltage and thermal state, respectively.
The platform-specific signal $u_j$ is motor current on OpenManipulator-X, the signed load register on SO-101, and commanded joint torque on Franka.
Let $\mathbf{x}_t$ collect the command, proprioceptive state, and actuator telemetry at time $t$.
The model used in our experiments takes the eight preceding feature vectors
$\mathcal{H}_t=\{\mathbf{x}_{t-8},\ldots,\mathbf{x}_{t-1}\}$ and appends the current vector $\mathbf{x}_t$, yielding the nine-token sequence
$\mathbf{X}_t=(\mathbf{x}_{t-8},\ldots,\mathbf{x}_t)$.

In parallel with the learned torque-surrogate output, we predict external forces and motor-condition scores from the same input features:
\begin{equation}
\begin{aligned}
\hat{\mathbf{f}}^{\text{ext}}(t)
&=f_{\text{force}}\!\Big(q^{\text{cmd}}(t),u(t),q(t),\\
&\hspace{16mm}\dot q(t),V(t),T(t),\mathcal H_t\Big)\in\mathbb R^3
\end{aligned}
\label{eq:force_pred}
\end{equation}
\begin{equation}
\mathbf c(t)=f_{\text{cond}}\!\left(\mathbf X_t\right)\in[0,1]^{n_m}
\label{eq:cond_pred}
\end{equation}

where $\hat{\mathbf{f}}^{\text{ext}}(t)$ is the predicted end-effector external force.
The architecture emits one condition score $c_j(t)$ per actuator channel. In the OpenManipulator-X condition benchmark, only the Joint~3 component is supervised and evaluated: $c_3$ estimates the probability of normal operation, and $c_3=0$ denotes the mechanically restricted condition. We do not assign a calibrated probabilistic interpretation to the unsupervised components in this experiment.

\subsection{Physical Interpretation through Manipulator Dynamics}
\label{sec:manipulator_connection}

We define $\mathbf{f}^{\text{ext}}$ as the force exerted by the environment on the robot, matching the sign of the recorded force labels, and write
\begin{equation}
\begin{aligned}
\boldsymbol{\tau}_{\mathrm{ID}}
&=M(q)\ddot q+C(q,\dot q)\dot q+\mathbf{g}(q) \\
&=\boldsymbol{\tau}_{\mathrm{act}}+\boldsymbol{\tau}_{\mathrm{ext}}, \\
\boldsymbol{\tau}_{\mathrm{ext}}
&=J_v(q)^\top\mathbf{f}^{\text{ext}}.
\end{aligned}
\label{eq:manipulator_connection}
\end{equation}
Here, $q\in\mathbb{R}^{n_a}$ contains the arm coordinates.
The quantities $M(q)$, $C(q,\dot q)\dot q$, and $\mathbf{g}(q)$ denote the joint-space inertia matrix, Coriolis and centrifugal generalized forces, and gravity generalized forces encoded by the simulator, respectively; $\boldsymbol{\tau}_{\mathrm{act}}$ denotes the physical joint-side actuator torque.
The translational Jacobian $J_v(q)$ is evaluated at the force-reference point; using a three-dimensional force rather than a six-dimensional wrench assumes that the residual moment about that point is negligible.
The gripper geometry is described separately in Sec.~\ref{sec:gripper_model}.
This convention gives
$\boldsymbol{\tau}_{\mathrm{ext}}=\boldsymbol{\tau}_{\mathrm{ID}}-\boldsymbol{\tau}_{\mathrm{act}}$,
which is the residual used by the classical force-estimation baselines.
Those methods approximate the actuator term by a linear current--torque model,
$\boldsymbol{\tau}_{\mathrm{act}}=\mathbf K_t\mathbf{i}+\mathbf{b}$, where $\mathbf K_t=\operatorname{diag}(K_{t,1},\ldots,K_{t,n_a})$, and map the resulting residual to Cartesian force.
On geared low-cost servos, however, friction, backlash, saturation, and thermal drift make this linear map unreliable; its errors therefore contaminate the inferred contact force.

\name predicts a torque surrogate $\boldsymbol{\tau}^{\mathrm{pred}}$ and an external-force estimate $\hat{\mathbf{f}}^{\mathrm{ext}}$ with dedicated heads. We consider two ways to couple the force prediction to the differentiable rollout. Under \emph{implicit coupling}, $\boldsymbol{\tau}^{\mathrm{pred}}$ alone drives the simulated body, while $\hat{\mathbf{f}}^{\mathrm{ext}}$ is directly supervised but is not applied to the differentiable simulator. The torque surrogate may therefore absorb interaction effects needed to reproduce the observed motion. Under \emph{explicit coupling}, $\hat{\mathbf{f}}^{\mathrm{ext}}$ is applied at the force-reference point and contributes $J_v(q)^\top\hat{\mathbf{f}}^{\mathrm{ext}}$ alongside $\boldsymbol{\tau}^{\mathrm{pred}}$, so the predicted interaction load also enters the state update. The rollout loss accordingly backpropagates through the applied force. Sec.~\ref{sec:force_coupling_ablation} compares the two coupling choices.

Independently of force coupling, we consider direct and residual parameterizations of the torque surrogate:
\begin{equation}
\begin{aligned}
\text{direct:}\quad
&\boldsymbol{\tau}^{\mathrm{pred}}_t
=g_\theta(\mathbf X_t),\\
\text{residual:}\quad
&\boldsymbol{\tau}^{\mathrm{pred}}_t
=\boldsymbol{\tau}_{\mathrm{base}}(u_t)
+\Delta g_\theta(\mathbf X_t).
\end{aligned}
\label{eq:torque_parameterizations}
\end{equation}
For instance, for the OpenManipulator-X arm joints, $\boldsymbol{\tau}_{\mathrm{base}}=\mathbf K_t\mathbf i_t$ with $K_t=1.3\,\mathrm{N\,m/A}$. We use the direct form because it avoids imposing a fixed linear telemetry-to-torque prior, whereas the residual form retains that prior as a nominal anchor. Sec.~\ref{sec:torque_parameterization_ablation} compares the two parameterizations.

\subsection{Neural Actuation Dataset (\dbname)}
To learn the actuator model, we first collect synchronized robot-state, actuator-telemetry, and force data using a twin-arm teleoperation system built on two identical OpenManipulator-X robots~\cite{robotis_openmanipulator_x}.
The system consists of a \emph{leader} arm kinesthetically operated by a human demonstrator and a \emph{follower} arm that mirrors the leader's motion while interacting with the environment (Fig.~\ref{fig:NAD_collection}).

The \emph{commanded target joint positions} sent to the follower's actuators are set to the leader's instantaneous joint states. Both arms are equipped with DYNAMIXEL XM430-W350 servos, which provide synchronized actuator-level measurements, including:
\begin{itemize}
    \item Joint positions $q_j(t)$ and velocities $\dot{q}_j(t)$
    \item Motor currents $i_j(t)$, PWM signals, and bus voltages
    \item Coil temperatures $T_j(t)$
\end{itemize}

For a subset of interaction trajectories, interaction force is measured with a fixture-mounted six-axis force/torque sensor. We use its three force channels as ground-truth end-effector contact-force measurements $\mathbf{f}^{\text{gt}}(t) \in \mathbb{R}^3$; the three moment channels are not used for supervision. Before training, force labels are transformed into the robot base frame and retain the environment-on-robot sign convention used in Eq.~\ref{eq:manipulator_connection}.

All data streams are temporally synchronized and logged per trajectory. For each trajectory of length $T$, we record time-aligned tuples containing (i) robot states (e.g., joint positions, velocities, and commanded targets), (ii) motor currents and actuator-side telemetry, and (iii) end-effector external forces. Together, these streams support pose-based surrogate learning and force estimation. The OpenManipulator-X model-development and evaluation subset used in our experiments comprises the following three components:

\noindent \textbf{Free motion (no external force).}
This subset characterizes intrinsic actuator dynamics and provides force-free baselines. We record a diverse set of trajectories (119,933 frames, avg. 20.49\,s), including (a) clockwise and counterclockwise circular end-effector motions, (b) individual joint sweeps covering approximately $90\%$ of feasible ranges, and (c) whole-arm primitives such as \emph{lean and extend}. These trajectories contain no intentional external contact.

\noindent \textbf{Force-labeled interactions.}
This subset provides force labels via two modalities.
\textit{(a) Known-weight loads.}
The experimental subset uses payloads with $m\in\{200,300,400,500\}\,\text{g}$; the full NAD dataset additionally contains 100\,g payload trajectories.
During \emph{loaded/grasped} intervals, we assign the nominal base-frame load $\mathbf{f}^{\text{gt}}_{\text{grav}}=m\,\mathbf{g}_{\text{base}}=[0,0,-mg]^\top$. This reference is exact for static or quasi-static holding; during motion, it provides nominal supervision that does not account for object inertia or gripping friction. Tasks include \emph{go up and stay still} (persistent load) and \emph{pick and place} (intermittent load during grasping), totaling 88{,}611 frames with an average duration of 21.63\,s.
\textit{(b) Axis-aligned sensor interactions.} Operators apply controlled push and pull forces along the force sensor's principal axes ($\pm X, \pm Y, \pm Z$). For each force direction, we also record matched no-interaction trajectories that follow the same motion patterns but are executed without physical contact. The measured wrench is transformed into the robot base frame to capture full 3D force components arising from both dominant-axis interactions and kinematic coupling (74{,}020 frames, avg. 10.50\,s).

We also collect data for the cross-platform evaluation in Sec.~\ref{sec:cross_platform}, using an SO-101 arm logged via the LeRobot stack~\cite{cadene2024lerobot} and a 7-DoF Franka Emika Panda industrial arm.
Franka provides an additional offline payload-force-estimation benchmark.

\begin{figure}
    \centering
    \includegraphics[width=\linewidth]{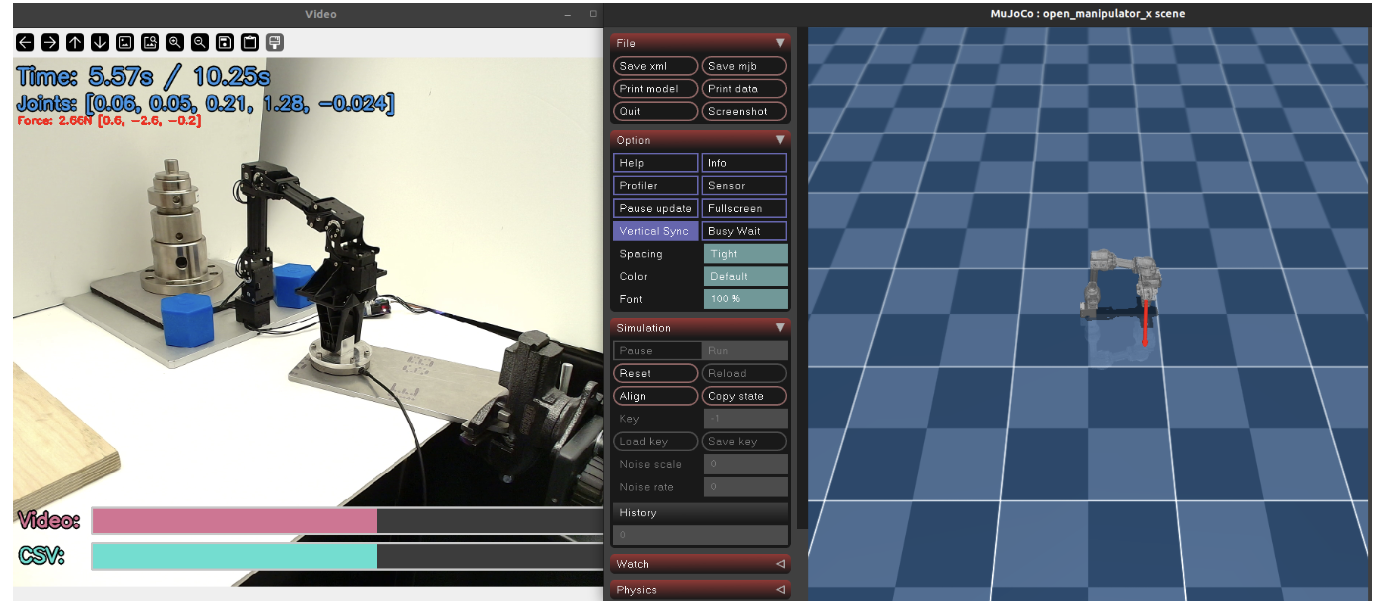}
    \vspace{-3.5mm}
\caption{\textbf{Data verification in NAD.}
We time-synchronize the real-robot video (left) with the logged trajectory and verify alignment in a GUI (right), where the red arrow visualizes the estimated external force.}
\label{fig:data_verification}
    \vspace{-2mm}
\end{figure}

\begin{wrapfigure}{r}{34mm}
  \vspace{-5mm}
  \hspace*{-4mm}
  \centerline{
  \includegraphics[width=35mm]{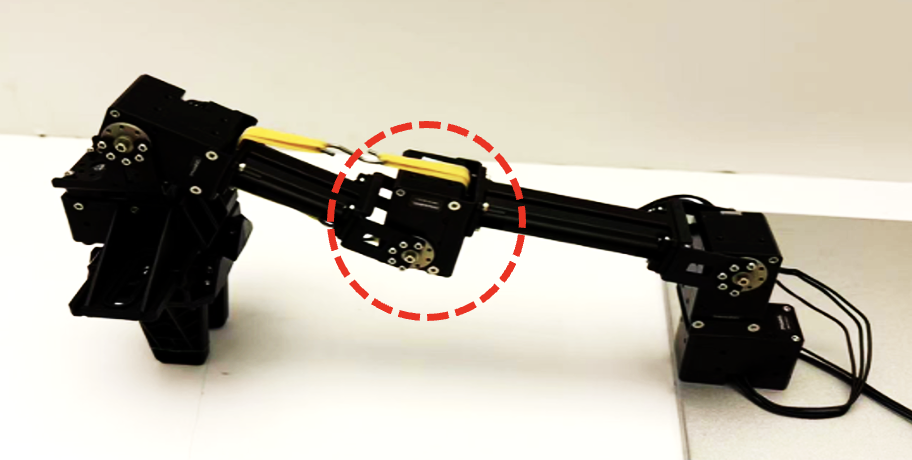}}
  \vspace*{-7mm}
\end{wrapfigure}
\noindent \textbf{Mechanically restricted motor operation.} In this task, we constrain Joint~3 with rubber bands to introduce additional mechanical resistance; see the inset. Under this asymmetric restriction, we collect \emph{pick and place} trajectories with and without a 200\,g payload (49{,}898 frames, avg. 21.20\,s), enabling disentanglement of contact-induced and condition-induced deviations.

We split the dataset into training, validation, and test sets in an 8:1:1 ratio.
All data in \dbname are calibrated and human-verified as shown in Fig.~\ref{fig:data_verification}. See Appendix~\ref{sec:appendix_dataset} for additional details.

\begin{table}[t]
\vspace{0mm}
\centering
\small
\renewcommand{\arraystretch}{1.15}
\caption{
\textbf{OpenManipulator-X model-development and evaluation subset.}
The 94.52 minutes sum the task-assignment durations used in the experiments; nominal trajectories reused for motor-condition comparison also appear in the free-motion and force-labeled categories.
See Appendix~\ref{sec:appendix_dataset} for a detailed task-level breakdown.}
\vspace{-1mm}
  \resizebox{\columnwidth}{!}{
\begin{tabular}{l l c}
\toprule
\textbf{Component} & \textbf{Description} & \textbf{Duration (min)} \\
\midrule
Free motion
& No external force
& $\sim 34.15$ \\

Force-labeled
& Known payloads or measured force
& $\sim 46.24$ \\

Motor condition
& Mechanically restricted
& $\sim 14.13$ \\
\midrule
\textbf{Total} &  & $\sim 94.52$ \\
\bottomrule
\end{tabular}}

\label{tab:dataset_summary}
\vspace{-2mm}
\end{table}

\begin{figure*}[t]
\vspace{-3mm}
  \centering
  \begin{overpic}[width=\linewidth]{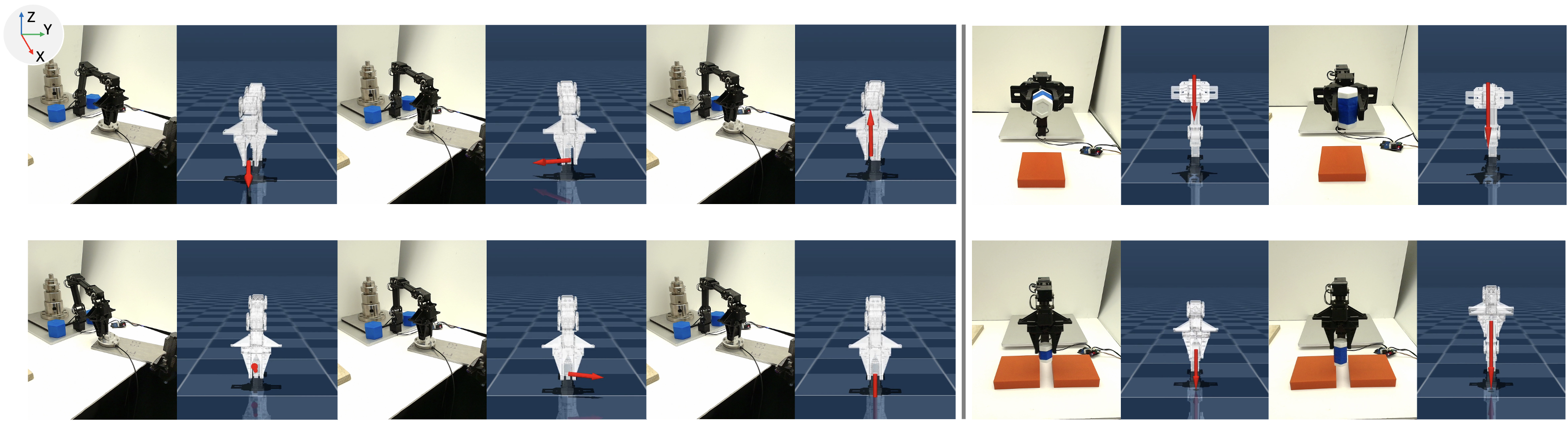}
  \end{overpic}
  \vspace{-5mm}
  \caption{\textbf{Robot arm simulation with external force estimation on the test set.}
  (a) \name estimates end-effector external forces along six directions (\(\pm X, \pm Y, \pm Z\)); red arrows visualize the estimated force vectors in simulation (paired with corresponding real-robot executions).
  (b) Force estimation during force-aware manipulation: lifting and holding 200\,g and 300\,g payloads (top) and pick-and-place with 200\,g and 300\,g payloads (bottom).
  Additional examples are shown in the accompanying video.}
  \label{fig: force_sensor}
  \vspace{-7mm}
\end{figure*}

\subsection{\name}

To capture temporal dependencies and nonlinear actuator dynamics, we employ a Transformer encoder that processes recent commands, states, and telemetry to produce multi-task actuator and interaction estimates.
In \name, the input sequence $\mathbf{X}_t\in\mathbb{R}^{L\times n_f}$ stacks per-time-step feature vectors $\mathbf{x}_t\in\mathbb{R}^{n_f}$.
It comprises $H_{\mathrm{hist}}=8$ historical frames followed by the current frame, so $L=H_{\mathrm{hist}}+1=9$.
In the primary low-cost configurations, each $\mathbf{x}_t$ aggregates three families of signals---(i)~commanded targets ($q^{\text{cmd}}, q_g^{\text{cmd}}$), (ii)~proprioception ($q, \dot{q}, q_g$), and (iii)~actuator telemetry ($u$, $V$, $T$)---together with tracking-error features ($e=q^{\text{cmd}}-q,\ e_g=q_g^{\text{cmd}}-q_g$). All features are normalized before being fed to the network. The Franka experiment uses the platform-specific feature vector described in Appendix~\ref{sec:appendix_impl}.
In simulation, the gripper feature $q_g$ is the single-finger prismatic-joint coordinate; it is distinct from the total physical jaw opening $a(\alpha)$ in Sec.~\ref{sec:gripper_model}.

An input projection and learned positional embeddings initialize
$\mathbf{H}_{\mathrm{enc}}^{(0)}=\mathbf{X}_t\mathbf{W}_{\mathrm{in}}+\mathbf{P}$,
where $\mathbf{W}_{\mathrm{in}}\in\mathbb{R}^{n_f\times d}$ and $\mathbf{P}\in\mathbb{R}^{L\times d}$.

We use pre-normalized residual blocks with a query-dependent elementwise gate on the attention output. For layer $l$,
\begin{align}
\mathbf{U}^{(l)}
&=\mathrm{LayerNorm}\!\left(\mathbf{H}_{\mathrm{enc}}^{(l-1)}\right), \\
\mathbf{A}^{(l)}
&=\left[
  \mathrm{MHA}\!\left(\mathbf{U}^{(l)}\right)
  \odot\sigma\!\left(\mathbf{U}^{(l)}\mathbf{W}_{g}^{(l)}\right)\right]\mathbf{W}_{o}^{(l)}, \\
\mathbf{Z}^{(l)}
&=\mathbf{H}_{\mathrm{enc}}^{(l-1)}+\mathbf{A}^{(l)}, \\
\mathbf{H}_{\mathrm{enc}}^{(l)}
&=\mathbf{Z}^{(l)}+\mathrm{FFN}_{\mathrm{GELU}}^{(l)}\!\left(\mathrm{LayerNorm}\!\left(\mathbf{Z}^{(l)}\right)\right).
\end{align}
Here, $\mathbf{W}_{g}^{(l)},\mathbf{W}_{o}^{(l)}\in\mathbb{R}^{d\times d}$, $\mathbf{U}^{(l)}\mathbf{W}_{g}^{(l)}$ is produced alongside the query projection, $\mathrm{MHA}$ denotes the concatenated per-head attention values before the output projection, and $\odot$ denotes elementwise multiplication.
After the final encoder layer and layer normalization, temporal mean pooling yields $\mathbf{h}_{\mathrm{pool}}\in\mathbb{R}^d$.
The configuration uses four encoder layers, hidden dimension $d=192$, four attention heads, and feed-forward dimension $d_{\mathrm{ff}}=384$.
Four output heads with SiLU hidden activations predict a torque surrogate, raw force, contact probability, and per-motor condition scores.

\noindent
\textbf{1. Torque-Surrogate Prediction.} \name predicts a torque surrogate as
\begin{equation}
\boldsymbol{\tau}^{\mathrm{pred}}(t)
=g_{\theta}\!\left(\,\mathbf{X}_t\right),
\label{eq:direct_torque}
\end{equation}
where $\boldsymbol{\tau}^{\mathrm{pred}}(t)\in\mathbb{R}^{n_m}$ denotes the torque-surrogate output and
$\mathbf{X}_t=(\mathbf{x}_{t-8},\ldots,\mathbf{x}_t)$ contains proprioception, commands, and actuator telemetry.
For the primary torque-actuated simulators, we use \emph{torque surrogate} as shorthand for this generalized effort: it is torque-valued on revolute channels and force-valued on a prismatic gripper channel.
The Franka configuration instead retains the robot model's stock affine PD position actuators; on that platform, the corresponding head supplies a simulator-control surrogate rather than a directly torque-valued input.
This design accommodates a nonlinear, history-dependent relationship between actuator telemetry and the effective generalized input required by the simulator.
The surrogate therefore avoids relying on a fixed linear current--torque prior.

A single MLP head jointly predicts the surrogate values for all $n_m$ actuator channels:
$\boldsymbol{\tau}^{\mathrm{pred}} = \mathrm{MLP}_{\tau}(\mathbf{h}_{\mathrm{pool}}) \in \mathbb{R}^{n_m}$,
where $\mathrm{MLP}_{\tau}$ uses SiLU hidden activations.

\noindent
\textbf{2. External Force Prediction.}
We predict end-effector external forces using a two-stage approach that decouples force regression from contact detection.
A force-specific hidden feature is first computed from the shared Transformer representation. Separate linear readouts from this feature predict the raw 3D force $\hat{\mathbf{f}}_{\mathrm{raw}}\in\mathbb{R}^3$ and the contact gate $g\in[0,1]$.
No dedicated force sensor is required at inference time.
The gate represents the probability of physical contact, and the final force output is $\hat{\mathbf{f}}=g\hat{\mathbf{f}}_{\mathrm{raw}}$.
During training, the gate target is $g_{\mathrm{gt}}=\mathbb{I}[\|\mathbf{f}_{\mathrm{gt}}\|_2>\epsilon]$.
We set $\epsilon=0.01$\,N solely to distinguish zero from nonzero supervision labels.

\noindent
\textbf{3. Motor Condition Estimation.}
A separate head predicts the per-motor condition vector
$\mathbf{c}=\sigma(\mathrm{MLP}_{\mathrm{cond}}(\mathbf{h}_{\mathrm{pool}}))\in[0,1]^{n_m}$.
The OpenManipulator-X benchmark supervises and evaluates only the Joint~3 component from the shared command, state, and telemetry history: a value near one indicates normal operation, whereas a value near zero indicates the mechanically restricted condition. The remaining output components are not interpreted as calibrated condition probabilities in this experiment.

\noindent
\textbf{Differentiable Simulation Integration.}
\name is integrated with differentiable physics simulators.
We embed the model within the simulation loop to enable gradient-based optimization through the system dynamics.
When differentiable rendering is enabled, image-space supervision follows the same computational path: the silhouette loss is backpropagated through the renderer, simulated configurations, and differentiable dynamics to the parameters of \name.

\noindent
\textit{Forward Dynamics.}
At time step $t$, the torque-surrogate head for the torque-actuated OpenManipulator-X and SO-101 simulators produces
$\boldsymbol{\tau}^{\mathrm{pred}}_t=g_\theta(\mathbf{X}_t)$.

Under implicit force coupling, the simulator advances the state using the predicted surrogate:
\begin{equation}
\begin{aligned}
\mathbf{s}_{t+1}
=F_t(\mathbf{s}_t,\boldsymbol{\tau}^{\mathrm{pred}}_t)
=\mathrm{DiffSim}\!\left(
\mathbf{s}_t,\boldsymbol{\tau}^{\mathrm{pred}}_t\right).
\end{aligned}
\label{eq:forward_dynamics}
\end{equation}
Here, $\mathbf{s}_t=[q_t,\dot q_t]$ is the simulated joint state.
The Franka configuration instead supplies the simulator-control surrogate to its affine PD position actuators, as described above.
Offline training and evaluation use the complete time-aligned effort-telemetry sequence recorded with each trajectory; the resulting multi-step evaluations are therefore recorded-telemetry-conditioned rollouts rather than forecasts of the actuator response to counterfactual future commands.
For online OpenManipulator-X deployment, prediction is strictly causal. At each control cycle, the robot reads the current state and available actuator telemetry, including the live measured current $\mathbf{i}_t$; \name then estimates the current surrogate, force, and condition scores, and a feedback controller can use $\hat{\mathbf f}_t$ to generate the next command. After inference, $\mathbf{x}_t$ is appended to the history used at the next control cycle; $\mathbf{x}_{t+1}$ enters only after the next telemetry packet arrives.
Online predictions therefore use no future current measurements.
Planning over counterfactual future commands would additionally require a model of the corresponding effort signal.

\noindent
\textit{Gradient Computation.}
For $\mathcal{L}=\sum_{k=1}^{T}\ell_k(\mathbf{s}_k)$, automatic differentiation accounts for the effect of $\boldsymbol{\tau}^{\mathrm{pred}}_t$ on $\mathbf{s}_{t+1}$ and performs backpropagation through time according to
\begin{align}
\frac{d\mathcal{L}}{d\theta}
&=\sum_{k=1}^{T}
\frac{\partial\ell_k}{\partial\mathbf{s}_k}
\frac{d\mathbf{s}_k}{d\theta}, \\
\frac{d\mathbf{s}_{t+1}}{d\theta}
&=
\frac{\partial F_t}{\partial\mathbf{s}_t}
\frac{d\mathbf{s}_t}{d\theta}
+
\frac{\partial F_t}{\partial\boldsymbol{\tau}^{\mathrm{pred}}_t}
\frac{d\,g_\theta(\mathbf{X}_t)}{d\theta}.
\label{eq:gradient_flow}
\end{align}
The total derivative includes the dependence of the simulated-state components in $\mathbf{X}_t$ on earlier predictions; consequently, losses at later steps backpropagate to preceding torque-surrogate predictions wherever the simulator sensitivity is nonzero.
Equivalently, the Markov state for this recurrence is the augmented state $(\mathbf{s}_t,\mathcal{H}_t)$; Eq.~\ref{eq:gradient_flow} suppresses the explicit history-buffer component for readability.

\noindent
\textit{Model Training.}
Since direct generalized-effort labels are unavailable on the low-cost platforms, the torque-surrogate head is supervised through pose trajectories: forward integration converts the predicted generalized inputs into future configurations, making measured poses effective surrogate targets.
We use Smooth L1 (Huber) regression losses. The following equations describe the primary OpenManipulator-X configuration.
Let $\operatorname{Huber}_{\beta}$ denote the elementwise Huber penalty with transition parameter $\beta$, and let $v_{t,k}\in\{0,1\}$ be the optional validity mask for force channel $k$.
The sets $\mathcal{T}_f$ and $\mathcal{T}_c$ index the input--label pairs used by the force, gate, and condition losses.
The regression terms are
\begin{align}
\mathcal{L}_{\mathrm{joint}}
&=\frac{1}{|\mathcal{T}|n_a}
\sum_{t\in\mathcal{T}}\sum_{j=1}^{n_a}
\operatorname{Huber}_{1}\!\left(
q_{t,j}^{\mathrm{sim}}-q_{t,j}^{\mathrm{real}}
\right), \\
\mathcal{L}_{\mathrm{grip}}
&=\frac{1}{|\mathcal{T}|}
\sum_{t\in\mathcal{T}}
\operatorname{Huber}_{1}\!\left(
\kappa_g(q_{g,t}^{\mathrm{sim}}-q_{g,t}^{\mathrm{real}})\right), \\
\mathcal{L}_{\mathrm{force}}
&=\frac{1}{3|\mathcal{T}_f|}
\sum_{t\in\mathcal{T}_f}\sum_{k=1}^{3}w_t v_{t,k}\,
\operatorname{Huber}_{\beta_f}\!\left(
\hat f_{t,k}-f_{t,k}^{\mathrm{gt}}\right).
\label{eq:regression_losses}
\end{align}
For the OpenManipulator-X prismatic gripper, $\kappa_g=1000$ converts the single-finger slide-coordinate residual from meters to millimeters before applying the loss.
The force loss uses $\beta_f=0.15$ for the force-sensor run and the final known-weight fine-tuning stage, and $\beta_f=1$ otherwise. Franka instead supervises only the vertical component with
$\mathcal{L}^{\mathrm{Franka}}_{\mathrm{force}}=|\mathcal{T}_f|^{-1}\sum_{t\in\mathcal{T}_f}w_t\operatorname{Huber}_{0.15}(\hat f_{t,z}-f^{\mathrm{gt}}_{t,z})$, where $f^{\mathrm{gt}}_{t,z}=-mg$ is the known-payload label.
The sample weight $w_t$ implements fixed nonzero-force-sample reweighting.
We set $v_{t,k}=1$, including for missing-value sentinels converted to zero; configurations that enable invalid-channel masking set the corresponding $v_{t,k}=0$.
The classification terms are
\begin{align}
\mathcal{L}_{\mathrm{gate}}
&=\frac{1}{|\mathcal{T}_f|}
\sum_{t\in\mathcal{T}_f}
\operatorname{BCE}(g_t,g_t^{\mathrm{gt}}), \\
\mathcal{L}_{\mathrm{cond}}
&=\frac{1}{|\mathcal{T}_c|}
\sum_{t\in\mathcal{T}_c}
\operatorname{BCE}(c_{t,3},m_{t,3}),
\label{eq:classification_losses}
\end{align}
where $m_{t,3}=1$ denotes normal operation of Joint~3 and $m_{t,3}=0$ denotes the mechanically restricted condition.
For selected SO-101 and Franka runs, finite-difference velocities computed from logged poses provide an auxiliary Smooth L1 rollout loss $\mathcal{L}_{\mathrm{vel}}$.
The complete objective is
\begin{equation}
\begin{aligned}
\mathcal{L}
&=\lambda_{\mathrm{joint}}\mathcal{L}_{\mathrm{joint}}
+\lambda_{\mathrm{grip}}\mathcal{L}_{\mathrm{grip}}
+\lambda_{\mathrm{force}}\mathcal{L}_{\mathrm{force}} \\
&\quad+\lambda_{\mathrm{gate}}\mathcal{L}_{\mathrm{gate}}
+\lambda_{\mathrm{cond}}\mathcal{L}_{\mathrm{cond}}
+\lambda_{\mathrm{vel}}\mathcal{L}_{\mathrm{vel}}.
\end{aligned}
\label{eq:training_objective}
\end{equation}
We use $\lambda_{\mathrm{vel}}=5$ for the velocity-supervised SO-101 runs and $5$ or $10$ across the Franka training stages; it is zero otherwise. Unavailable task losses are likewise assigned zero weight.
For visual-supervision experiments, the silhouette objective is added to Eq.~\ref{eq:training_objective} and backpropagated through the path described above.
We train with AdamW~\cite{loshchilov2019decoupled} for up to 100000 epochs; in practice the best checkpoints emerge within 30000 epochs.
Unless otherwise specified, experiments are conducted on a standard workstation with an Intel Core Ultra 7 265K processor (20 cores, 5.0 GHz), 32 GB of memory, and an NVIDIA GeForce RTX 5080 GPU with 16 GB of VRAM; the no-load benchmark in Appendix~\ref{sec:appendix_impl} uses an RTX 4090, and the hand-eye calibration measurements in Appendix~\ref{supp_sec:diff_rendering} use an A6000.
Network training and inference were implemented in JAX with JIT compilation~\cite{jax2018github} and GPU acceleration, and differentiable simulation for rigid-body dynamics was based on the MuJoCo engine and its MJX JAX implementation~\cite{todorov2012mujoco,mujoco_mjx}.
Additional implementation details are provided in the Appendix.

Because the gripper is driven by a rotary motor while its fingers translate along prismatic joints, relating motor commands to jaw position requires a kinematic mapping.
We describe this mechanism next.

\subsection{Gripper Kinematics}
\label{sec:gripper_model}

\begin{figure}[t]
  \centering
  \includegraphics[width=\linewidth]{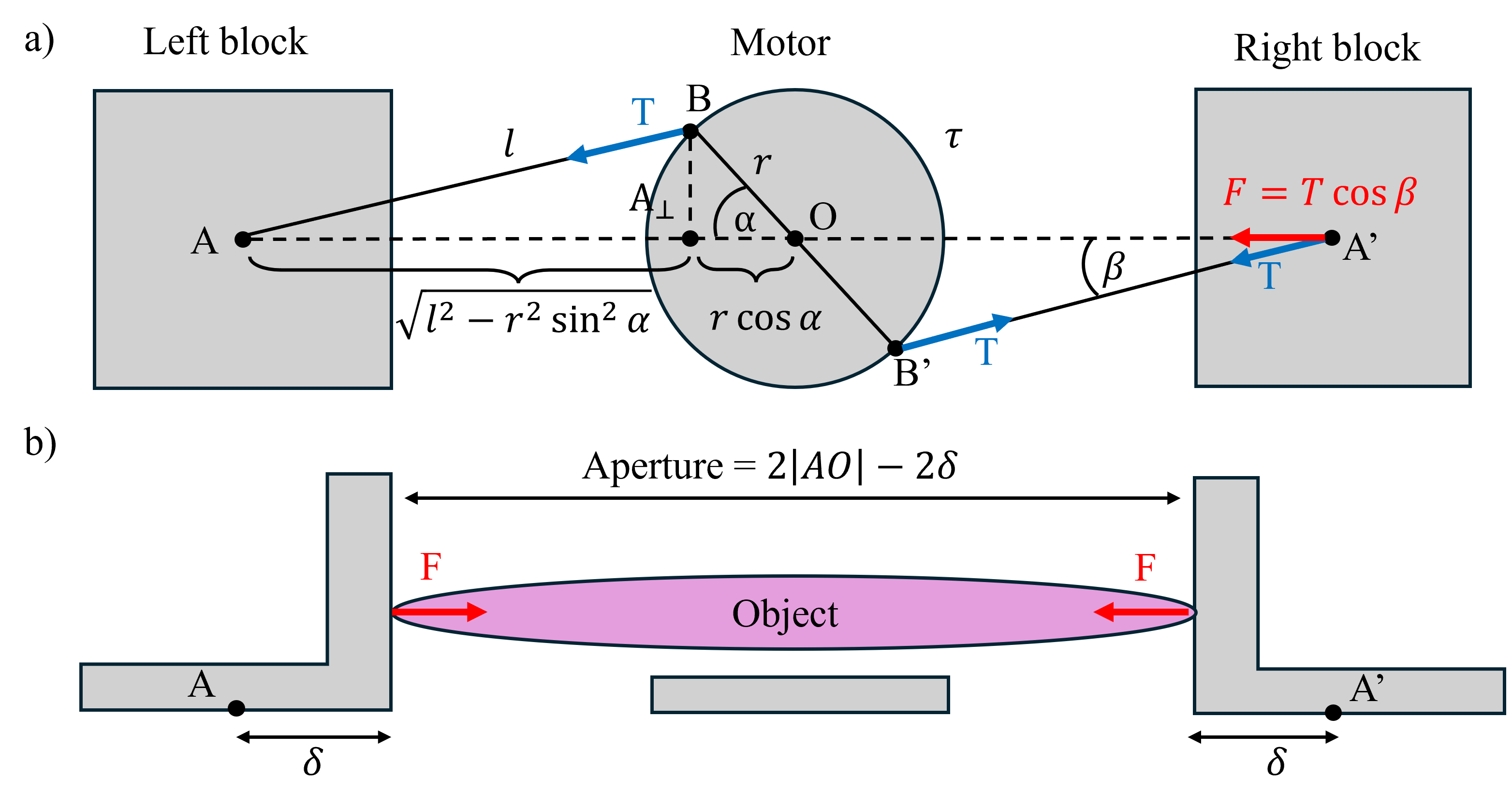}
  \caption{\textbf{Kinematics and force transmission of the symmetric gripper.}
  Top (a) and side (b) views show the rotary motor driving two symmetric prismatic jaws.
  Each link transmits an axial force $T$, whose component along the closing direction gives the per-jaw force $F=T\cos\beta$.}
  \label{fig:gripper_geometry}
\end{figure}

Fig.~\ref{fig:gripper_geometry} defines the physical gripper geometry.
Let $B$ and $B'$ denote the two crank pins, $r=|OB|=|OB'|$ the crank radius, $l=|AB|=|A'B'|$ the link length, $\alpha$ the acute angle between either crank radius and the slider axis, and $\delta$ the offset from each slider pin ($A$ or $A'$) to the corresponding inner jaw face.
The symmetric jaw aperture is
\begin{equation}
a(\alpha)
=2|AO|-2\delta
=2\!\left(\sqrt{l^2-r^2\sin^2\alpha}+r\cos\alpha\right)-2\delta.
\label{eq:gripper_aperture}
\end{equation}
Eq.~\ref{eq:gripper_aperture} provides the ideal motor-angle-to-jaw-aperture relation for the physical mechanism; the simulator represents the gripper with a calibrated single-finger prismatic coordinate.
For a quasistatic symmetric grasp, the links act as two-force members carrying axial force $T$.
If $\beta$ is the angle between a link and the slider axis, each jaw applies the closing force $F=T\cos\beta$; torque balance about the motor axis gives $\tau=2Tr\sin\gamma$, where $\gamma=\angle OBA=\angle OB'A'$.
Thus, the gripper's force transmission depends on both motor torque and mechanism configuration.

\section{Experiments}

\subsection{Rollout Accuracy}
\label{sec:rollout_acc_no_load}
We evaluate the fidelity of our differentiable simulator by comparing predicted trajectories against ground truth on a held-out test dataset. For each joint $j$ at horizon $T$, we compute the mean absolute error (MAE) across $N$ test trajectories:
$
\mathrm{MAE}_{j}(T)=\frac{1}{N}\sum_{n=1}^{N}\frac{1}{T}\sum_{t=1}^{T}\bigl|q^{(n)}_{j,t}-\hat{q}^{(n)}_{j,t}\bigr|
$,
where $q^{(n)}_{j,t}$ and $\hat{q}^{(n)}_{j,t}$ denote the ground-truth and predicted positions, respectively. Tab.~\ref{tab:simulation-accuracy-all} reports rollout errors across horizons. The model remains accurate over short horizons for both revolute and prismatic joints, and the errors remain moderate over the evaluated longer horizons. Because these experiments replay the recorded current sequence, they evaluate recorded-current-conditioned trajectory propagation rather than counterfactual planning; model-predictive control would additionally require a model of the current response to candidate commands. The consistent trend across joints indicates that physics-in-the-loop learning captures effective history-dependent input--output behavior while retaining the structure of differentiable rigid-body dynamics.
Force MAE is the componentwise mean over the evaluated trajectories, time steps, and three Cartesian output channels: $\mathrm{MAE}_{F}(T)=(3NT)^{-1}\sum_{n,t,k}|f^{(n)}_{t,k}-\hat f^{(n)}_{t,k}|$. For payload-based benchmarks, the nominal reference is $\mathbf f^{\mathrm{gt}}=[0,0,-mg]^\top$, so the metric compares lateral predictions with zero and the vertical prediction with $-mg$. The directional force-sensor benchmark provides measured three-axis force references. For gauge pushing, the scalar gauge reading is assigned to the known contact axis and the orthogonal reference components are zero; the tabulated F values retain the same three-channel convention, whereas the trace-level values in Fig.~\ref{fig:teaser} use the contact-axis force magnitude. For OpenManipulator-X, Grip denotes single-finger slide-coordinate MAE in millimeters, rather than the total jaw aperture $a(\alpha)$.

\begin{table}[t]
\centering
\footnotesize
\setlength{\tabcolsep}{12pt}
\caption{\textbf{Gradient behavior across rollout horizons $H$.} Here, input-grad denotes the norm of the loss gradient at the simulator's generalized-effort input; $\cos$ is the parameter-gradient cosine similarity relative to $H=128$.}
\vspace{-2mm}
\resizebox{0.7\columnwidth}{!}{
\hspace{-3mm}
\label{tab:horizon_gradient}
\begin{tabular}{r ccc}
    \toprule
    $H$ & input-grad & $\|\nabla_\theta \mathcal{L}\|$ & $\cos$ \\
    \midrule
    64  & $2.73\!\times\!10^{-2}$ & 8.77   & 0.96 \\
    128 & $2.59\!\times\!10^{-2}$ & 17.33  & 1.00 \\
    256 & $1.72\!\times\!10^{-2}$ & 22.35  & 0.99 \\
    320 & $1.45\!\times\!10^{-2}$ & 22.91  & 0.99 \\
    500 & $9.37\!\times\!10^{-3}$ & 20.29  & 0.98 \\
    \bottomrule
\end{tabular}
}
\vspace{-1mm}
\end{table}

Across the rollout horizons $H\!\in\!\{64,128,256,320,500\}$ (Tab.~\ref{tab:horizon_gradient}), simulator-input gradient norms remain $O(10^{-2})$, and parameter-gradient cosine similarity remains at least 0.96 relative to $H{=}128$; $\|\nabla_\theta \mathcal{L}\|$ plateaus around $H{=}256\text{--}320$, motivating the curriculum used in training.

\begin{table}[t]
  \centering
  \caption{\textbf{Simulation accuracy on the test set.}
J1--J4: joint-angle MAE (deg). Grip: single-finger slide-coordinate MAE (mm).
Rollouts of 600 steps span approximately 10 seconds at the platform sampling rate.}
  \vspace{-1mm}
  \label{tab:simulation-accuracy-all}
  \setlength{\tabcolsep}{2.5pt}
  \resizebox{\columnwidth}{!}{
  \begin{tabular}{l|ccccc|ccccc|ccccc}
    \toprule
    & \multicolumn{5}{c|}{\textbf{@100 steps}}
    & \multicolumn{5}{c|}{\textbf{@300 steps}}
    & \multicolumn{5}{c}{\textbf{@600 steps}} \\
    \textbf{Task} & J1 & J2 & J3 & J4 & Grip & J1 & J2 & J3 & J4 & Grip & J1 & J2 & J3 & J4 & Grip \\
    \midrule
    backward\_forward & 2.4 & 4.3 & 4.4 & 4.2 & 0 & 2.3 & 5.9 & 3.9 & 4.2 & 0 & 3.1 & 4.8 & 4.9 & 4.5 & 0 \\
    circular\_ccw & 1.7 & 5.2 & 2.2 & 2.6 & 0 & 1.5 & 3.1 & 2.0 & 3.4 & 0 & 2.5 & 2.3 & 2.1 & 2.8 & 0 \\
    circular\_cw & 3.6 & 3.2 & 1.4 & 2.2 & 0 & 3.1 & 1.9 & 1.2 & 2.6 & 0 & 2.6 & 1.7 & 2.0 & 2.8 & 0 \\
    go\_up\_stay\_still & 2.5 & 1.8 & 2.3 & 1.2 & 0 & 3.1 & 2.1 & 3.1 & 2.5 & 0 & 3.1 & 1.5 & 3.5 & 2.7 & 0 \\
    joint\_sweep\_1 & 2.3 & 2.5 & 1.8 & 1.4 & 0 & 1.9 & 1.0 & 2.6 & 1.8 & 0 & 2.9 & 1.9 & 2.0 & 1.6 & 0 \\
    joint\_sweep\_2 & 2.1 & 4.3 & 4.6 & 1.2 & 0 & 1.8 & 6.7 & 8.4 & 1.1 & 0 & 2.4 & 4.6 & 5.2 & 1.7 & 0 \\
    joint\_sweep\_3 & 3.7 & 4.5 & 3.1 & 1.9 & 0 & 3.4 & 2.6 & 2.8 & 2.8 & 0 & 3.1 & 2.6 & 2.5 & 3.4 & 0 \\
    joint\_sweep\_4 & 2.6 & 4.8 & 2.5 & 3.4 & 0 & 2.9 & 3.3 & 2.4 & 3.2 & 0 & 2.4 & 2.9 & 2.2 & 4.6 & 0 \\
    \midrule
    joint\_sweep\_5 & 1.7 & 3.0 & 1.7 & 3.1 & 0.4 & 2.6 & 3.9 & 2.3 & 2.7 & 0.5 & 5.5 & 3.3 & 5.1 & 3.0 & 0.7 \\
    pick\_place\_empty & 2.6 & 2.5 & 2.2 & 3.5 & 1.1 & 1.9 & 3.6 & 2.7 & 3.2 & 1.1 & 3.1 & 2.7 & 2.0 & 3.9 & 1.0 \\
    \midrule
    Average & 2.5 & 3.6 & 2.6 & 2.5 & 0.2 & 2.5 & 3.4 & 3.1 & 2.8 & 0.2 & 3.1 & 2.8 & 3.2 & 3.1 & 0.2 \\
    \bottomrule
  \end{tabular}
  }
  \vspace{-4mm}
\end{table}

\subsection{Force Estimation Accuracy}
\label{sec:rollout_acc_with_force}
We evaluate force prediction on three complementary benchmarks: (i)~directional contact measured by a calibrated six-axis force/torque sensor, (ii)~known payloads whose gravitational loads define force labels, and (iii)~contact with a force gauge across multiple directions and heights. The experimental setup is shown in Fig.~\ref{fig: force_sensor}.

\begin{table}[t]
\vspace{1mm}
  \centering

  \caption{\textbf{Simulation and force prediction accuracy on the force-sensor test set.}
J1--J4: joint-angle MAE (deg). Grip: single-finger slide-coordinate MAE (mm). F: force MAE (N).}

  \label{tab:force-sensor-prediction-accuracy}
  \setlength{\tabcolsep}{1.5pt}
  \resizebox{\columnwidth}{!}{
  \begin{tabular}{l|ccccc c|ccccc c|ccccc c}
    \toprule
    & \multicolumn{6}{c|}{\textbf{@100 steps}}
    & \multicolumn{6}{c|}{\textbf{@300 steps}}
    & \multicolumn{6}{c}{\textbf{@500 steps}} \\
    \textbf{Task} & J1 & J2 & J3 & J4 & Grip & F & J1 & J2 & J3 & J4 & Grip & F & J1 & J2 & J3 & J4 & Grip & F \\
    \midrule
    \multicolumn{19}{c}{\textit{With External Force Contact}} \\
    \midrule
    force\_X$+$ & 0.9 & 2.3 & 0.9 & 1.6 & 1.0 & 0.36 & 1.1 & 1.8 & 2.4 & 1.3 & 1.0 & 0.46 & 0.9 & 1.9 & 2.7 & 1.7 & 1.0 & 0.50 \\
    force\_X$-$ & 1.4 & 1.1 & 0.8 & 1.1 & 1.0 & 0.39 & 1.6 & 2.3 & 1.3 & 2.3 & 1.0 & 0.42 & 2.0 & 2.9 & 1.5 & 3.0 & 1.0 & 0.44 \\
    force\_Y$+$  & 2.3 & 5.7 & 2.8 & 0.6 & 0.1 & 0.36 & 2.1 & 4.5 & 2.6 & 0.9 & 0.2 & 0.38 & 2.7 & 4.8 & 2.1 & 0.9 & 0.3 & 0.44 \\
    force\_Y$-$ & 1.4 & 1.6 & 3.1 & 0.5 & 0.1 & 0.38 & 1.5 & 3.0 & 3.1 & 1.3 & 0.1 & 0.36 & 1.4 & 3.6 & 3.1 & 1.5 & 0.2 & 0.38 \\
    force\_Z$+$ & 3.3 & 6.2 & 3.3 & 0.8 & 0.1 & 0.39 & 2.3 & 5.6 & 2.5 & 1.9 & 0.2 & 0.43 & 2.5 & 4.4 & 2.2 & 2.3 & 0.3 & 0.46 \\
    force\_Z$-$ & 1.0 & 5.9 & 0.6 & 0.6 & 1.0 & 0.36 & 2.9 & 6.6 & 2.8 & 2.5 & 1.0 & 0.64 & 3.8 & 6.4 & 3.5 & 2.0 & 1.0 & 0.57 \\
    \midrule
    \multicolumn{19}{c}{\textit{Reference (No External Force)}} \\
    \midrule
    force\_X$+$ & 0.6 & 2.0 & 1.5 & 1.2 & 1.0 & 0.01 & 0.8 & 1.8 & 1.0 & 1.1 & 1.0 & 0.02 & 1.0 & 1.6 & 1.1 & 1.1 & 1.0 & 0.01 \\
    force\_X$-$ & 2.0 & 2.5 & 0.5 & 1.1 & 1.0 & 0.00 & 1.4 & 2.3 & 0.5 & 1.0 & 1.0 & 0.00 & 1.5 & 2.2 & 0.8 & 1.3 & 1.0 & 0.00 \\
    force\_Y$+$ & 2.0 & 1.0 & 2.0 & 0.6 & 0.1 & 0.00 & 2.1 & 3.3 & 2.1 & 1.1 & 0.2 & 0.00 & 1.9 & 3.7 & 1.7 & 1.1 & 0.4 & 0.00 \\
    force\_Y$-$ & 1.0 & 1.8 & 1.2 & 1.3 & 0.1 & 0.00 & 1.6 & 4.5 & 2.0 & 2.2 & 0.1 & 0.00 & 1.2 & 4.0 & 2.4 & 1.6 & 0.3 & 0.00 \\
    force\_Z$+$ & 1.2 & 1.7 & 2.1 & 0.5 & 0.1 & 0.00 & 1.6 & 3.8 & 2.2 & 1.2 & 0.2 & 0.00 & 1.5 & 2.9 & 2.0 & 1.3 & 0.3 & 0.00 \\
    force\_Z$-$ & 0.6 & 2.3 & 0.9 & 0.9 & 1.0 & 0.02 & 0.7 & 1.6 & 1.1 & 1.1 & 1.0 & 0.01 & 0.9 & 1.3 & 1.0 & 1.1 & 1.0 & 0.01 \\
    \midrule
    {Avg} & {1.48} & {2.84} & {1.64} & {0.90} & {0.55} & {0.19} & {1.64} & {3.43} & {1.97} & {1.49} & {0.58} & {0.23} & {1.78} & {3.31} & {2.01} & {1.58} & {0.65} & {0.23} \\
    \bottomrule
  \end{tabular}
  }
\end{table}

\begin{table}[t]
  \centering
 \caption{\textbf{Simulation and force prediction accuracy on the payload-based test set.} J1--J4: joint-angle MAE (deg). Grip: single-finger slide-coordinate MAE (mm). F: force MAE (N).}
  \label{tab:weight-prediction-accuracy}
  \setlength{\tabcolsep}{1.5pt}
  \newlength{\colTask}\setlength{\colTask}{0.8cm}
  \newlength{\colWeight}\setlength{\colWeight}{0.8cm}
  \resizebox{\columnwidth}{!}{
  \begin{tabular}{w{l}{\colTask} w{c}{\colWeight}|ccccc c|ccccc c|ccccc c}
    \toprule
    & & \multicolumn{6}{c|}{\textbf{@100 steps}}
    & \multicolumn{6}{c|}{\textbf{@300 steps}}
    & \multicolumn{6}{c}{\textbf{@600 steps}} \\
    \textbf{Task} & \textbf{Weight} & J1 & J2 & J3 & J4 & Grip & F & J1 & J2 & J3 & J4 & Grip & F & J1 & J2 & J3 & J4 & Grip & F \\
    \midrule
    \multirow{3}{*}{\makecell[c]{go up\\and stay}} & 200\,g & 1.3 & 5.0 & 2.6 & 2.4 & 0.1 & 0.12 & 3.4 & 3.9 & 4.2 & 2.3 & 0.1 & 0.10 & 3.1 & 3.3 & 4.3 & 2.9 & 0.1 & 0.16 \\
    & 300\,g & 2.6 & 5.3 & 5.7 & 3.7 & 0.1 & 0.20 & 1.8 & 3.4 & 5.7 & 5.4 & 0.0 & 0.17 & 3.5 & 2.5 & 3.4 & 4.6 & 0.0 & 0.20 \\
    & 400\,g & 2.2 & 4.1 & 1.4 & 5.0 & 0.1 & 0.07 & 2.9 & 2.1 & 2.1 & 6.1 & 0.1 & 0.12 & 2.0 & 1.4 & 3.0 & 7.1 & 0.2 & 0.11 \\
    \midrule
    \multirow{4}{*}{\makecell[c]{pick and\\ place}} & 200\,g & 0.7 & 3.1 & 4.3 & 3.4 & 1.1 & 0.24 & 2.7 & 3.9 & 4.0 & 4.0 & 1.1 & 0.08 & 2.9 & 5.1 & 2.9 & 3.0 & 1.0 & 0.11 \\
    & 300\,g & 0.5 & 4.5 & 3.7 & 0.9 & 0.3 & 0.00 & 1.2 & 5.1 & 4.8 & 1.5 & 0.2 & 0.00 & 2.7 & 7.8 & 4.9 & 3.2 & 0.3 & 0.09 \\
    & 400\,g & 0.8 & 2.1 & 2.8 & 2.9 & 1.1 & 0.19 & 2.8 & 1.4 & 3.4 & 3.6 & 1.1 & 0.06 & 2.4 & 2.0 & 3.8 & 3.3 & 1.0 & 0.03 \\
    & 500\,g & 2.7 & 1.8 & 1.7 & 3.3 & 1.1 & 0.05 & 4.8 & 2.8 & 1.7 & 2.3 & 1.1 & 0.02 & 4.2 & 6.3 & 2.3 & 2.3 & 0.9 & 0.04 \\
    \midrule
    \multicolumn{2}{l|}{Avg} & 1.54 & 3.70 & 3.17 & 3.09 & 0.56 & 0.12 & 2.80 & 3.23 & 3.70 & 3.60 & 0.53 & 0.08 & 2.97 & 4.06 & 3.51 & 3.77 & 0.50 & 0.11 \\
    \bottomrule
  \end{tabular}
  }
\end{table}

\noindent
\textbf{External Force Sensing.} Tab.~\ref{tab:force-sensor-prediction-accuracy} reports 3D external force prediction across six directions (\(\pm X,\pm Y,\pm Z\)), with forces up to 9\,N applied at the end-effector. On contact trajectories, the force MAE is 0.36--0.39\,N at 100 steps and 0.38--0.57\,N at 500 steps. On reference trajectories without contact, \name predicts near-zero forces (0.00--0.02\,N). Payloads at or below $\sim$50\,g ($\sim$0.5\,N) lie close to the noise floor of this low-cost platform; reliable detection in this regime remains future work.

\noindent
\textbf{Payload Benchmark.} Tab.~\ref{tab:weight-prediction-accuracy} summarizes force prediction accuracy across payloads and rollout horizons. \name achieves an average force MAE of 0.08--0.12\,N across tasks and horizons, with single-task errors at most 0.24\,N. The go-up-and-stay task exhibits relatively larger per-row errors (up to 0.20\,N) under sustained static loading.

\begin{wrapfigure}{r}{0.41\columnwidth}
\vspace{-4mm}
\centering
\hspace{-2mm}\includegraphics[width=0.41\columnwidth]{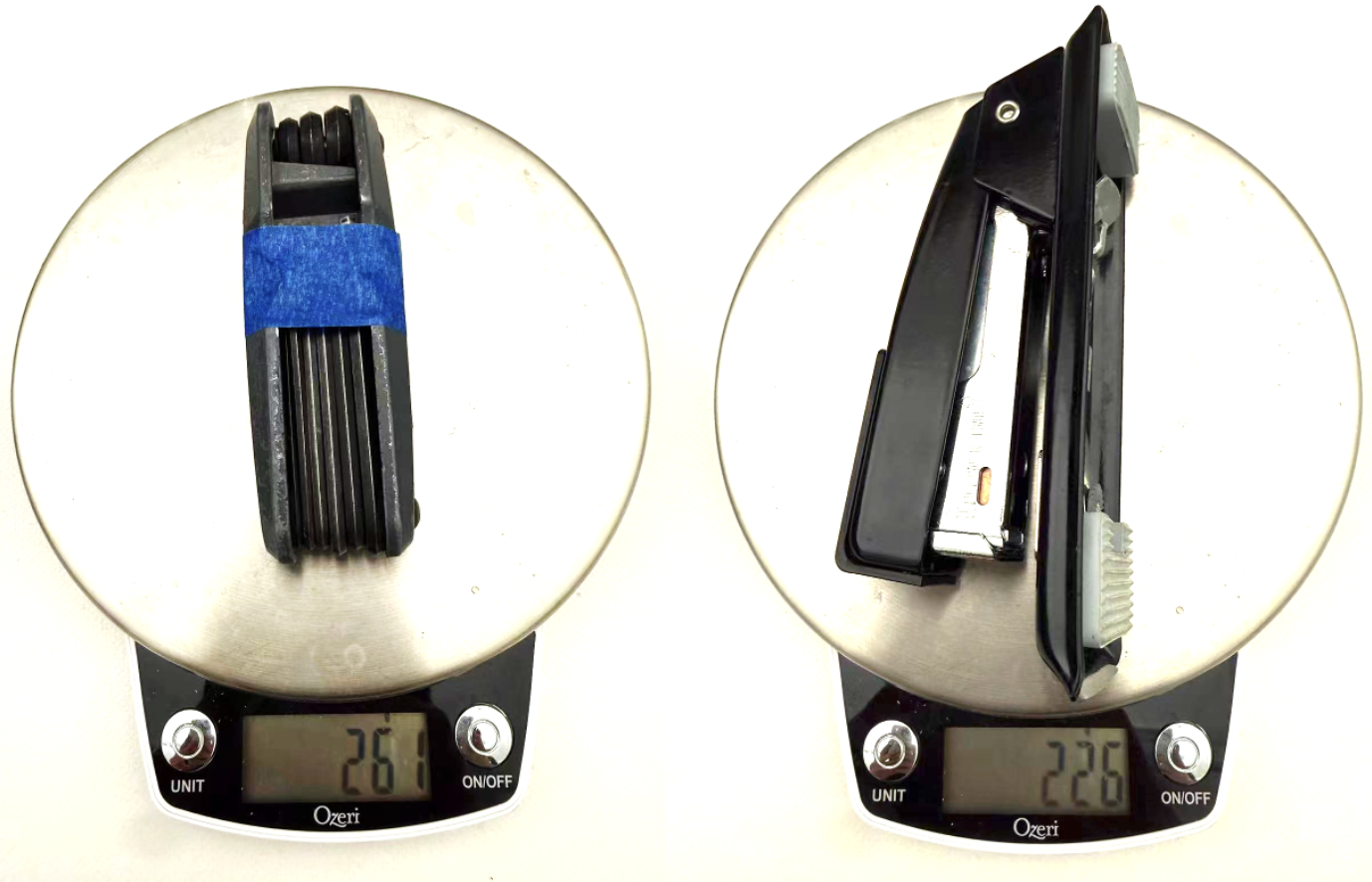}
\vspace{-4mm}
\end{wrapfigure}
\noindent\textbf{Unseen Contact Geometries.}
The force head predicts base-frame end-effector force from the shared telemetry representation and does not take explicit object geometry as input. We therefore evaluate the pretrained model on two unseen objects (inset) whose shapes and surface properties are not represented in training: a 261\,g payload and a 226\,g payload. We measure the force during the stationary holding phase.
Predicted forces are 2.80\,N (GT: 2.56\,N) and 2.40\,N (GT: 2.21\,N), indicating some generalization to unseen contact geometries in this setting.

\begin{figure*}[t]

  \centering
  \includegraphics[width=\textwidth]{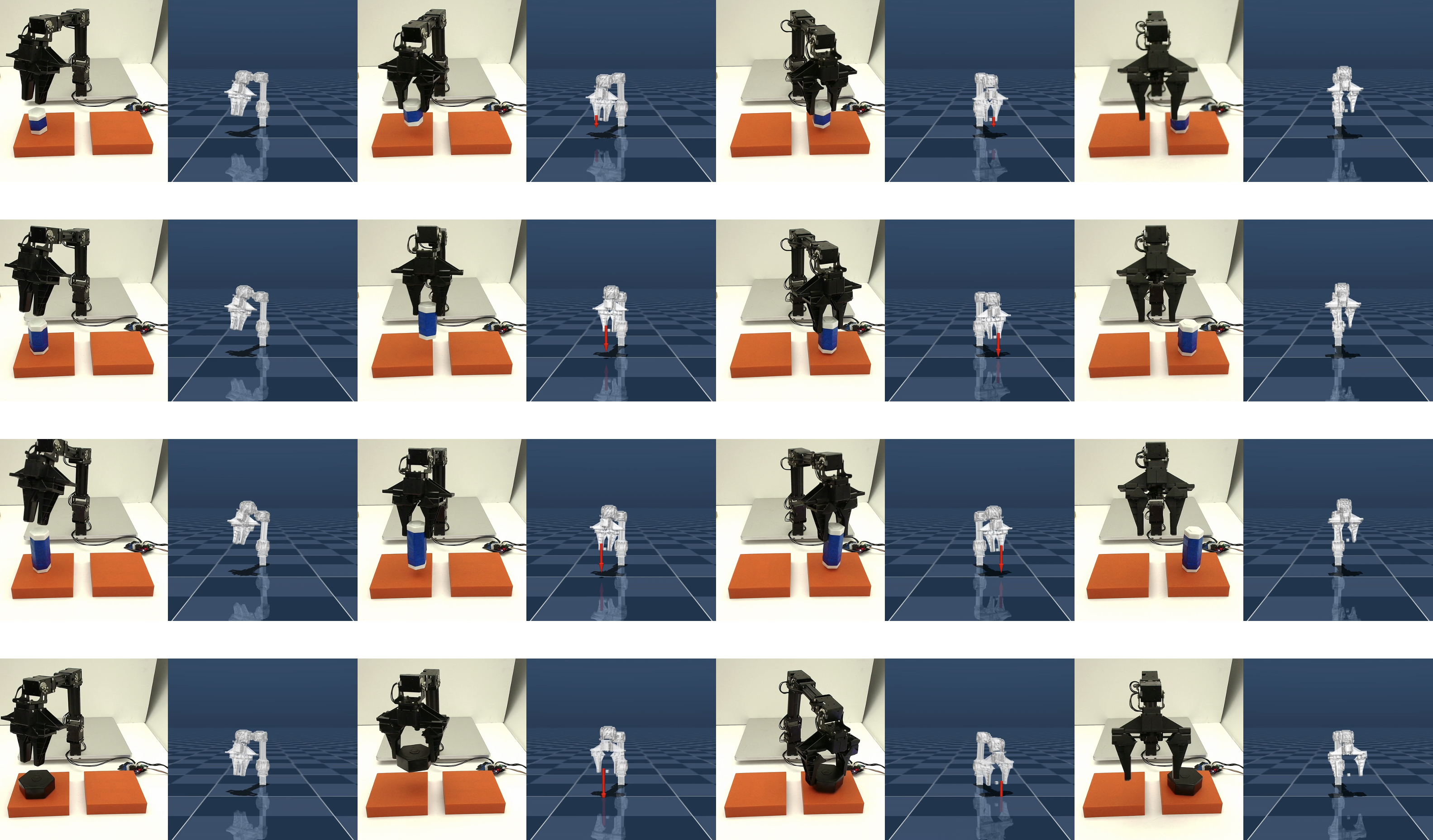}
  \caption{\textbf{Force-aware pick-and-place rollouts.} From top to bottom: 200\,g, 300\,g, 400\,g, and 500\,g payloads. Each row pairs the real execution with the simulated rollout and overlays the predicted end-effector external force.}
  \label{fig:force-pick-place-qual}
\end{figure*}

\begin{figure*}[t]
\vspace{-4mm}
  \centering
  \includegraphics[width=\textwidth]{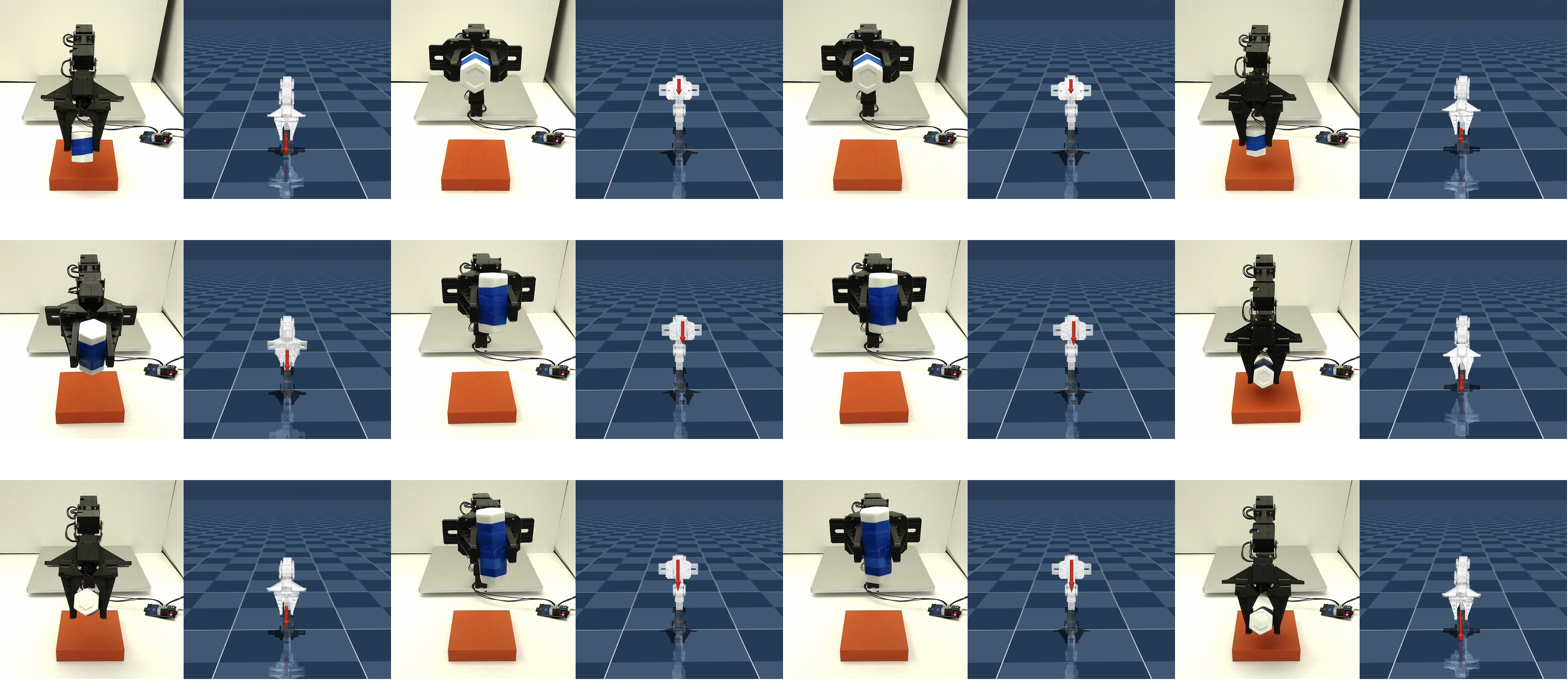}
  \caption{\textbf{Force-aware lift-and-hold rollouts.} From top to bottom: 200\,g, 300\,g, and 400\,g payloads. Each row pairs the real execution with the simulated rollout and overlays the predicted end-effector external force.}
  \label{fig:force-lift-hold-qual}
\end{figure*}

\noindent\textbf{Qualitative Payload Rollouts.}
Figs.~\ref{fig:force-pick-place-qual} and~\ref{fig:force-lift-hold-qual} show paired hardware executions and simulated rollouts for the two payload-based manipulation tasks. The force overlays capture contact transitions during pick-and-place and sustained loading during lift-and-hold, complementing the aggregate errors in Tab.~\ref{tab:weight-prediction-accuracy}.

\begin{figure*}[t]

  \centering
  \includegraphics[width=\textwidth]{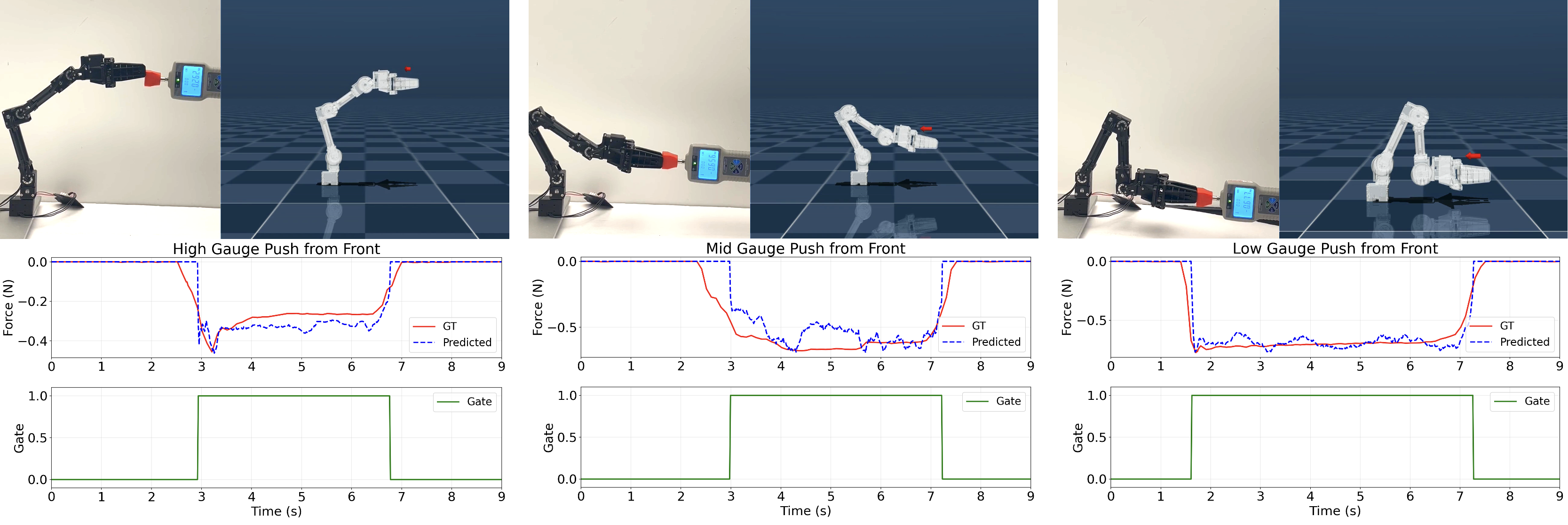}
  \vspace{-4mm}
  \caption{\textbf{Force prediction for front-direction gauge pushing.} For high, middle, and low contact positions, we show the physical setup and simulated scene together with the measured and predicted signed contact-axis force. The bottom panels show the corresponding binary contact interval.}
  \label{fig:gauge-pushing-qual}
  \vspace{-6mm}
\end{figure*}

\noindent\textbf{Gauge Pushing.}
We further test contact with a force gauge from the front (horizontal pushing) and top (vertical pressing) at low, middle, and high end-effector positions. Tab.~\ref{tab:gauge-pushing} reports state and force errors over 100-, 300-, and 600-step rollouts. Average force MAE increases from 0.08\,N at 100 steps to 0.10\,N at 600 steps and remains at most 0.12\,N for every setting. Joint errors grow moderately with horizon, whereas single-finger slide-coordinate MAE remains approximately 0.21\,mm. Fig.~\ref{fig:gauge-pushing-qual} shows that the predicted force also tracks contact onset, sustained loading, and release for representative front pushes.

\begin{table}[t]

  \centering
  \caption{\textbf{Simulation and force prediction accuracy on gauge pushing.} The end-effector contacts the gauge from the front or top at three positions. J1--J4: joint-angle MAE (deg). Grip: single-finger slide-coordinate MAE (mm). F: force MAE (N).}
  \label{tab:gauge-pushing}
  \setlength{\tabcolsep}{2pt}
  \resizebox{\columnwidth}{!}{
  \begin{tabular}{l|ccccc c|ccccc c|ccccc c}
    \toprule
    & \multicolumn{6}{c|}{\textbf{@100 steps}}
    & \multicolumn{6}{c|}{\textbf{@300 steps}}
    & \multicolumn{6}{c}{\textbf{@600 steps}} \\
    \textbf{Task} & J1 & J2 & J3 & J4 & Grip & F & J1 & J2 & J3 & J4 & Grip & F & J1 & J2 & J3 & J4 & Grip & F \\
    \midrule
    high\_push\_front & 0.91 & 1.68 & 1.18 & 0.62 & 0.00 & 0.07 & 1.62 & 1.29 & 0.53 & 0.32 & 0.00 & 0.08 & 1.48 & 1.16 & 1.44 & 1.09 & 0.00 & 0.09 \\
    high\_push\_top   & 0.73 & 1.01 & 0.78 & 1.61 & 0.01 & 0.10 & 1.38 & 1.35 & 1.97 & 1.92 & 0.01 & 0.11 & 1.62 & 1.57 & 2.72 & 2.01 & 0.01 & 0.12 \\
    mid\_push\_front  & 0.66 & 0.52 & 0.96 & 0.61 & 0.30 & 0.06 & 0.85 & 0.51 & 0.60 & 0.99 & 0.31 & 0.07 & 0.95 & 0.56 & 0.48 & 0.98 & 0.31 & 0.08 \\
    mid\_push\_top    & 0.49 & 0.54 & 0.87 & 0.74 & 0.30 & 0.09 & 0.64 & 0.43 & 0.73 & 1.07 & 0.31 & 0.10 & 0.89 & 0.52 & 0.75 & 0.78 & 0.31 & 0.11 \\
    low\_push\_front  & 0.61 & 0.75 & 1.83 & 0.23 & 0.32 & 0.08 & 0.87 & 0.56 & 1.65 & 0.33 & 0.32 & 0.09 & 1.29 & 0.88 & 1.12 & 0.98 & 0.33 & 0.10 \\
    low\_push\_top    & 1.00 & 0.36 & 1.05 & 0.29 & 0.31 & 0.08 & 1.17 & 0.84 & 2.11 & 0.74 & 0.32 & 0.09 & 1.35 & 0.76 & 2.27 & 1.04 & 0.32 & 0.10 \\
    \midrule
    Avg & 0.73 & 0.81 & 1.11 & 0.68 & 0.21 & 0.08 & 1.09 & 0.83 & 1.27 & 0.90 & 0.21 & 0.09 & 1.26 & 0.91 & 1.46 & 1.15 & 0.21 & 0.10 \\
    \bottomrule
  \end{tabular}
  }
\end{table}

\subsection{Learned Torque Surrogates}
\label{sec:current_torque}

\begin{figure}[t]
  \vspace{-2mm}
  \centering
  \includegraphics[width=0.9\linewidth]{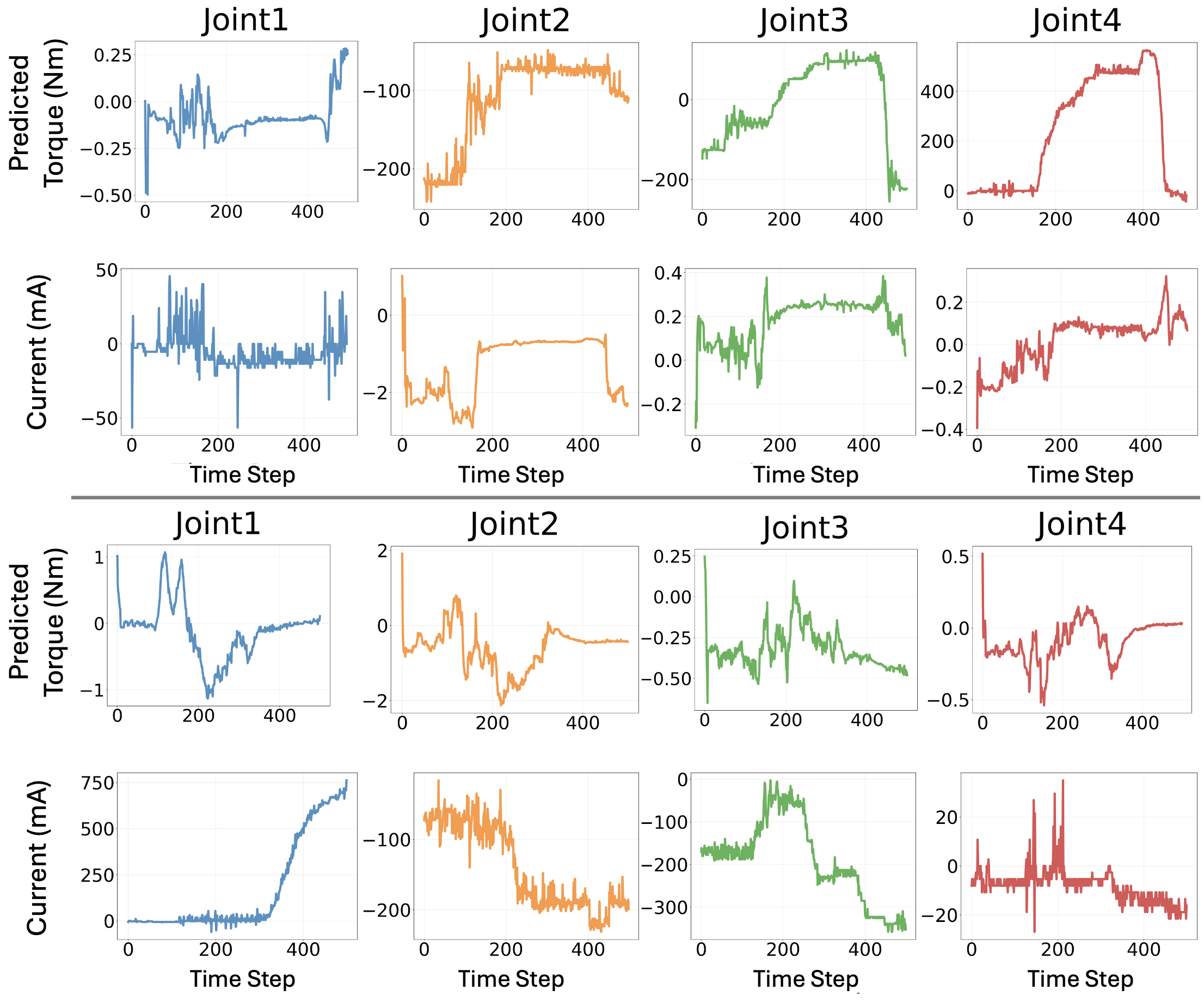}
  \vspace{-2mm}
  \caption{\textbf{Measured motor current and torque-surrogate response for the arm joints.} Two representative rollouts illustrate the learned temporal association; the responses serve as diagnostic outputs rather than direct measurements of motor torque.}
  \label{fig:current_torque_nonlinear}
  \vspace{-8mm}
\end{figure}

Fig.~\ref{fig:current_torque_nonlinear} compares torque-surrogate responses with measured motor currents over time for the four revolute joints of OpenManipulator-X. Because the platform lacks joint-torque sensing, the comparison provides a qualitative view of the nonlinear, history-dependent association learned from actuator telemetry; it does not establish a calibrated current--torque relation.

\subsection{Comparison with Baselines}
\label{sec:classical_baselines}
To compare \name with model-based alternatives under a common low-cost-servo input protocol, we implement three adapted baselines on the payload benchmark: (i)~\textit{ID-Linear}, the acceleration-based inverse-dynamics residual $\hat{\boldsymbol{\tau}}_{\text{ext}}=\boldsymbol{\tau}_{\text{ID}}(q,\dot q,\ddot q)-(\mathbf K_t\boldsymbol{i}+\mathbf b)$ with per-joint $K_{t,j}$ calibrated on free motion; (ii)~\textit{ID-Friction}, which augments (i) with Stribeck friction and backlash terms; and (iii)~the Generalized Momentum Observer (\textit{GMO})~\cite{de2005sensorless}, which avoids $\ddot q$ through momentum integration with an auto-tuned gain. The first two are simplified benchmark adaptations and differ from the camera-assisted contact-localization method of Magrini et al.~\cite{magrini2014estimation} and the measured-motor-torque-based convex friction-uncertainty estimator of Linderoth et al.~\cite{linderoth2013robotic}.
All three baselines convert motor current to torque through the linear current--torque map and recover end-effector force as $\hat{\mathbf f}=(J_v^\top)^\dagger\hat{\boldsymbol\tau}_{\mathrm{ext}}$; they consume \emph{ground-truth} robot states at every step, whereas \name predicts forces from its own simulated rollout.
First, fitting $K_t$ per joint on OpenManipulator-X reveals the limits of the linear assumption: two joints yield negative unconstrained slopes, consistent with high-ratio gearboxes masking the underlying current--torque relation.
Second, even after Stribeck and backlash compensation, the mean force MAE of the best adapted classical baseline (GMO) is $5.5$ times that of \name (0.66\,N vs.\ 0.12\,N).
In the no-contact regime, classical methods produce $33$--$59\%$ false-positive contact rates, whereas \name remains near zero.
Adding Stribeck friction over the linear baseline provides only modest improvement (1.23 versus 1.41\,N), indicating that these fixed parametric models do not capture the dominant nonlinearities of the geared servos on this platform.

\begingroup
\begin{table}[t]

  \centering
  \caption{\textbf{Force MAE (N) against model-based sensorless baselines} on the payload benchmark at a 100-step horizon (lower is better). Baselines consume ground-truth states at every step; \name uses simulated rollout states.}
  \label{tab:classical_baselines}
  \setlength{\tabcolsep}{5pt}
  \resizebox{0.95\columnwidth}{!}{
  \begin{tabular}{l|ccc|cccc|c}
    \toprule
    & \multicolumn{3}{c|}{\textbf{Go Up \& Stay}} & \multicolumn{4}{c|}{\textbf{Pick \& Place}} & \\
    \textbf{Method} & 200\,g & 300\,g & 400\,g & 200\,g & 300\,g & 400\,g & 500\,g & \textbf{Avg} \\
    \midrule
    ID-Linear    & 1.37 & 1.81 & 2.30 & 0.72 & 0.95 & 1.22 & 1.47 & 1.41 \\
    ID-Friction & 1.06 & 1.59 & 2.15 & 0.62 & 0.82 & 1.10 & 1.31 & 1.23 \\
    GMO~\cite{de2005sensorless}            & 0.58 & 0.66 & 1.23 & 0.33 & 0.47 & 0.63 & 0.75 & 0.66 \\
    \midrule
    \textbf{\name} & \textbf{0.12} & \textbf{0.20} & \textbf{0.07} & \textbf{0.24} & \textbf{0.00} & \textbf{0.19} & \textbf{0.05} & \textbf{0.12} \\
    \bottomrule
  \end{tabular}
  }
\end{table}
\endgroup

\subsection{Online Adaptation}
\label{sec:online_adaptation}
When actuator characteristics change across hardware instances or operating conditions, a pretrained model may require local recalibration. Starting from the pretrained checkpoint, we therefore fine-tune \name in situ on 12 newly collected trajectories (7,810 frames). The adaptation runs for 70 epochs and takes 356 seconds end-to-end, corresponding to approximately 5.1 seconds per epoch.

\begin{figure}[t]

  \centering
  \includegraphics[width=\linewidth]{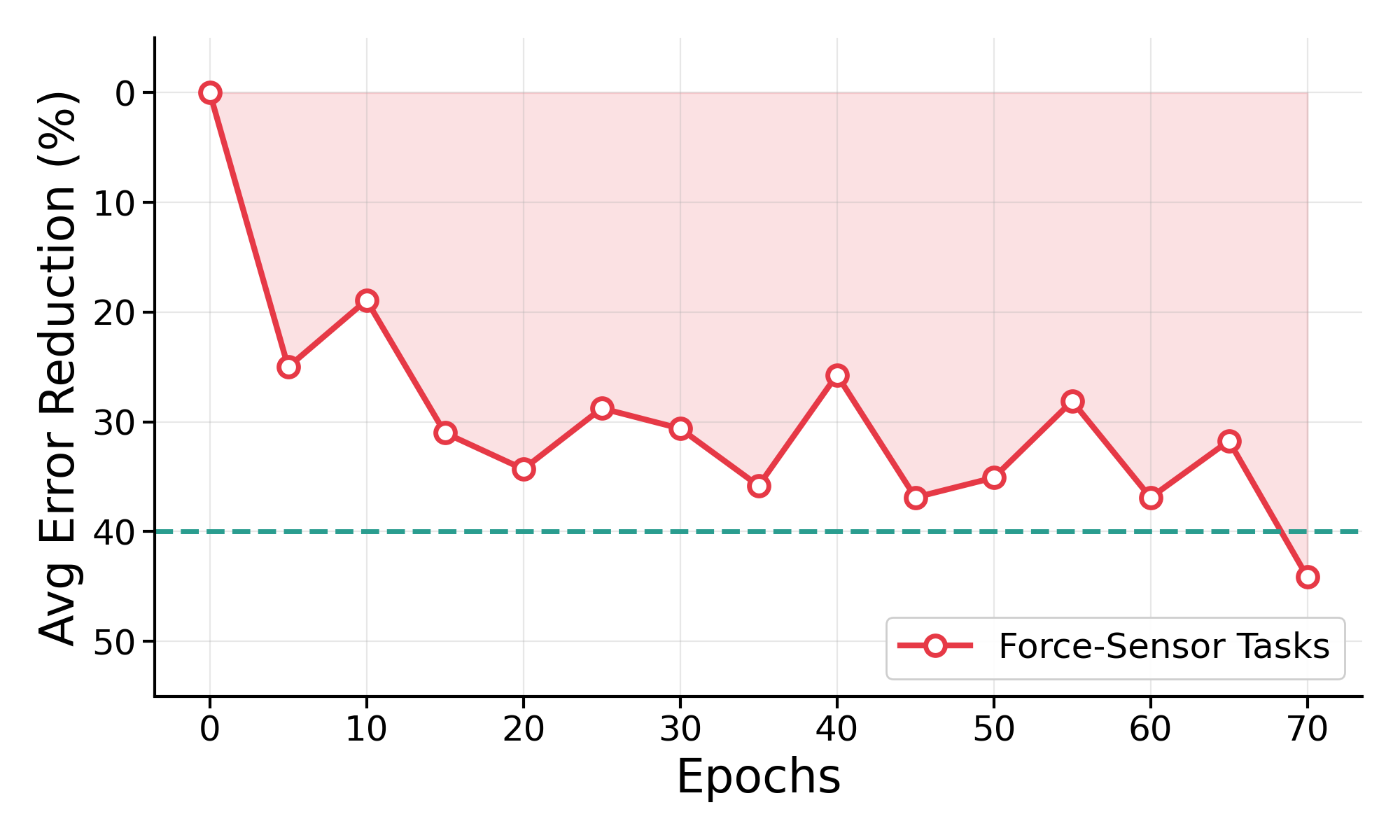}
  \vspace{-6mm}
  \caption{\textbf{Online adaptation progress.} The curve-level aggregate shows approximately 44\% average tracking-error reduction after 70 epochs with 12 trajectories (7,810 frames). Fine-tuning takes 356 seconds in total (approximately 29.7 seconds per trajectory and 5.1 seconds per epoch).}
  \label{fig:online-adaptation-progress}
  \vspace{-2mm}
\end{figure}

Tab.~\ref{tab:online-adaptation} reports per-joint position, single-finger slide-coordinate, and force errors before (B) and after (A) adaptation on the force-sensor benchmark. Adaptation reduces aggregate position-tracking error across all three rollout horizons, while force MAE remains comparable and improves slightly at the longer horizons. Fig.~\ref{fig:online-adaptation-progress} reports approximately 44\% average tracking-error reduction after 70 epochs under the curve-level aggregation used for that progress plot; this statistic is distinct from the horizon-wise Avg rows in Tab.~\ref{tab:online-adaptation}. In this setting, recalibration required 12 trajectories and 356 seconds rather than full retraining.

\begin{table*}[t]

  \centering
  \caption{\textbf{Simulation and force prediction accuracy before (B) and after (A) online adaptation.} J1--J4: joint-angle MAE (deg). Grip: single-finger slide-coordinate MAE (mm). F: force MAE (N).}
  \vspace{-1mm}
  \label{tab:online-adaptation}
  \setlength{\tabcolsep}{1.1pt}
  \resizebox{\textwidth}{!}{
  \begin{tabular}{l l| BA| BA| BA| BA| BA| BA| BA| BA| BA| BA| BA| BA| BA| BA| BA| BA| BA| BA}
    \toprule
    & & \multicolumn{12}{c|}{\textbf{@100 steps}}
      & \multicolumn{12}{c|}{\textbf{@300 steps}}
      & \multicolumn{12}{c}{\textbf{@500 steps}} \\
    & & \multicolumn{2}{c}{J1} & \multicolumn{2}{c}{J2} & \multicolumn{2}{c}{J3} & \multicolumn{2}{c}{J4} & \multicolumn{2}{c}{Grip} & \multicolumn{2}{c|}{F}
      & \multicolumn{2}{c}{J1} & \multicolumn{2}{c}{J2} & \multicolumn{2}{c}{J3} & \multicolumn{2}{c}{J4} & \multicolumn{2}{c}{Grip} & \multicolumn{2}{c|}{F}
      & \multicolumn{2}{c}{J1} & \multicolumn{2}{c}{J2} & \multicolumn{2}{c}{J3} & \multicolumn{2}{c}{J4} & \multicolumn{2}{c}{Grip} & \multicolumn{2}{c}{F} \\
    \textbf{Task} &
      & \cellcolor{colB}B & \cellcolor{colA}A & \cellcolor{colB}B & \cellcolor{colA}A & \cellcolor{colB}B & \cellcolor{colA}A & \cellcolor{colB}B & \cellcolor{colA}A & \cellcolor{colB}B & \cellcolor{colA}A & \cellcolor{colB}B & \cellcolor{colA}A
      & \cellcolor{colB}B & \cellcolor{colA}A & \cellcolor{colB}B & \cellcolor{colA}A & \cellcolor{colB}B & \cellcolor{colA}A & \cellcolor{colB}B & \cellcolor{colA}A & \cellcolor{colB}B & \cellcolor{colA}A & \cellcolor{colB}B & \cellcolor{colA}A
      & \cellcolor{colB}B & \cellcolor{colA}A & \cellcolor{colB}B & \cellcolor{colA}A & \cellcolor{colB}B & \cellcolor{colA}A & \cellcolor{colB}B & \cellcolor{colA}A & \cellcolor{colB}B & \cellcolor{colA}A & \cellcolor{colB}B & \cellcolor{colA}A \\
    \midrule
    \multicolumn{38}{c}{\textit{With External Force Contact}} \\
    \midrule
    \multicolumn{2}{l|}{force\_X$+$} & 0.9 & 0.3 & 2.1 & 0.7 & 1.0 & 0.8 & 0.8 & 0.8 & 1.0 & 1.0 & 0.36 & 0.37
               & 2.1 & 1.2 & 2.3 & 1.0 & 0.7 & 1.5 & 0.9 & 0.8 & 1.0 & 1.0 & 0.46 & 0.41
               & 2.7 & 1.6 & 4.0 & 1.3 & 0.6 & 1.9 & 0.9 & 0.8 & 1.0 & 1.0 & 0.52 & 0.41 \\
    \multicolumn{2}{l|}{force\_X$-$} & 0.2 & 0.5 & 1.1 & 1.7 & 0.9 & 1.2 & 0.7 & 1.3 & 1.0 & 1.0 & 0.39 & 0.39
               & 1.7 & 0.7 & 1.6 & 1.6 & 1.4 & 0.7 & 0.8 & 1.2 & 1.0 & 1.0 & 0.41 & 0.40
               & 2.5 & 0.9 & 1.9 & 1.6 & 2.0 & 0.8 & 0.9 & 1.2 & 1.0 & 1.0 & 0.42 & 0.41 \\
    \multicolumn{2}{l|}{force\_Y$+$} & 2.1 & 1.8 & 5.4 & 3.4 & 0.8 & 1.1 & 1.2 & 1.3 & 0.0 & 0.1 & 0.37 & 0.37
               & 2.3 & 1.5 & 4.5 & 3.2 & 1.1 & 1.1 & 1.6 & 1.0 & 0.1 & 0.1 & 0.39 & 0.37
               & 3.0 & 1.2 & 3.7 & 2.3 & 2.5 & 1.2 & 3.1 & 1.7 & 0.3 & 0.3 & 0.45 & 0.39 \\
    \multicolumn{2}{l|}{force\_Y$-$} & 1.1 & 1.2 & 0.6 & 1.3 & 1.1 & 1.8 & 1.8 & 1.8 & 0.1 & 0.1 & 0.38 & 0.38
               & 1.6 & 1.0 & 3.1 & 2.6 & 1.2 & 1.4 & 2.1 & 2.0 & 0.1 & 0.1 & 0.36 & 0.36
               & 1.7 & 1.0 & 2.3 & 2.0 & 2.5 & 1.1 & 2.1 & 1.6 & 0.2 & 0.2 & 0.38 & 0.37 \\
    \multicolumn{2}{l|}{force\_Z$+$} & 2.9 & 4.0 & 5.2 & 3.6 & 1.3 & 3.1 & 0.6 & 0.5 & 0.1 & 0.1 & 0.39 & 0.39
               & 2.7 & 2.1 & 4.0 & 2.7 & 1.4 & 1.5 & 0.7 & 0.9 & 0.2 & 0.1 & 0.44 & 0.42
               & 5.0 & 1.8 & 3.2 & 2.0 & 1.7 & 1.5 & 1.4 & 1.5 & 0.3 & 0.2 & 0.46 & 0.45 \\
    \multicolumn{2}{l|}{force\_Z$-$} & 0.7 & 0.9 & 6.2 & 3.5 & 2.0 & 0.6 & 1.0 & 0.7 & 1.0 & 1.0 & 0.36 & 0.36
               & 2.0 & 0.7 & 7.1 & 2.3 & 1.8 & 0.8 & 3.3 & 1.2 & 1.0 & 1.0 & 0.61 & 0.38
               & 2.6 & 1.0 & 7.2 & 1.6 & 2.0 & 0.8 & 3.0 & 1.1 & 1.0 & 1.0 & 0.54 & 0.40 \\
    \midrule
    \multicolumn{38}{c}{\textit{Reference (No External Force)}} \\
    \midrule
    \multicolumn{2}{l|}{force\_X$+$} & 1.4 & 0.4 & 1.6 & 0.4 & 1.0 & 1.0 & 0.8 & 0.8 & 1.0 & 1.0 & 0.01 & 0.08
               & 1.5 & 0.5 & 1.3 & 0.6 & 0.8 & 0.8 & 1.0 & 1.2 & 1.0 & 1.0 & 0.02 & 0.08
               & 1.3 & 0.5 & 1.1 & 0.6 & 0.6 & 1.0 & 1.4 & 1.9 & 1.0 & 1.0 & 0.01 & 0.05 \\
    \multicolumn{2}{l|}{force\_X$-$} & 0.4 & 0.7 & 3.2 & 0.6 & 2.4 & 0.5 & 0.7 & 1.0 & 1.0 & 1.0 & 0.00 & 0.00
               & 0.7 & 0.6 & 2.4 & 1.2 & 1.9 & 0.5 & 0.9 & 0.9 & 1.0 & 1.0 & 0.00 & 0.04
               & 0.8 & 0.6 & 2.2 & 1.0 & 2.2 & 0.7 & 1.1 & 0.8 & 1.0 & 1.0 & 0.00 & 0.05 \\
    \multicolumn{2}{l|}{force\_Y$+$} & 1.4 & 1.8 & 1.0 & 2.8 & 1.4 & 1.4 & 1.3 & 1.9 & 0.1 & 0.1 & 0.00 & 0.00
               & 2.2 & 1.7 & 3.3 & 3.3 & 1.0 & 1.2 & 1.8 & 1.4 & 0.2 & 0.2 & 0.00 & 0.01
               & 2.2 & 1.2 & 4.4 & 2.4 & 1.0 & 1.1 & 1.3 & 1.3 & 0.4 & 0.3 & 0.00 & 0.01 \\
    \multicolumn{2}{l|}{force\_Y$-$} & 2.0 & 1.2 & 2.2 & 1.9 & 0.8 & 0.4 & 2.7 & 1.3 & 0.0 & 0.1 & 0.00 & 0.01
               & 2.5 & 1.5 & 4.8 & 2.7 & 0.7 & 0.8 & 3.2 & 1.8 & 0.1 & 0.1 & 0.00 & 0.01
               & 2.0 & 1.2 & 4.9 & 2.1 & 0.7 & 1.3 & 2.7 & 2.0 & 0.3 & 0.2 & 0.00 & 0.01 \\
    \multicolumn{2}{l|}{force\_Z$+$} & 0.6 & 0.9 & 1.3 & 0.8 & 0.4 & 1.3 & 1.4 & 1.3 & 0.1 & 0.1 & 0.00 & 0.00
               & 1.3 & 1.2 & 3.4 & 1.6 & 0.4 & 1.0 & 1.2 & 1.0 & 0.2 & 0.2 & 0.00 & 0.00
               & 1.0 & 1.1 & 3.4 & 1.5 & 0.5 & 1.2 & 0.9 & 1.3 & 0.3 & 0.3 & 0.00 & 0.00 \\
    \multicolumn{2}{l|}{force\_Z$-$} & 1.8 & 0.5 & 1.9 & 0.8 & 1.1 & 0.9 & 0.5 & 0.7 & 1.0 & 1.0 & 0.03 & 0.14
               & 1.7 & 0.6 & 0.9 & 0.7 & 1.1 & 0.8 & 0.8 & 0.8 & 1.0 & 1.0 & 0.02 & 0.11
               & 1.6 & 0.5 & 0.8 & 0.7 & 1.0 & 0.7 & 0.8 & 1.1 & 1.0 & 1.0 & 0.01 & 0.08 \\
    \midrule
    \multicolumn{2}{l|}{Avg} & 1.28 & 1.19 & 2.65 & 1.81 & 1.19 & 1.17 & 1.13 & 1.12 & 0.55 & 0.55 & 0.19 & 0.21
          & 1.86 & 1.09 & 3.22 & 1.96 & 1.11 & 1.02 & 1.53 & 1.18 & 0.59 & 0.58 & 0.23 & 0.22
          & 2.20 & 1.04 & 3.25 & 1.58 & 1.44 & 1.10 & 1.64 & 1.35 & 0.66 & 0.65 & 0.23 & 0.22 \\
    \bottomrule
  \end{tabular}
  }
\end{table*}

\subsection{\texorpdfstring{Cross-Platform Validation}{Cross-Platform Validation}}
\label{sec:cross_platform}

\begin{figure}[t]
  \vspace{-1mm}
  \centering
  \includegraphics[width=\columnwidth]{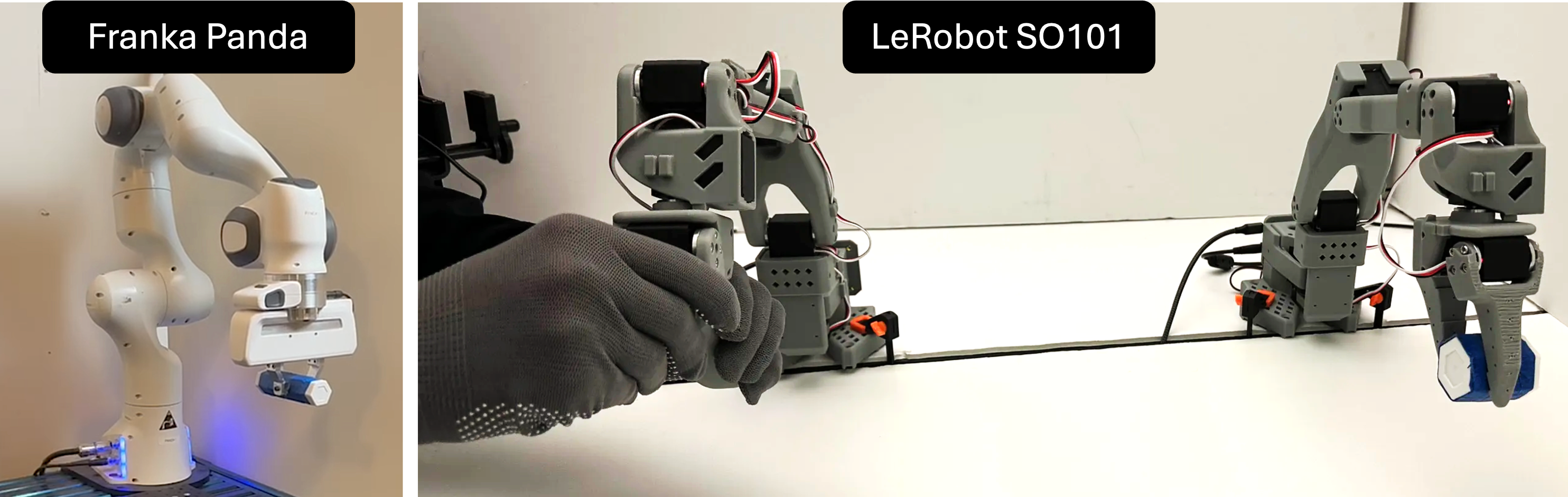}
  \vspace{-1mm}
  \caption{\textbf{Cross-platform validation.} We train \name from scratch on a 7-DoF Franka Emika Panda and a 6-DoF SO-101 low-cost arm with STS3215 servos. Franka provides an offline payload-force-estimation benchmark, whereas SO-101 provides both rollout and force evaluation. Together with OpenManipulator-X, these robots span three actuator families and costs from approximately \$500 to more than \$30{,}000.}
  \label{fig:cross_platform}
  \vspace{-6mm}
\end{figure}

For clarity, OpenManipulator-X has four arm joints plus one gripper actuator, SO-101 has five arm joints plus one gripper actuator, and Franka has seven arm joints.
The Franka results below report external force only; the SO-101 table reports its arm joints and treats gripper actuation separately.

\noindent\textbf{7-DoF Franka Emika Panda: external-force estimation.}
The Franka model is command-conditioned: its input contains a target joint pose together with the current state, commanded joint torque, motor-side state, and gripper features.
Because the logged commanded-position signal is a sparse sequence of setpoints, the offline feature construction replaces that channel with the smooth proxy
$\tilde{\mathbf q}^{\mathrm{cmd}}_t=1.03\,\mathbf q^{\mathrm{rec}}_{t+5}$, constructed from the recorded joint positions.
Because this proxy depends on a future recorded state, the Franka force results constitute a future-state-conditioned offline benchmark rather than an online evaluation.
For causal hardware deployment, this channel must instead be populated by the commanded pose already available at time $t$ from the controller, teleoperation interface, or inverse-kinematics module; no future measured state is then required.
Accordingly, the Franka analysis focuses on external-force estimation.
The force head is supervised by the known-payload vertical load and operates in parallel with the simulator-control head on the shared Transformer representation.
Across 200--600\,g lift-and-hold trajectories, the three-component force MAE ranges from 0.26 to 0.42\,N.
Tab.~\ref{tab:franka-detail} reports the payload-wise results at 100- and 500-step horizons.

\begin{table}[t]

  \centering
  \caption{\textbf{Franka Panda external-force output MAE (N)} at 100- and 500-step future-state-conditioned offline rollouts. F averages the three output components against the nominal payload reference $[0,0,-mg]^\top$. Consistent with the vertical-loading protocol, training supervision is applied to $f_z$, while the lateral outputs enter the metric through their zero references.}
  \label{tab:franka-detail}
  \setlength{\tabcolsep}{12pt}
  \begin{tabular}{lcc}
    \toprule
    \textbf{Payload} & \textbf{F @100} & \textbf{F @500} \\
    \midrule
    200\,g & 0.42 & 0.31 \\
    300\,g & 0.35 & 0.28 \\
    400\,g & 0.36 & 0.27 \\
    500\,g & 0.27 & 0.26 \\
    600\,g & 0.30 & 0.28 \\
    \midrule
    Avg & 0.34 & 0.28 \\
    \bottomrule
  \end{tabular}
\end{table}

\noindent\textbf{6-DoF low-cost arm (SO-101).}
SO-101 uses STS3215 serial-bus servos, extending the cross-platform evaluation beyond the DYNAMIXEL family used in the primary NAD experiments. Its model uses the servos' signed load registers as the effort-related input; the raw current registers are not used. The complete SO-101 corpus contains 65,916 frames (1,037.7\,s at an aggregate rate of approximately 63.5\,Hz) across ten combinations of two tasks and five payload conditions. For a controlled comparison across nonzero payloads, the evaluation subset contains 40,426 frames (649\,s, approximately 62.3\,Hz) across six combinations: go-up-and-stay and pick-and-place with 300\,g, 400\,g, and 500\,g payloads. At a 500-step horizon, force MAE is 0.47--0.73\,N, and the largest joint error is approximately $9.8^\circ$ on Joint~3 (Tab.~\ref{tab:so101-detail}). The larger joint errors relative to OpenManipulator-X may reflect differences in actuator and platform dynamics together with the finite per-platform training set; the short calibration procedure in Sec.~\ref{sec:online_adaptation} provides a practical mechanism for platform-specific refinement.

\begin{table}[t]

  \centering
  \caption{\textbf{SO-101 per-joint and force MAE} at a 500-step rollout. Joint errors are in degrees and force error (F) is in N.}
  \label{tab:so101-detail}
  \setlength{\tabcolsep}{2.5pt}
  \resizebox{\columnwidth}{!}{
  \begin{tabular}{l|ccccc c|ccccc c}
    \toprule
    & \multicolumn{6}{c|}{\textbf{Go Up \& Stay}} & \multicolumn{6}{c}{\textbf{Pick \& Place}} \\
    \textbf{Task} & J1 & J2 & J3 & J4 & J5 & F & J1 & J2 & J3 & J4 & J5 & F \\
    \midrule
    300\,g & 2.27 & 8.20 & 6.40 & 5.40 & 2.37 & 0.64 & 3.73 & 5.63 & 4.17 & 6.17 & 2.47 & 0.73 \\
    400\,g & 2.83 & 9.30 & 9.63 & 4.00 & 5.73 & 0.57 & 3.53 & 6.83 & 2.43 & 5.23 & 5.13 & 0.63 \\
    500\,g & 2.27 & 7.63 & 9.77 & 7.13 & 2.67 & 0.47 & 3.80 & 5.60 & 1.53 & 6.73 & 3.33 & 0.54 \\
    \bottomrule
  \end{tabular}
  }
\end{table}

\FloatBarrier
\subsection{Motor Condition Estimation}
\label{sec:motor_condition}

\begin{figure}[t]
\vspace{-3mm}
  \centering
  \begin{overpic}[width=0.75\linewidth]{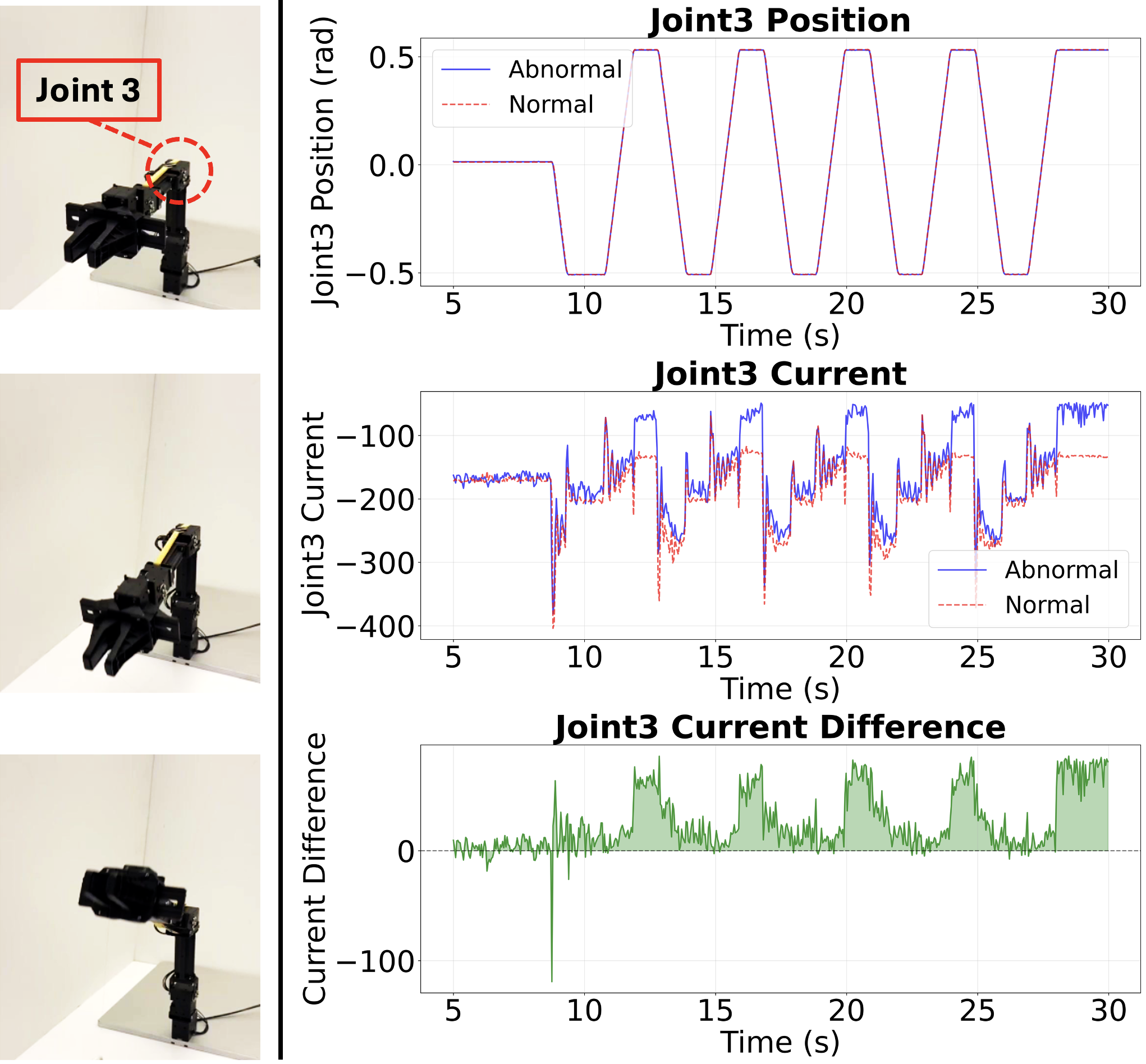}
    \put(10,-4.5){(a)}
    \put(65,-4.5){(b)}
  \end{overpic}
  \vspace{2mm}
  \caption{\textbf{Motor-condition experiment.} (a)~Rubber bands mechanically restrict Joint~3. (b)~Under matched commands, the unrestricted and restricted trials follow similar position trajectories (top), but the restricted joint draws more current (middle), producing a measurable difference (bottom).}
  \label{fig:motor_condition}
  \vspace{-2mm}
\end{figure}

Next, we create a controlled mechanical restriction by wrapping rubber bands around Joint~3 to introduce additional resistance (Fig.~\ref{fig:motor_condition}(a)). Under identical position commands, the restricted joint draws more current while following a similar trajectory (Fig.~\ref{fig:motor_condition}(b)). This current deviation enables sensorless classification of the restricted condition. For evaluation, the restricted-class score is $1-c_3$. Under the native decision granularity of each method, \name achieves 91.0\% accuracy, 96.2\% recall, and 0.95 AUC-ROC on a pick-and-place task with a 200\,g object (Tab.~\ref{tab:motor-condition-detection}), with higher values than the threshold, SVM, and Random Forest baselines. The task distinguishes unrestricted from mechanically restricted operation and does not address motor-damage diagnosis or general motor-health assessment.
\begin{table}[H]
\vspace{-2mm}
  \centering
    \setlength{\tabcolsep}{5pt}
  \caption{\textbf{Motor-condition classification under controlled mechanical restriction.} Precision and recall treat the mechanically restricted condition as the positive class. The handcrafted baselines classify overlapping 64-step windows, whereas \name uses a nine-frame context and produces framewise predictions. All methods are evaluated on the same four held-out trajectories.}
  \label{tab:motor-condition-detection}
  \vspace{-1.5mm}
\hspace{-4mm}
  \resizebox{0.7\columnwidth}{!}{
  \begin{tabular}{lcccc}
    \toprule
    \textbf{Metric} & Thres. & SVM & RF & \textbf{Ours} \\
    \midrule
    Accuracy  & 58.6\% & 59.9\% & 67.1\% & \textbf{91.0\%} \\
    Precision & 0.0\% & 52.6\% & 62.3\% & \textbf{84.5\%} \\
    Recall    & 0.0\% & 31.7\% & 52.4\% & \textbf{96.2\%} \\
    AUC-ROC   & 0.45 & 0.62 & 0.72 & \textbf{0.95} \\
    \bottomrule
  \end{tabular}
  }
\end{table}

\noindent\textbf{Baseline Protocols.}
The threshold baseline classifies a window using the mean absolute Joint~3 current, with the threshold selected on the training set.
The SVM uses an RBF kernel, and the Random Forest uses 200 trees with maximum depth 15.
Both operate on the same 20 handcrafted features: six current statistics (mean, standard deviation, maximum, minimum, skewness, and kurtosis); three tracking-error statistics (mean, maximum, and standard deviation); four temporal features (mean and maximum current derivative, mean acceleration, and current trend); three spectral features (dominant-frequency power, low-to-high-frequency power ratio, and spectral entropy); three physics-inspired features (current during motion, current-to-velocity ratio, and power estimate); and one first-half-versus-second-half current-asymmetry feature.
\name is trained on 32 trajectories (eight from each task--condition combination), whereas the three handcrafted baselines are fitted on 16 trajectories (four from each combination). All methods are evaluated on the same four held-out trajectories (two per condition).
They classify 64-step windows with 50\% overlap, yielding 601 training and 152 test windows; the SVM and Random Forest features are standardized using training-set statistics only.
\name instead produces a prediction at each time step from its native nine-frame input (eight historical frames plus the current frame) and is evaluated on the same four held-out test trajectories.
We therefore compute the metrics at each method's native decision granularity, with the same four held-out trajectories providing the common evaluation set.

\subsection{Runtime Performance}
\label{sec:runtime_performance}
Tab.~\ref{tab:neural-actuator-specs} shows that JAX with JIT compilation yields sub-millisecond GPU inference latency and throughput above the 60\,Hz control rate. Batch processing further improves throughput, and the parameter footprint remains modest.

\begin{table}[t]
  \centering
  \caption{\textbf{Model parameters and runtime performance.}}
  \vspace{-1mm}
  \label{tab:neural-actuator-specs}
    \resizebox{\columnwidth}{!}{
  \begin{tabular}{@{}lcc@{\hspace{1.5em}}lcc@{}}
    \toprule
    \multicolumn{3}{@{}c}{\textit{Model Sizes}} & \multicolumn{3}{c@{}}{\textit{Inference}} \\
    \cmidrule(r){1-3} \cmidrule(l){4-6}
    Metric & Value & Unit & Metric & Value & Unit \\
    \midrule
    Parameters & 1.44M & -- & Mean time & 0.25 & ms \\
    FLOPs (fwd) & 5.46M & -- & P95 time & 0.31 & ms \\
    \multirow{2}{*}{\makecell[c]{FP32 parameter memory}} & \multirow{2}{*}{5.50} & \multirow{2}{*}{MiB} & Thru. (b=1) & 4{,}019 & Hz \\
    & & & Thru. (b=32) & 10{,}992 & Hz \\
    \bottomrule
  \end{tabular}
  }
\end{table}

\subsection{Force-Aware Imitation Learning for Real-Robot Control}
\label{sec:imitation_learning}
To evaluate the utility of \name as a pretrained force-perception module for downstream control, we train a behavior-cloning (BC) controller using a frozen \name\ module.
Following~\cite{kobayashi2025ilbit}, we collect expert demonstrations via teleoperation and train a policy network to predict commanded joint positions (see details in Appendix~\ref{supp_sec:imitation_learning}).
We train the policy using demonstrations with 100\,g, 200\,g, 300\,g, and 500\,g payloads, and evaluate on 400\,g for object lift-and-hold and 500\,g for pick-and-place.
The pretrained \name module is trained on the full 100--500\,g payload set, while the BC demonstrations include the 500\,g condition used in the pick-and-place evaluation.
The downstream protocol therefore evaluates task-level transfer within the payload range represented during pretraining.
\textbf{Force-aware policy input.} The BC policy receives a history of joint positions and gripper aperture augmented with the \textbf{external force signal} predicted by the frozen \name module, $\hat{\mathbf{f}}_{\text{ext}}$.
This augments position-only control with force feedback under varying payloads and intermittent contacts. The \name module remains frozen throughout policy training; it is pretrained on teleoperated NAD trajectories and is not optimized jointly with the downstream policy. \textbf{Position-only baseline.} The baseline trains the same policy architecture from the position-and-aperture history without the predicted-force input.
Tab.~\ref{tab:bc-results} shows higher success rates for the force-aware variant than for the position-only baseline in the evaluated manipulation trials (Fig.~\ref{fig:teaser}).

\begin{table}[t]
  \centering
  \footnotesize
\caption{\textbf{Behavior cloning success rates with \name.} Results are averaged over 40 trials.}
  \vspace{-1mm}
\label{tab:bc-results}
  \begin{tabular}{lcc}
    \toprule
    \textbf{Task} & \textbf{w/o \name} & \textbf{w/ \name} \\
    \midrule
    Pick-and-Place & 80\% & \textbf{92.5\%} \\
    Go Up-and-Stay & 85\% & \textbf{95\%} \\
    \bottomrule
  \end{tabular}
\end{table}

\subsection{Visual Supervision}
\label{sec:diff_rendering}
\begin{figure}
  \centering
  \begin{subfigure}[b]{\linewidth}
    \includegraphics[width=\linewidth]{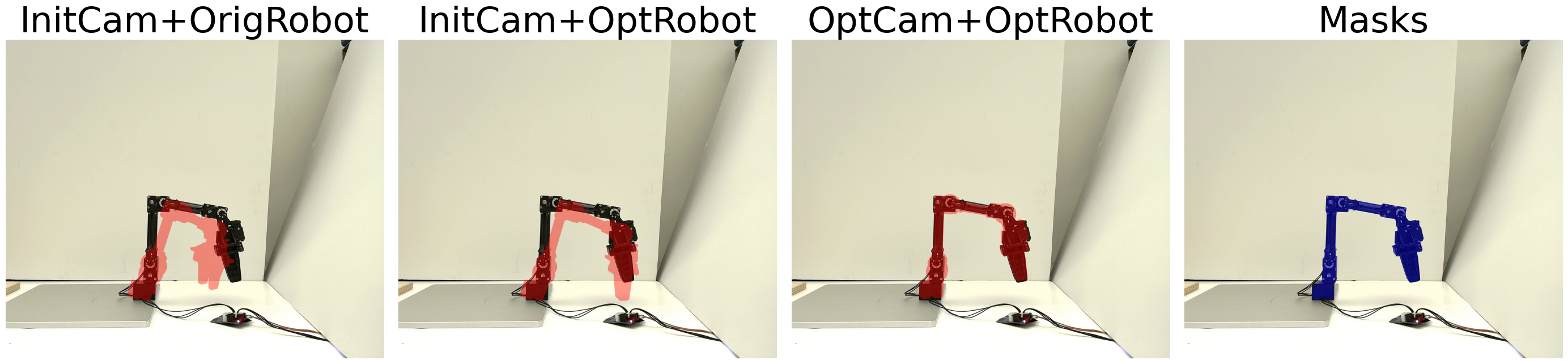}
  \end{subfigure}
  \vspace{-4pt}
  \begin{subfigure}[b]{\linewidth}
    \includegraphics[width=\linewidth]{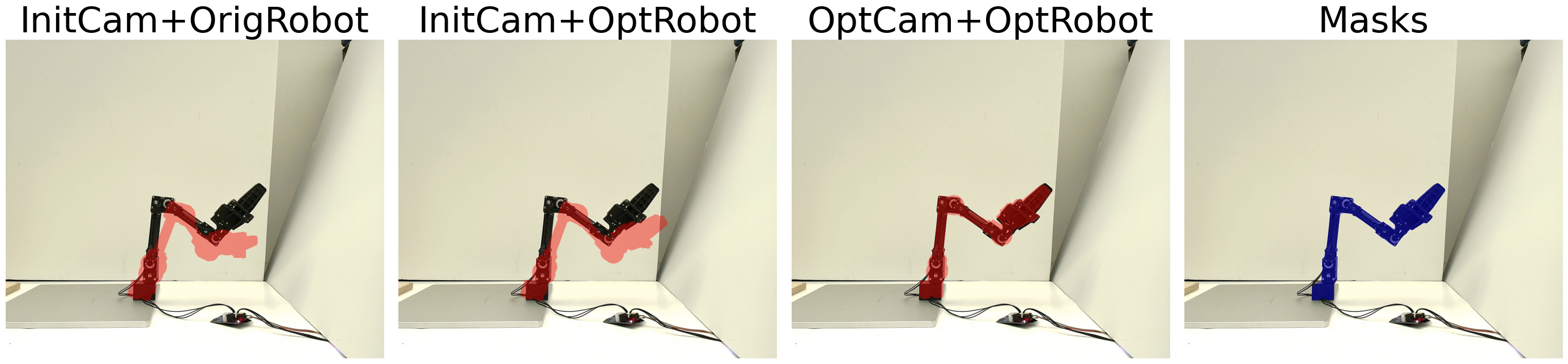}
  \end{subfigure}
  \caption{\textbf{Image-space refinement.} The evaluated pipeline
  refines the hand--eye transform and \name parameters through the
  simulated robot state; the visualization reports silhouette alignment.}
  \label{fig:visual_supervision}
\end{figure}

We evaluate an image-space refinement pipeline that composes the
neural actuator, differentiable dynamics, forward kinematics, and a
differentiable renderer. A geometric three-point initialization provides the
camera-from-base transform $T_{cb}$, after which the silhouette objective
refines the camera extrinsics and \name parameters for the evaluated
sequence. Camera intrinsics are estimated with VGGT~\cite{wang2025vggt} and held
fixed; SAM3 produces the observed robot masks~\cite{carion2025sam3segmentconcepts}.
Following the EasyHeC formulation~\cite{Chen_2023}, gradients propagate from
mask alignment through the renderer and simulated states to \name. The resulting
silhouette alignment reaches a mean IoU of 0.8515
(Fig.~\ref{fig:visual_supervision}); this experiment evaluates alignment quality
and does not by itself quantify improvements in actuator or rollout accuracy.
Appendix~\ref{supp_sec:diff_rendering} gives the complete objective and
optimization details.

\begin{table*}
  \centering
  \caption{\textbf{Torque-parameterization and force-coupling ablations.}
  Torque-parameterization results use full trajectories: the OpenManipulator-X (OMX) columns give average/worst per-joint MAE on the no-gripper tasks, whereas the SO-101 columns give worst per-joint MAE and mean force MAE over six tasks. For force coupling, the columns give the worst task--joint cell and all-task mean force MAE on the combined payload benchmarks. OMX coupling entries summarize three-seed results as mean (range); for SO-101, the implicit entry uses the released checkpoint, whereas the explicit entry summarizes three from-scratch seeds.}
  \label{tab:torque_force_ablation}
  \setlength{\tabcolsep}{5pt}
  \resizebox{\textwidth}{!}{
  \begin{tabular}{llcccc}
    \toprule
    \textbf{Design choice} & \textbf{Variant}
    & \textbf{OMX joint MAE ($^\circ$)}
    & \textbf{OMX force MAE (N)}
    & \textbf{SO-101 joint MAE ($^\circ$)}
    & \textbf{SO-101 force MAE (N)} \\
    \midrule
    Torque parameterization
      & Direct & \textbf{0.30 / 0.39} & -- & \textbf{1.52} & \textbf{0.22} \\
      & Residual & 0.49 / 0.64 & -- & 1.53 & 0.38 \\
    \midrule
    Force coupling
      & Implicit & \textbf{0.88 (0.84--0.90)} & \textbf{0.06} & \textbf{2.50} & \textbf{0.22} \\
      & Explicit & 1.33 (1.00--1.53) & 0.08 & 6.9 (6.35--7.34) & 0.33 \\
    \bottomrule
  \end{tabular}
  }
\end{table*}

\subsection{Ablation on Torque Parameterization}
\label{sec:torque_parameterization_ablation}
We compare direct prediction with the residual form in Eq.~\ref{eq:torque_parameterizations} over full trajectories. On the OpenManipulator-X no-gripper tasks, direct prediction reduces the average/worst per-joint MAE from $0.49^\circ/0.64^\circ$ to $0.30^\circ/0.39^\circ$. On the SO-101 six-task protocol, the worst per-joint MAE is nearly unchanged ($1.53^\circ$ versus $1.52^\circ$), while force MAE decreases from $0.38$ to $0.22$\,N. The residual form offers a nominal effort anchor, but its fixed linear prior appears less well matched to the nonlinear, history-dependent behavior of low-cost servos; we therefore use direct prediction (Tab.~\ref{tab:torque_force_ablation}).

\begin{figure}
  \centering
  \begin{subfigure}[t]{\linewidth}
    \centering
    \includegraphics[width=\linewidth,trim=0 14 0 0,clip]{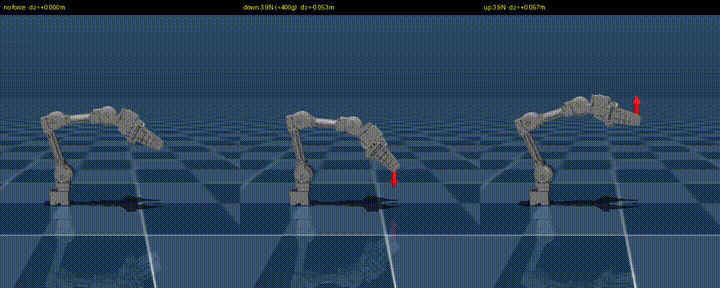}
    \caption{OpenManipulator-X}
  \end{subfigure}
  \vspace{1mm}
  \begin{subfigure}[t]{\linewidth}
    \centering
    \includegraphics[width=\linewidth]{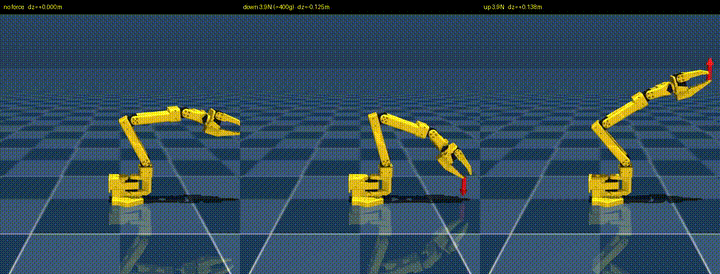}
    \caption{SO-101}
  \end{subfigure}
  \vspace{-3mm}
  \caption{\textbf{Illustration of explicit external-force coupling.}
  Predicted external forces are applied at the grasp point to illustrate their injection into the differentiable simulation. From left to right, each panel shows no applied force, a downward force, and an upward force; the red arrows indicate the injection direction. The resulting arm motion illustrates the action of the injected generalized load.}
  \label{fig:explicit_force_coupling}
\end{figure}

\subsection{Ablation on Force Coupling}
\label{sec:force_coupling_ablation}
We next compare the implicit and explicit variants on the combined payload benchmarks. Implicit coupling is more accurate on both platforms (Tab.~\ref{tab:torque_force_ablation}), with a modest difference on OpenManipulator-X and a larger gap on SO-101. Explicit coupling introduces the predicted force as an additional generalized load, as illustrated in Fig.~\ref{fig:explicit_force_coupling}. Its sensitivity to calibration mismatch in the simulator---including the force-reference frame, kinematics, and inertial parameters---may leave a load that the torque-surrogate head must counteract, particularly during sustained payload holds. In this paper, we use implicit coupling.

\subsection{Ablation on Network Architecture}
To isolate the effect of network architecture, we compare our Transformer against three alternatives with approximately matched parameter counts ($\sim1.4\,\mathrm{M}$). Let $\boldsymbol{x}_t \in \mathbb{R}^{F}$ denote the feature vector containing joint states, motor currents, and goals. We consider: (i)~an MLP with LayerNorm that predicts the torque surrogate from a flattened history, $\boldsymbol{\tau}_t = f_{\theta}(\boldsymbol{x}_{t-L+1:t})$; (ii)~a GRU~\cite{chung2014empirical}, $\boldsymbol{h}_t = \mathrm{GRU}_{\theta}(\boldsymbol{x}_t, \boldsymbol{h}_{t-1})$, with a learnable initial state; and (iii)~an LSTM~\cite{hochreiter1997long}, $(\boldsymbol{h}_t, \boldsymbol{c}_t) = \mathrm{LSTM}_{\theta}(\boldsymbol{x}_t, \boldsymbol{h}_{t-1}, \boldsymbol{c}_{t-1})$. At a 500-step horizon, the Transformer is best or tied for best on four of the six metrics, including force MAE (Tab.~\ref{tab:ablation_nn}).

\begin{table}
  \centering
  \caption{\textbf{Comparison of neural architectures for actuation modeling.} We report rollout and force prediction accuracy at a 500-step prediction horizon. Bold marks the best result per column; ties are bolded jointly.}
  \label{tab:ablation_nn}
  \vspace{-1mm}
  \setlength{\tabcolsep}{10pt}
  \resizebox{\columnwidth}{!}{
  \begin{tabular}{lccccc c}
    \toprule
    \textbf{Model} & J1 & J2 & J3 & J4 & Grip & F \\
    \midrule
    MLP  & 4.55 & 5.81 & 3.48 & 2.16 & 0.91 & 0.47 \\
     GRU & 1.83 & \textbf{2.08} & \textbf{1.68} & {1.70} & \textbf{0.65} & 0.49 \\
     LSTM & 2.91 & 7.66 & 3.21 & 3.08 & 0.71 & 0.41 \\    \midrule
    \textbf{Ours} & \textbf{1.78} & {3.31} & {2.01} & \textbf{1.58} & \textbf{0.65} & \textbf{0.23}  \\
    \bottomrule
  \end{tabular}
  }
\end{table}

\begin{figure*}
  \centering
  \includegraphics[width=\linewidth]{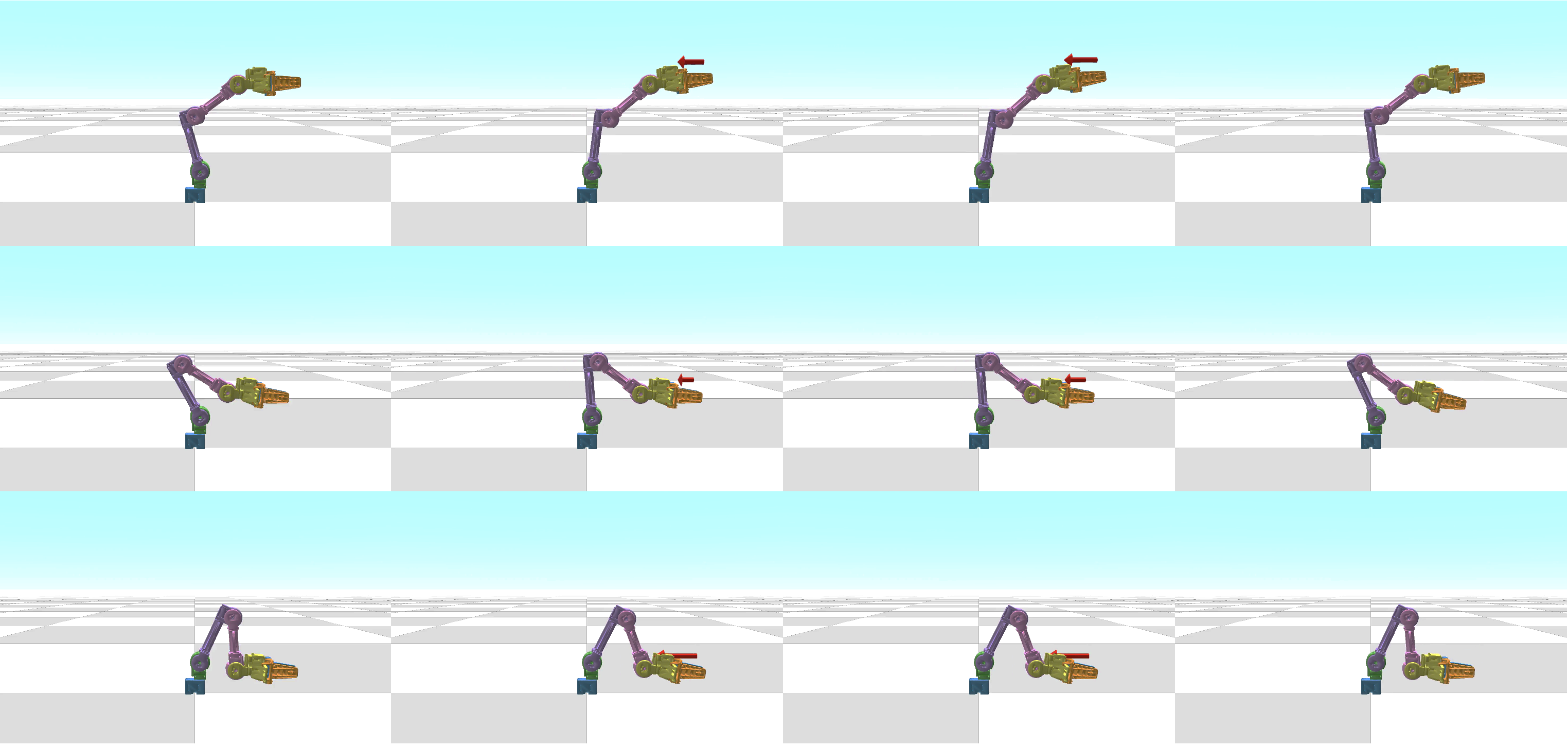}
  \vspace{-4mm}
  \caption{\textbf{\name with NVIDIA Warp.} Representative articulated-arm rollouts under different initial configurations and external push directions (red arrows), simulated with \texttt{wp.sim.FeatherstoneIntegrator}.}
  \label{fig:warp-backend}
  \vspace{-2mm}
\end{figure*}

\subsection{Additional Physics Backend}
\label{sec:warp_backend}
The learned actuator can also be coupled to differentiable rigid-body simulators other than MuJoCo. We integrate \name with NVIDIA Warp~\cite{warp2022} through a lightweight adapter that maps generalized joint states and control signals between the actuator model and Warp's articulated-body simulator. The implementation advances the robot with \texttt{wp.sim.FeatherstoneIntegrator} and records each rollout with \texttt{wp.Tape()} for reverse-mode differentiation. We use a gravitational acceleration of $-9.81\,\mathrm{m/s^2}$, an environment rate of 60\,Hz, five simulation substeps per frame (300\,Hz internally), and a per-joint armature value of 0.01. Fig.~\ref{fig:warp-backend} shows representative differentiated rollouts under multiple initial configurations and external pushes.

\section{Conclusion}
We presented \name, a differentiable neural actuator model for low-cost robots that supports torque-surrogate prediction and contact-gated force perception without dedicated force/torque sensors at inference time.
Instead of assuming a fixed linear current--torque relation, \name learns a history-dependent mapping from command, state, and telemetry histories to a torque surrogate, while jointly estimating external forces and contact.
We provide a twin-arm teleoperation pipeline for collecting force-labeled data and a multi-task Transformer that predicts torque surrogates, forces, contact probabilities, and motor-condition scores; its torque-surrogate head is trained through differentiable simulation without ground-truth joint-torque measurements.
Experiments across three platforms spanning three actuator families and costs from approximately \$500 to more than \$30{,}000 show accurate motion prediction, force estimation on payload benchmarks, motor-condition estimation, and improved behavior-cloning performance with \name.
\textit{Limitations and Future Work.} Long-horizon recorded-telemetry-conditioned trajectory propagation accumulates error and eventually degrades tracking accuracy; the few-trajectory online adaptation routine in Sec.~\ref{sec:online_adaptation} mitigates this drift.

Online inference also requires live effort-related actuator telemetry.
Although causal closed-loop deployment is supported as new telemetry arrives, the current model does not predict the effort response to unexecuted commands and therefore cannot perform counterfactual rollouts over candidate future command sequences without an additional effort-response model.
The force head estimates a single 3D resultant force at the end effector; it neither localizes distributed or multipoint contacts nor estimates a full wrench, and it requires force labels for training.
Future work could explore multipoint force prediction and more scalable ways to learn force estimation from visual cues.

\section{Acknowledgments}

This work was conducted as part of the MIT--Amazon Science Hub.
We thank Yuxiang Ma and Yuxin Song for assistance with the hardware experiments.
We thank the anonymous reviewers for their constructive feedback.

\bibliographystyle{plainnat}
\bibliography{references}
\clearpage
\appendix
\renewcommand\thefigure{\Alph{section}\arabic{figure}}
\renewcommand\thetable{\Alph{section}\arabic{table}}

This appendix provides dataset details
(Sec.~\ref{sec:appendix_dataset}), implementation details
(Sec.~\ref{sec:appendix_impl}), the visual-supervision formulation
(Sec.~\ref{supp_sec:diff_rendering}), an architecture comparison
(Sec.~\ref{supp_sec:arch_comp}), a force-sensing taxonomy
(Sec.~\ref{supp_sec:force_sensing_taxonomy}), and the imitation-learning setup
(Sec.~\ref{supp_sec:imitation_learning}).

\subsection{Additional Details of the Neural Actuation Dataset (\dbname)}
\label{sec:appendix_dataset}

\paragraph{Task Design}
We collect the Neural Actuation Dataset (\dbname) with a twin-arm leader--follower teleoperation system to capture actuator behavior across a broad range of motions, loads, and contact wrenches.
Tab.~\ref{tab:task_composition} summarizes the task suite, which is organized into three categories.
\textbf{(i) Free motion} tasks contain no external contact and are used to characterize nominal actuation dynamics under representative workspace motions, including end-effector circular trajectories (clockwise/counterclockwise), per-channel sweeps (actuator channels~1--5), and scripted motion primitives (lean-back/extend-forward, pick-and-place without payload, and go-up/hold-still).
\textbf{(ii) Force-labeled} tasks introduce controlled external wrenches in two ways: (a) \emph{payload} variants of go-up/hold-still and pick-and-place with task-specific payloads spanning 100--500\,g to induce repeatable gravity loading, and (b) \emph{directional interaction} trials in $\pm X$, $\pm Y$, and $\pm Z$ using an external fixture instrumented with a six-axis F/T sensor.
For each directional trial, we additionally record a matched \emph{no-interaction} counterpart that executes the same commanded motion but without physical contact, enabling paired characterization of contact-induced effects.
\textbf{(iii) Motor-condition} tasks replicate pick-and-place sequences under
nominal and mechanically restricted operation.
We randomly split the 10 trajectories for each task instance into training, validation, and test sets in an 8:1:1 ratio.

\begin{figure}[ht]
  \centering
  \begin{overpic}[width=\linewidth]{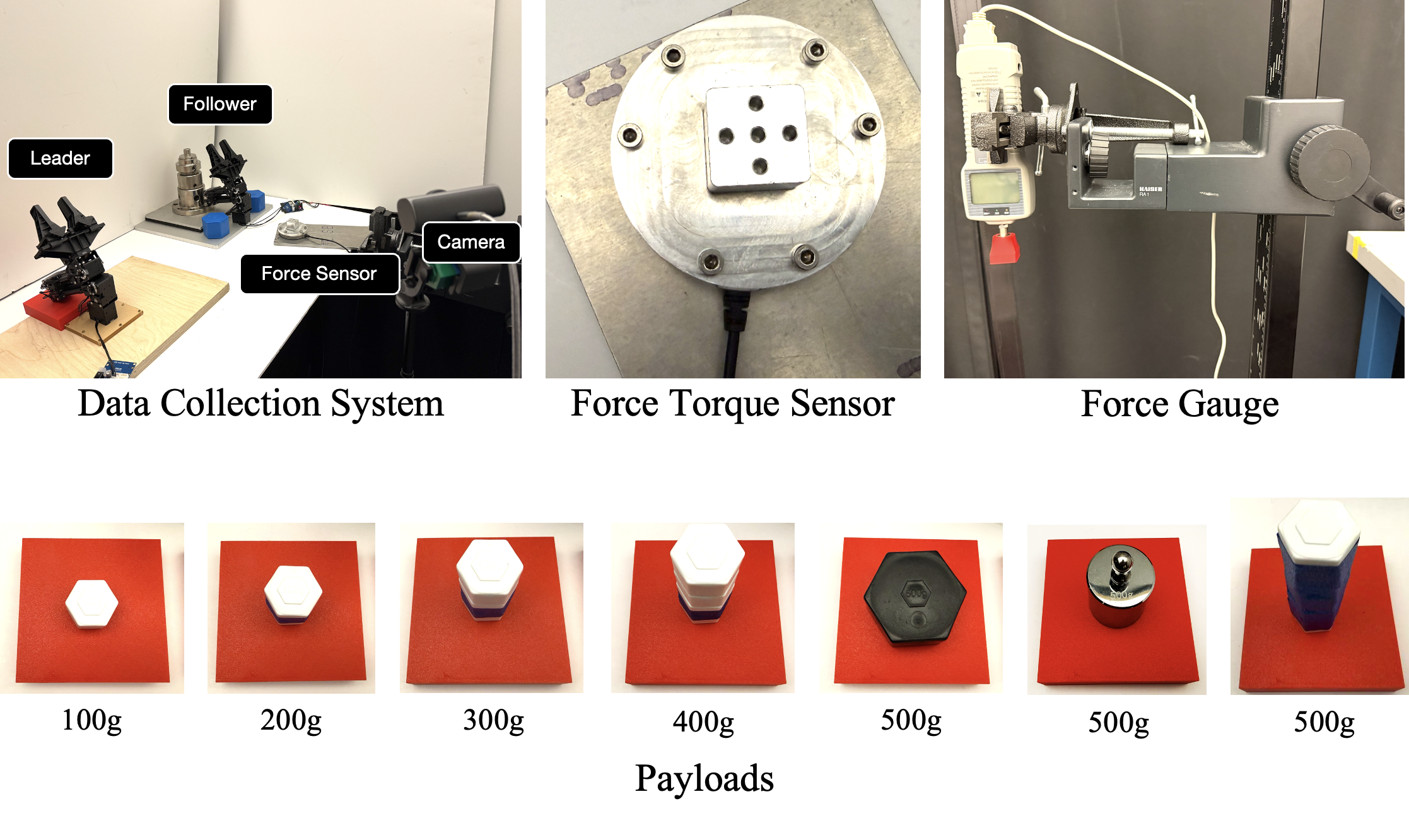}
    \put(18,25.5){(a)}
    \put(50,25.5){(b)}
    \put(82,25.5){(c)}
    \put(48,-1){(d)}
  \end{overpic}
\vspace{-4mm}
\caption{\textbf{Data collection hardware.}
(a) Twin-arm leader--follower teleoperation setup with a fixed external RGB
camera and a fixture-mounted six-axis force/torque (F/T) sensor.
(b) Close-up of the F/T sensor mounted on the rigid fixture. (c) Force gauge with a stand. (d) Payload set (100\,g, 200\,g, 300\,g, 400\,g, 500\,g) used to induce controlled loading during data collection.}
\label{fig:force_sensor_example}
\end{figure}

\begin{figure}
  \centering
  \begin{subfigure}[b]{\linewidth}
    \includegraphics[width=\linewidth]{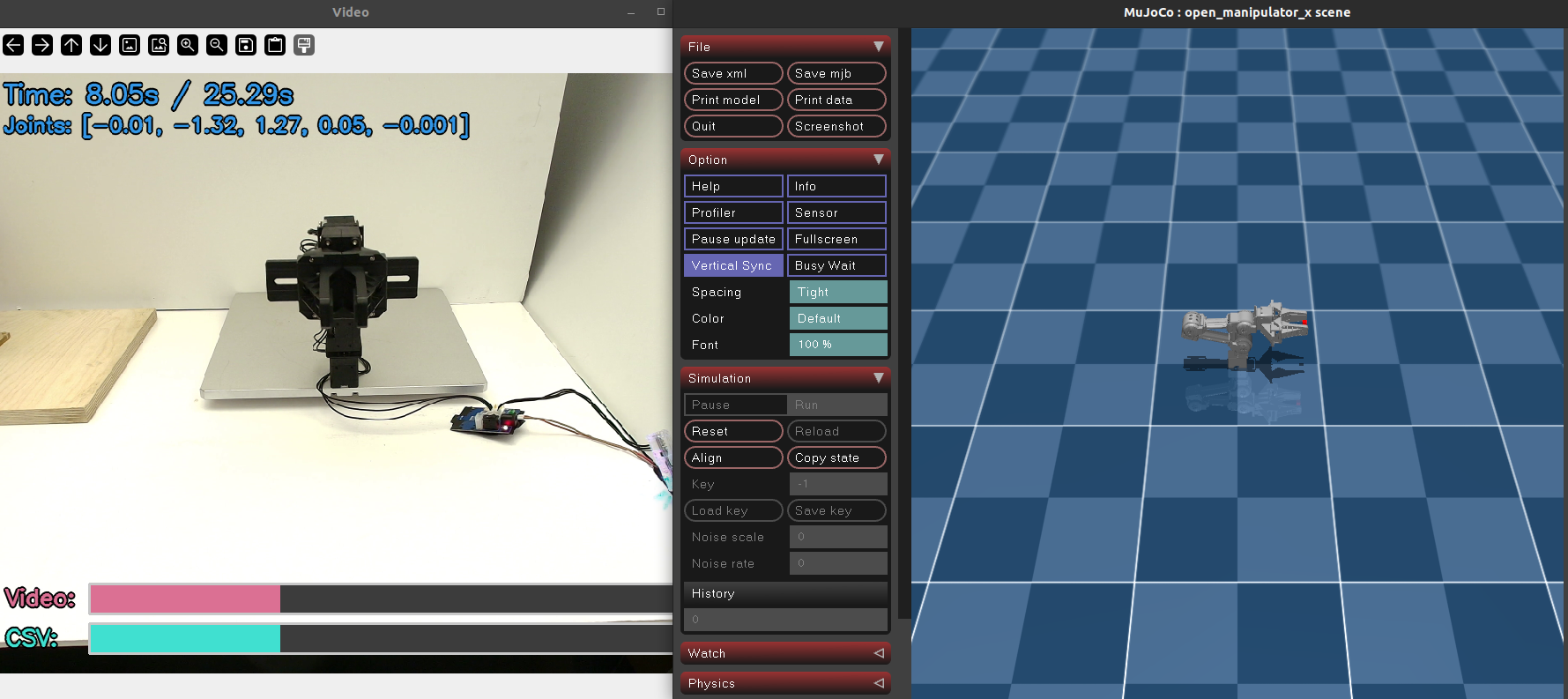}
    \caption{Free motion}
    \label{fig: free motion}
  \end{subfigure}

  \begin{subfigure}[b]{\linewidth}
    \includegraphics[width=\linewidth]{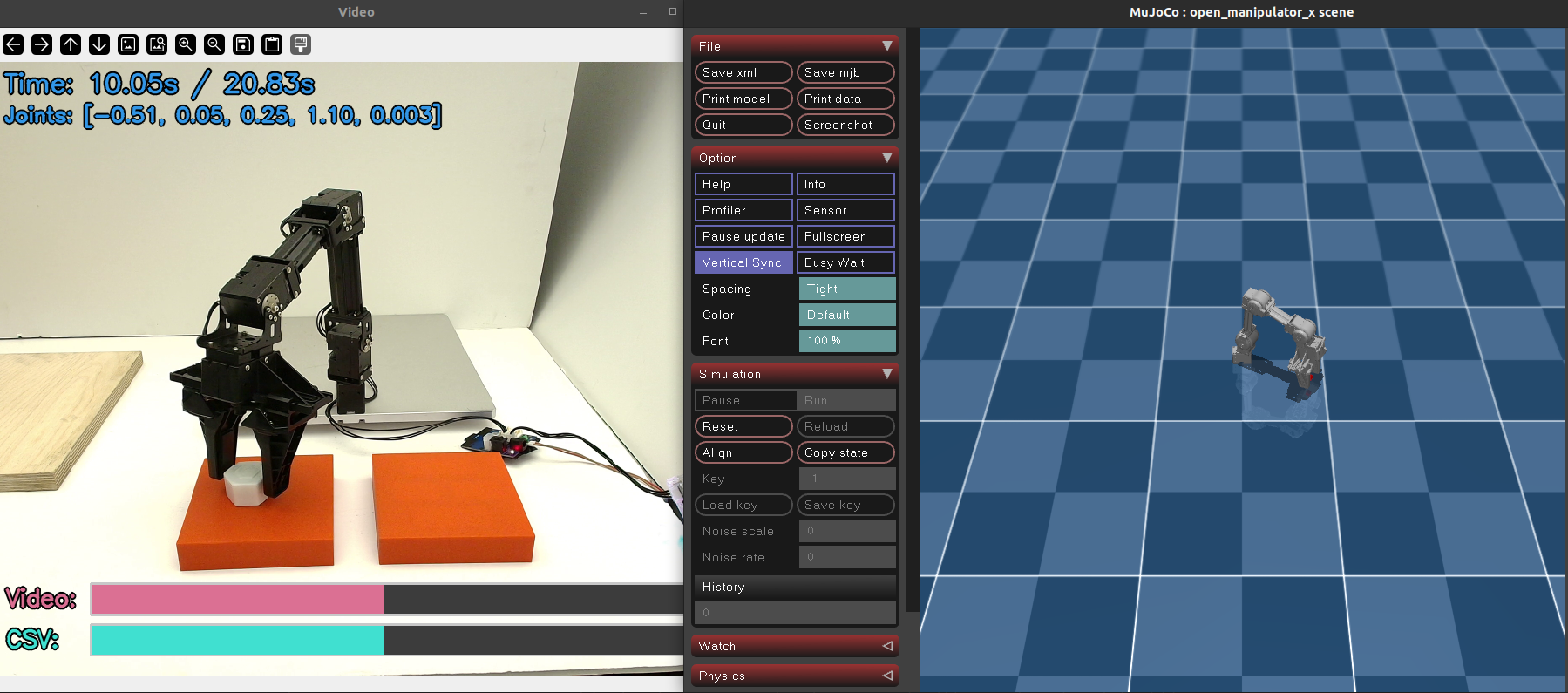}
    \caption{Payload-labeled}
    \label{fig: weight labeled}
  \end{subfigure}

  \begin{subfigure}[b]{\linewidth}
    \includegraphics[width=\linewidth]{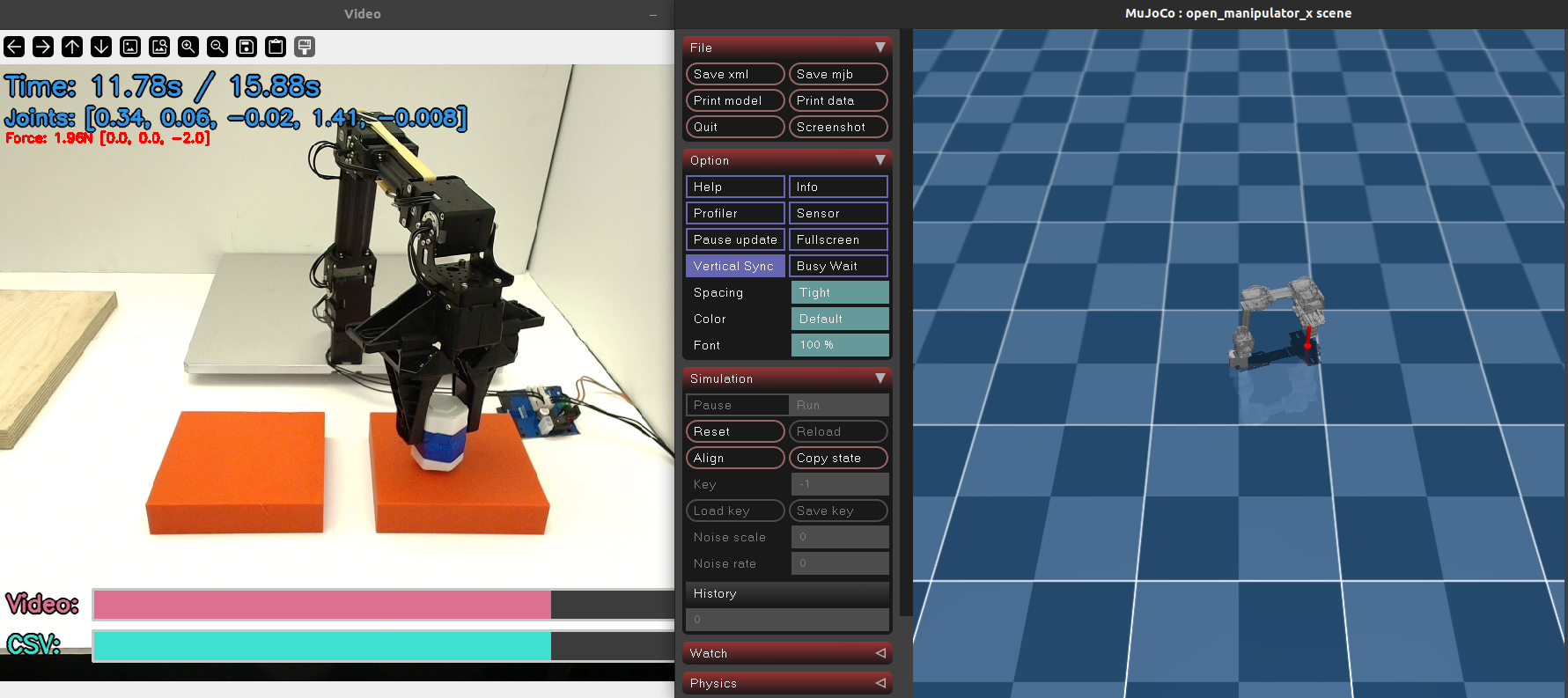}
    \caption{Mechanically restricted}
    \label{fig: motor condition}
  \end{subfigure}

  \begin{subfigure}[b]{\linewidth}
    \includegraphics[width=\linewidth]{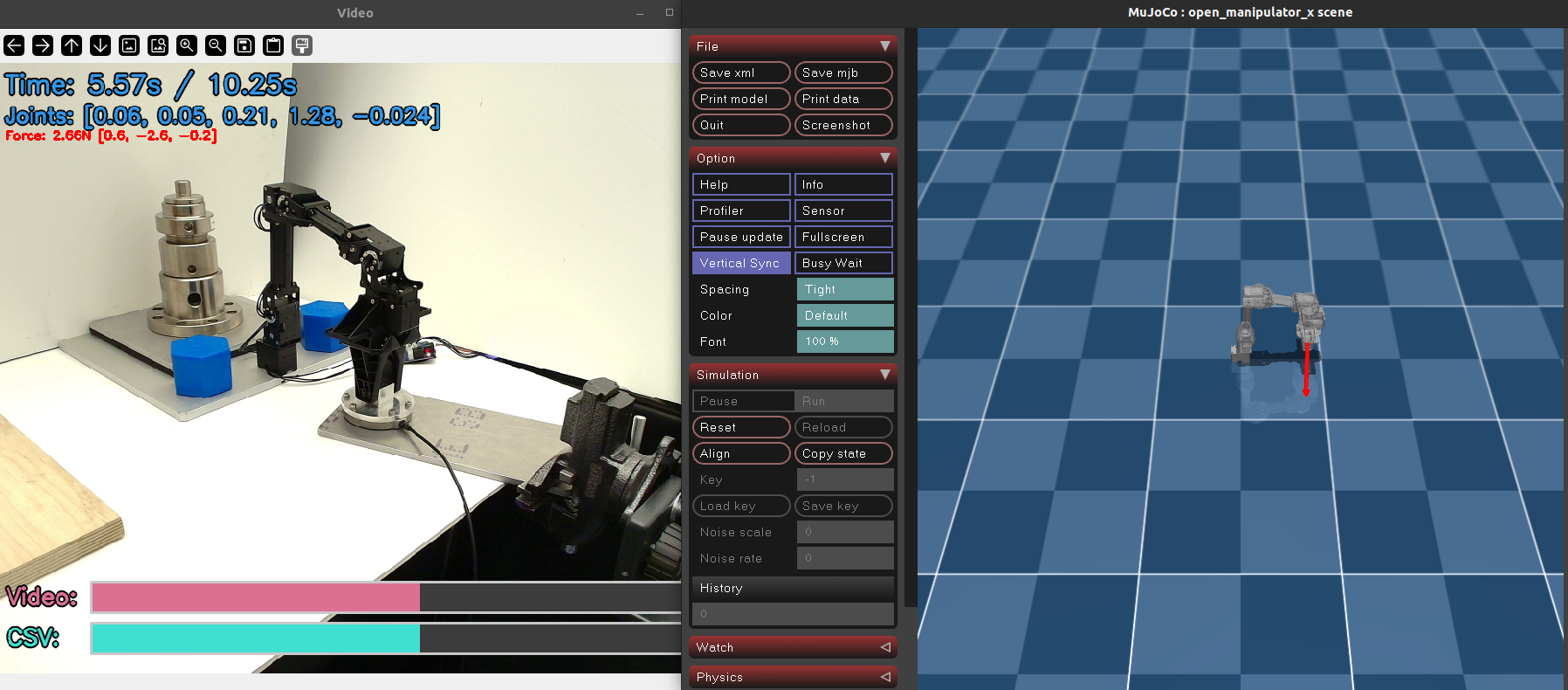}
    \caption{Force-sensor interaction}
    \label{fig: force sensor}
  \end{subfigure}

  \caption{\textbf{Representative data-collection configurations.}
  Free motion, payload supervision, mechanically restricted operation, and
  fixture-based force sensing are shown from top to bottom.}
  \label{fig: data collection}
\end{figure}

\paragraph{Data Capture Setup}
Fig.~\ref{fig:force_sensor_example} illustrates the collection apparatus.
We perform teleoperation in a leader--follower manner: a human operator drives the leader arm, while the follower arm tracks the leader's motion to execute the corresponding interaction in the physical scene (Fig.~\ref{fig:force_sensor_example}(a)).
To obtain force labels, the target object is mounted on a rigid fixture
instrumented with a calibrated six-axis F/T sensor
(Fig.~\ref{fig:force_sensor_example}(b)), which measures the interaction wrench
$(F_x,F_y,F_z,\tau_x,\tau_y,\tau_z)$. We use only the three translational force
channels as learning targets and express them in the robot base frame. For
payload-based trials, we additionally attach standardized weights
(100--500\,g; Fig.~\ref{fig:force_sensor_example}(d)) to induce repeatable gravity
loading.
A fixed external RGB camera observes the workspace and records synchronized visual observations.
All streams (robot proprioception, F/T readings, and camera frames) are timestamped and time-aligned per trajectory.
For reproducibility, we (i) zero the F/T sensor prior to each recording session and after any re-mounting, (ii) keep the camera pose and the sensor--object fixture unchanged throughout a session, and (iii) use a consistent coordinate convention across trials, defining the $\pm X/\pm Y/\pm Z$ interaction directions in the F/T sensor frame.

\begin{wrapfigure}{r}{18mm}
\vspace{-4mm}
  \hspace*{-2mm}
  \centerline{
  \includegraphics[width=33mm]{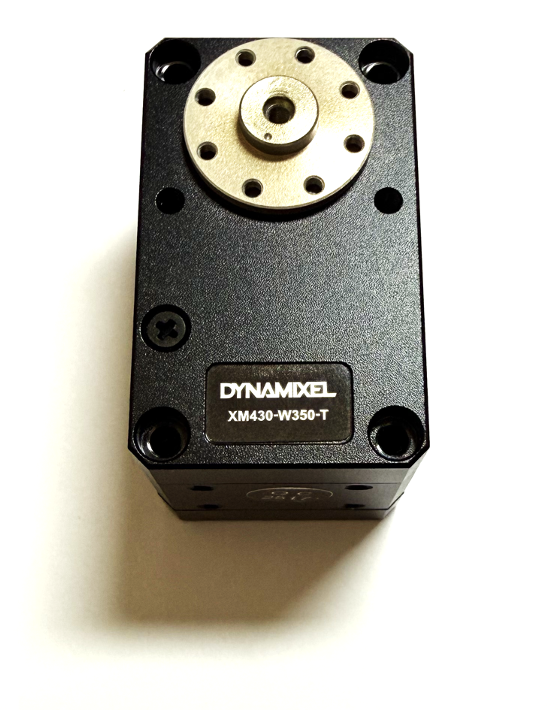}}
  \vspace*{-9mm}
\end{wrapfigure}
\paragraph{Robotic Arm}
We use two OpenManipulator-X arms with identical kinematic structures to form the leader and follower.
Each arm is a 5-DoF platform (4 revolute joints + a 1-DoF parallel gripper) driven by DYNAMIXEL XM430-W350-T motors\footnote{\url{https://emanual.robotis.com/docs/en/dxl/x/xm430-w350/}}. The gear ratio of this motor is $353.5:1$.
The system is powered at 12\,V and communicates over a TTL-level multidrop bus; the arms are controlled from a PC via an OpenCR interface. Each arm has a reach of 380\,mm, a payload capacity of 500\,g, repeatability below 0.2\,mm, a maximum joint speed of 46\,RPM, and a gripper stroke of 20--75\,mm.
Using two identical arms reduces kinematic mismatch between demonstration and execution and improves the repeatability of the collected trajectories.

\newcolumntype{C}[1]{>{\centering\arraybackslash}m{#1}}

\begin{table*}
\centering
\small
\renewcommand{\arraystretch}{1.25}
\setlength{\tabcolsep}{6pt}

\caption{\textbf{Task Composition of the Dataset.}
The dataset consists of free-motion, force-labeled, and motor-condition data.
Each task instance contains 10 trajectories, which are randomly split into 8 training, 1 validation, and 1 test trajectory. The two nominal motor-condition rows reuse trajectories listed in the free-motion and force-labeled categories.}
\label{tab:task_composition}

\begin{tabular}{
C{3.0cm} |
C{3.2cm} | C{3.5cm} |
C{2.0cm} | C{2.0cm}
}
\toprule
\textbf{Category}
& \multicolumn{2}{c|}{\textbf{Task}}
& \textbf{\#Frames}
& \textbf{Duration (s)} \\
\midrule

\multirow{10}{*}{\makecell[c]{Free motion\\(No external force)}}
& \multirow{2}{*}{Circular trajectory}
& Clockwise & 8615 & 147.06\\
& & Counterclockwise & 8428 & 143.90\\
\cline{2-5}

& \multirow{5}{*}{Joint sweep}
& Motor 1 & 22935 & 392.04\\
& & Motor 2 & 8688 & 148.47\\
& & Motor 3 & 11357 & 193.95 \\
& & Motor 4 & 15012 & 256.57\\
& & Motor 5 & 7338 & 125.21\\
\cline{2-5}

& \multicolumn{2}{c|}{Lean back and extend forward} & 12261 & 209.35\\
\cline{2-5}
& \multicolumn{2}{c|}{Pick \& place (empty)} & 15288 & 261.08\\
\cline{2-5}
& \multicolumn{2}{c|}{Go up and stay still} & 10011 & 171.14\\
\midrule

\multirow{21}{*}{Force-labeled}
& \multirow{4}{*}{\makecell[c]{Go up and stay still\\(with weight)}}
& 100\,g & 10976 & 187.61 \\
& & 200\,g & 10606 & 181.26 \\
& & 300\,g & 11259 & 192.40\\
& & 400\,g & 11291 & 192.89\\
\cline{2-5}

& \multirow{5}{*}{\makecell[c]{Pick and place\\(with weight)}}
& 100\,g & 12483 & 213.35 \\
& & 200\,g & 13245 & 226.15 \\
& & 300\,g & 13129 & 224.31 \\
& & 400\,g & 15124 & 258.54 \\
& & 500\,g & 13957 & 238.49\\
\cline{2-5}

& \multirow{12}{*}{Force sensor}
& $+X$ & 5005 & 85.17 \\
& & $+X$ w/o interaction & 5005 & 85.21 \\
& & $-X$ & 5169 & 87.94 \\
& & $-X$ w/o interaction & 5169 & 88.00 \\
& & $+Y$ & 7444 & 126.74 \\
& & $+Y$ w/o interaction & 7444 & 126.78 \\
& & $-Y$ & 6921 & 117.88 \\
& & $-Y$ w/o interaction & 6921 & 117.91 \\
& & $+Z$ & 6906 & 117.59 \\
& & $+Z$ w/o interaction & 6906 & 117.63 \\
& & $-Z$ & 5565 & 94.69 \\
& & $-Z$ w/o interaction & 5565 & 94.73 \\
\midrule

\multirow[c]{4}{*}{Motor condition}
& \multirow[c]{2}{*}{\makecell[c]{Pick \& place\\w/ weight (200\,g)}}
& Motor restricted
& 10113 & 168.26 \\
& & Nominal motor & 13245 & 226.15 \\
\cline{2-5}

& \multirow[c]{2}{*}{\makecell[c]{Pick \& place\\(empty)}}
& Motor restricted & 11252 & 192.41 \\
& & Nominal motor & 15288 & 261.08 \\
\bottomrule

\end{tabular}
\end{table*}

\noindent \textbf{Payload-labeled manipulation (SO-101).}
Tab.~\ref{tab:so101_dataset_overview} summarizes the SO-101 collection. Following the same protocol as the OpenManipulator-X setup described in the main paper, we collect teleoperated trajectories under known payload conditions ($m \in \{0, 200, 300, 400, 500\}\,\text{g}$) with two tasks: \emph{go up and stay still} and \emph{pick and place}. For each of the 10 task--payload combinations, we record 10 trajectories: eight for training, one for validation, and one for testing. The LeRobot logs contain $65{,}916$ frames over $1{,}037.7\,\text{s}$, corresponding to an aggregate rate of approximately $63.5\,\text{Hz}$.

\begin{table}
\centering
\setlength{\extrarowheight}{0pt}
\resizebox{\columnwidth}{!}{
\begin{tabular}{
>{\centering\arraybackslash}p{2.5cm} |
>{\centering\arraybackslash}p{2.2cm} | >{\centering\arraybackslash}p{2.2cm} |
>{\centering\arraybackslash}p{1.8cm} | >{\centering\arraybackslash}p{2.2cm}
}
\toprule
\textbf{Category}
& \multicolumn{2}{c|}{\textbf{Task}}
& \textbf{\#Frames}
& \textbf{Duration (s)} \\
\midrule
\multirow{10}{*}{\makecell[c]{Payload-labeled}}
& \multirow{5}{*}{\makecell[c]{Go up and\\stay still}}
  & Empty & 6753 & 87.5 \\
& & 200\,g  & 7312 & 117.4 \\
& & 300\,g  & 7227 & 116.1 \\
& & 400\,g  & 7659 & 123.0 \\
& & 500\,g  & 7630 & 122.5 \\
\cline{2-5}
& \multirow{5}{*}{\makecell[c]{Pick and\\place}}
  & Empty & 6125 & 98.4 \\
& & 200\,g  & 5300 & 85.1 \\
& & 300\,g  & 5553 & 89.2 \\
& & 400\,g  & 5725 & 92.0 \\
& & 500\,g  & 6632 & 106.5 \\
\bottomrule
\end{tabular}
}
\caption{\textbf{SO-101 dataset overview.}}
\label{tab:so101_dataset_overview}
\end{table}

The task breakdown in Tab.~\ref{tab:task_composition} contains 350 OpenManipulator-X trajectory assignments across 35 rows. The two nominal motor-condition rows reuse 20 trajectories already listed under free motion and force-labeled interaction, yielding 330 distinct OpenManipulator-X trajectories. Together with 100 trajectories spanning ten SO-101 task--payload combinations, the complete \dbname contains 430 distinct trajectories. Individual training and evaluation protocols use task-specific subsets of this broader collection, and collection videos are available for all 430 trajectories.

\subsection{Implementation Details}
\label{sec:appendix_impl}

\paragraph{Loss Functions}
We use robust regression losses for continuous targets and binary
cross-entropy for classification targets. Using implementation-specific weights, Eq.~\ref{eq:training_objective} can be written as
\begin{equation}
\begin{aligned}
\mathcal{L}
&=w_{\text{pos}}\mathcal{L}_{\text{pos}}
+w_{\text{force}}\mathcal{L}_{\text{force}}
+w_{\text{gate}}\mathcal{L}_{\text{gate}} \\
&\quad+w_{\text{cond}}\mathcal{L}_{\text{cond}}
+w_{\text{vel}}\mathcal{L}_{\text{vel}}.
\end{aligned}
\end{equation}
For the primary OpenManipulator-X configuration, $\mathcal{L}_{\text{pos}}=\mathcal{L}_{\text{arm}}+
w_{\text{grip}}\mathcal{L}_{\text{grip}}$ combines Smooth L1 losses on arm
joint positions (radians) and the single-finger slide coordinate (millimeters).
$\mathcal{L}_{\text{force}}$ is a Huber loss on the predicted base-frame 3D
force. It uses $\beta=0.15$ for the force-sensor run and the final known-weight fine-tuning stage; $\beta=1$ otherwise. Franka uses a scalar Huber loss with $\beta=0.15$ on $f_z$ alone, supervised by the known-payload label $-mg$. Nonzero-force samples receive
$w_{\text{focal}}$ times the weight of zero-force samples. For the final
known-weight fine-tuning stage and the SO-101 runs, the force loss is masked per channel so that
only valid labels contribute; other runs supervise all three channels after missing-value sentinels are converted to zero. $\mathcal{L}_{\text{gate}}$ is a binary cross-entropy loss for contact gating. For the OpenManipulator-X motor-condition task, $\mathcal{L}_{\text{cond}}$ is binary cross-entropy on the Joint~3 component of the per-motor output. The condition term is used
only for motor-condition tasks, where $w_{\text{cond}}=100$; otherwise
$w_{\text{cond}}=0$. Selected SO-101 and Franka runs use finite-difference joint velocities for auxiliary supervision, with $w_{\text{vel}}=5$ on SO-101 and $5$ or $10$ across the Franka training stages; it is zero otherwise.
The OpenManipulator-X weights are $w_{\text{pos}}=100$,
$w_{\text{grip}}=0.02$,
$w_{\text{force}}=30$, $w_{\text{gate}}=1$, and $w_{\text{focal}}=5$
($3$ for the force-sensor run). SO-101 instead uses one equally weighted six-channel pose loss over its five arm joints and rotary jaw. Franka uses seven arm-joint weights $[1,1,1,1,1.5,2,1.5]$ and replays the recorded position-controlled gripper, so the gripper does not contribute to learning the simulator-control surrogate.
Each training step uses the features at time $t$ to advance the simulated state toward $q_{t+1}$; force, gate, and condition labels are aligned at $t+1$. The offline force evaluators instead compare the output computed from the time-$t$ features with the force label time-stamped at $t$, resulting in a one-sample difference in label alignment between training and evaluation. The condition label is constant within each motor-condition trajectory.

\paragraph{Differentiable Simulation}
Training uses task-specific truncated rollouts of up to 320 steps.
For the primary OpenManipulator-X and SO-101 320-step runs, we begin with 128-step supervision and increase the horizon to 256 and then 320. The no-load runs instead use 128, 192, and 256 steps, motor-condition training remains at 128 steps, and the Franka runs use a fixed 128-step horizon. The simulator uses
4 substeps for OpenManipulator-X, 5 for SO-101, and 8 for Franka. For the torque-actuated OpenManipulator-X and SO-101 simulators, we use
direct torque-surrogate prediction, $\boldsymbol{\tau}=g_{\theta}(\mathbf{X}_t)$. Under implicit coupling, the torque surrogate alone advances the simulator. Secs.~\ref{sec:torque_parameterization_ablation} and~\ref{sec:force_coupling_ablation} evaluate the residual and explicit alternatives, respectively.
The Franka simulator instead retains the stock affine PD position actuators, and the dynamics head supplies their control channels; its raw output is therefore treated as a simulator-control surrogate rather than a direct torque estimate in $\mathrm{N\,m}$.

\paragraph{Optimization Schedule} The learning rate follows a cosine-annealing schedule from a base rate of $10^{-4}$ ($5{\times}10^{-5}$ on SO-101) whose decay horizon is set independently of the epoch budget, so extended runs continue at a small floor rate instead of restarting the decay. The known-weight configuration uses staged warm-starts for the no-gripper and force-refinement stages at $5{\times}10^{-5}$ and then $5{\times}10^{-6}$; the SO-101 extended configuration is likewise fine-tuned from its scratch run. The three Franka stages use $4{\times}10^{-5}$, $1.2{\times}10^{-5}$, and $10^{-5}$, respectively. The training code maintains both raw and exponential-moving-average parameter tracks.

\paragraph{Franka Feature Configuration}
The offline Franka force-estimation experiment uses a 52-D input per frame: a seven-joint target pose; current arm positions and velocities; normalized gripper width and goal width; seven commanded torques; motor-side positions and velocities; and seven arm tracking errors plus one gripper error.
The recorded commanded-position signal contains only a sparse sequence of stepwise setpoints, so the offline feature construction substitutes the smooth commanded-pose proxy $\tilde{\mathbf q}^{\mathrm{cmd}}_t=1.03\,\mathbf q^{\mathrm{rec}}_{t+5}$ for the target-position channel.
The recorded position-controlled gripper is replayed in simulation.
Because this offline proxy is constructed from a future recorded position, the Franka force results constitute a future-state-conditioned offline benchmark rather than an online evaluation.
In causal deployment, the commanded-pose channel must instead receive the command available at time $t$ from the external controller, teleoperation interface, or inverse-kinematics module.

\paragraph{Data and Runtime} For the no-load free-motion benchmark, we train on 80 trajectories (8 per task $\times$ 10 tasks), split across the no-gripper (8-task) and with-gripper (2-task) runs, with a batch size of 16 (12 for the no-gripper fine-tuning stage). Training converges in approximately 1.5 hours on a single NVIDIA RTX 4090 GPU.

\subsection{Visual Supervision through Differentiable Rendering}
\label{supp_sec:diff_rendering}
We use differentiable silhouette rendering to train \name
together with the hand--eye transform from image-space alignment. Hand--eye
calibration estimates the rigid camera-from-base transform
$\mathbf{T}_{cb}\in SE(3)$ that projects base-frame robot geometry consistently
into the image. Following EasyHeC, the visual objective compares rendered and
observed robot silhouettes and backpropagates through rendering, kinematics,
differentiable dynamics, and the actuator model. The IoU reflects the
joint image-space refinement; it does not, by itself, isolate the contribution of
the actuator update or establish an improvement in rollout accuracy.

\paragraph{Supervision signals}
For each selected frame $i \in \mathcal{I}$, the following quantities are available.

\textbf{Segmentation mask.}
A binary robot silhouette $M_i \in \{0,1\}^{H_{\mathrm{img}} \times W_{\mathrm{img}}}$ is obtained using SAM3~\cite{carion2025sam3segmentconcepts}. This mask provides pixel-level silhouette supervision in the image plane.

\textbf{Camera intrinsics.}
A camera intrinsic matrix $\mathbf{K}_i \in \mathbb{R}^{3 \times 3}$ is estimated using VGGT and analytically transformed back to the original image coordinate system. Camera intrinsics are treated as fixed during optimization.

\textbf{Camera pose initialization.}
An initial estimate of the camera-from-base transformation $\mathbf{T}_{cb}^{(0)} \in SE(3)$ is obtained via a geometric three-point initialization. This initialization aligns depth-lifted image keypoints with corresponding robot keypoints computed from forward kinematics, and serves as the starting point for differentiable refinement.

\paragraph{Pose parameterization}
We parameterize the camera extrinsic transformation in the Lie algebra. Let
\begin{equation}
\boldsymbol{\xi} \in \mathbb{R}^{6}, \qquad
\mathbf{T}_{cb}(\boldsymbol{\xi}) = \exp(\widehat{\boldsymbol{\xi}}) \in SE(3),
\end{equation}
where $\widehat{\boldsymbol{\xi}}\in\mathfrak{se}(3)$ is the matrix representation of the twist coordinates and $\exp(\cdot)$ is the matrix exponential.
The optimization variable $\boldsymbol{\xi}$ is initialized by applying the logarithm map to $\mathbf{T}_{cb}^{(0)}$.

\paragraph{Differentiable actuator rollout and silhouette rendering}
For a selected frame $i$ at control step $t_i$, \name predicts
the torque surrogate from the input sequence $\mathbf{X}_{t_i}$, and differentiable
simulation produces $\mathbf{s}^{\mathrm{sim}}_i(\theta)$. The offline visual
experiment uses the synchronized current sequence recorded with the trajectory,
so it is recorded-current-conditioned in the same sense as Sec.~\ref{sec:method}.
We use the simulated joint configuration and optionally add a small, regularized
per-frame correction $\Delta\mathbf{q}_i$:
\begin{equation}
\qquad
\mathbf{q}_i(\theta)=\mathbf{q}^{\mathrm{sim}}_i(\theta)
+\Delta\mathbf{q}_i.
\label{eq:visual_state}
\end{equation}
Forward kinematics then yields each link pose:
\begin{equation}
\mathbf{T}^{(l)}_{bl,i} = \mathrm{FK}_l(\mathbf{q}_i), \qquad l = 1,\dots,N_{\mathrm{link}} .
\end{equation}
Each link mesh is transformed into the camera frame by
$\mathbf{T}_{cb}(\boldsymbol{\xi})\mathbf{T}^{(l)}_{bl,i}$ and rendered with a
differentiable rasterizer, producing a soft silhouette
$\mathbf{s}^{(l)}_i\in[0,1]^{H_{\mathrm{img}}\times W_{\mathrm{img}}}$.

Following EasyHeC, individual link silhouettes are aggregated using saturation:
\begin{equation}
\hat{M}_i
=
\min\!\left(1,\; \sum_{l=1}^{N_{\mathrm{link}}} \mathbf{s}^{(l)}_i \right),
\label{eq:rendered_mask}
\end{equation}
yielding the rendered soft robot mask $\hat{M}_i \in [0,1]^{H_{\mathrm{img}} \times W_{\mathrm{img}}}$.

\begin{figure*}
  \centering
  \includegraphics[width=\textwidth]{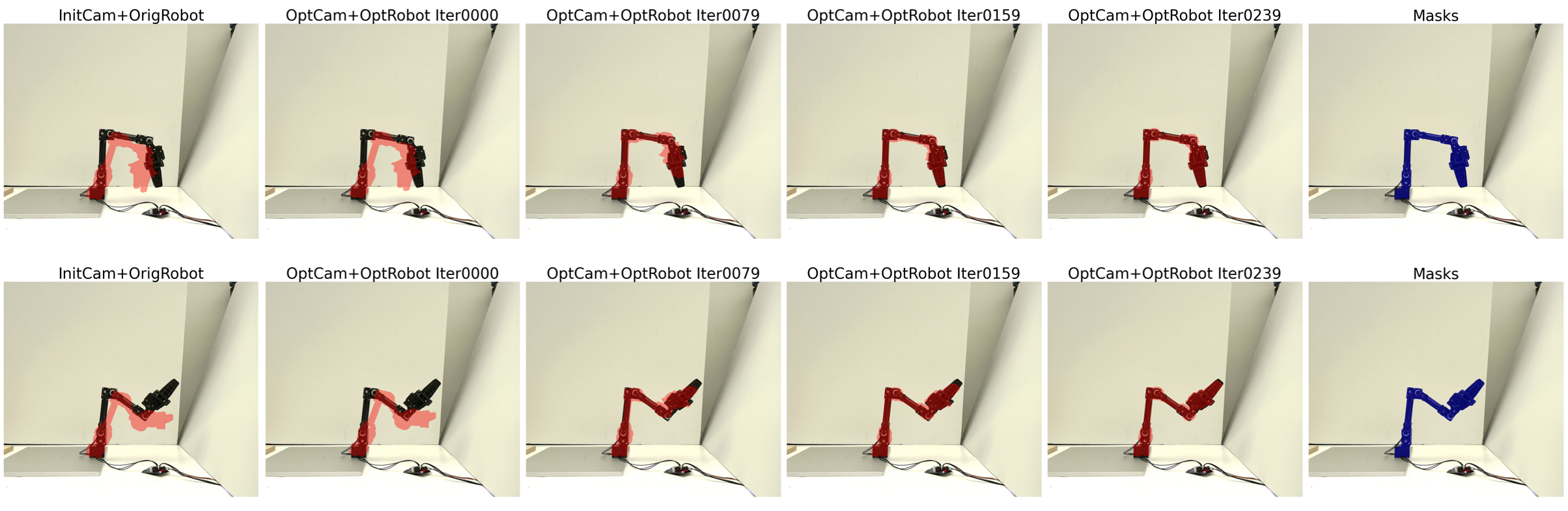}
  \caption{\textbf{Visual refinement through differentiable
  rendering.} Rows show two robot configurations. Columns show the initial
  alignment, representative optimization snapshots through iteration 239, and
  the observed segmentation mask. Red denotes the rendered robot silhouette;
  blue denotes the observed mask. The complete optimization contains 1,845
  gradient steps (Tab.~\ref{tab:runtime_metrics}).}
  \label{fig:appendix_diff_render}
\end{figure*}

\paragraph{Silhouette alignment objective}
We minimize the discrepancy between the rendered soft silhouette
$\hat{M}_i\in[0,1]^{H_{\mathrm{img}}\times W_{\mathrm{img}}}$ and the observed binary mask
$M_i\in\{0,1\}^{H_{\mathrm{img}}\times W_{\mathrm{img}}}$:

\begin{equation}
\begin{aligned}
\mathcal{L}_{\text{mask}}
&=
\frac{1}{|\mathcal{I}|}
\sum_{i \in \mathcal{I}}
\left\|
\hat{M}_i - M_i
\right\|_F^2 \\
&=
\frac{1}{|\mathcal{I}|}
\sum_{i \in \mathcal{I}}
\sum_{u,v}
\left(
\hat{M}_i(u,v) - M_i(u,v)
\right)^2 .
\end{aligned}
\label{eq:mask_loss}
\end{equation}

\paragraph{Joint visual optimization of camera and \name}
In the visual-training experiment, we jointly optimize the
\name parameters, camera extrinsics, and optional per-frame state
corrections. Small per-frame state
corrections absorb residual synchronization, sensing, and geometry errors, and
are quadratically regularized:
\begin{equation}
\mathcal{L}_{\text{reg}}^{q}
=
\lambda_q
\cdot
\frac{1}{|\mathcal{I}|}
\sum_{i \in \mathcal{I}}
\|\Delta \mathbf{q}_i\|_2^2 .
\end{equation}
The visual term is
\begin{equation}
\mathcal{L}_{\text{visual}}
=
\mathcal{L}_{\text{mask}}
+\mathcal{L}_{\text{reg}}^{q}.
\label{eq:total_pose}
\end{equation}
Combining the visual term with the trajectory losses in Sec.~\ref{sec:method}, we solve
\begin{equation}
\min_{\theta,\boldsymbol{\xi},\{\Delta\mathbf{q}_i\}}
\quad
\mathcal{L}_{\text{traj}}+\lambda_{\text{visual}}
\mathcal{L}_{\text{visual}}.
\label{eq:visual_training}
\end{equation}
We select the positive weights $\lambda_q$ and
$\lambda_{\text{visual}}$ on the training sequence to balance trajectory
fidelity, silhouette alignment, and the magnitude of the state corrections,
and then hold them fixed for all frames.
Consequently, gradients from $\mathcal{L}_{\text{mask}}$ propagate to $\theta$ through the
chain $\hat M_i\rightarrow\mathrm{FK}\rightarrow
\mathbf{q}^{\mathrm{sim}}_i\rightarrow\mathrm{DiffSim}\rightarrow g_\theta$.
Because the camera extrinsics and $\Delta\mathbf q_i$ are optimized jointly,
the silhouette IoU measures the combined refinement and is not an isolated
ablation of the actuator update.

\paragraph{Optimization details}
We optimize $\theta$, $\boldsymbol{\xi}$, and the optional state
corrections with first-order gradients and retain the lowest-loss solution.

\paragraph{Accuracy metrics}
We evaluate calibration quality with soft intersection-over-union
(IoU) between the rendered soft mask $\hat M_i$ and the observed binary mask $M_i$:
\begin{equation}
\begin{aligned}
\mathrm{IoU}_{\mathrm{soft}}
&=\frac{1}{|\mathcal{I}|}
\sum_{i\in\mathcal{I}}
\frac{\sum_{u,v}M_i(u,v)\hat M_i(u,v)}{D_i}, \\
D_i
&=\sum_{u,v}\left[M_i(u,v)+\hat M_i(u,v)
-M_i(u,v)\hat M_i(u,v)\right].
\end{aligned}
\end{equation}

\paragraph{Quantitative and qualitative results}
We report both silhouette alignment accuracy and runtime statistics for joint camera--actuator refinement on a single real-world video sequence. All measurements are conducted on a single NVIDIA A6000 GPU. The evaluation uses six images sampled from a single video; these frames are intentionally non-consecutive and correspond to substantially different robot configurations, providing diverse geometric constraints for calibration. In Tab.~\ref{tab:accuracy_metrics}, \emph{OptRobot} denotes the optimized robot-side variables: the \name parameters $\theta$ and, when enabled, the regularized state corrections $\{\Delta\mathbf q_i\}$. The \emph{Robot refinement} row keeps the camera fixed, whereas \emph{Joint refinement} also optimizes the camera extrinsics $\boldsymbol\xi$. Tab.~\ref{tab:accuracy_metrics} summarizes silhouette alignment accuracy, while Tab.~\ref{tab:runtime_metrics} reports the corresponding runtime statistics. Qualitative visualizations of the joint refinement process at different optimization stages are provided in Fig.~\ref{fig:appendix_diff_render}.

\begin{table}
\centering
\caption{Silhouette alignment accuracy (IoU) on real-world data.}
\label{tab:accuracy_metrics}
\begin{tabular}{lcc}
\toprule
Method & Mean IoU & Std IoU \\
\midrule
Initialization (InitCam + InitRobot) & 0.2589 & 0.0441 \\
Robot refinement (InitCam + OptRobot) & 0.2974 & 0.0754 \\
Joint refinement (OptCam + OptRobot) & 0.8515 & 0.0065 \\
\bottomrule
\end{tabular}
\end{table}

\begin{table}
\centering
\caption{Runtime statistics per sequence on real-world data.}
\label{tab:runtime_metrics}
\begin{tabular}{lcc}
\toprule
Method  & Iterations & Total time (s) \\
\midrule
Joint refinement & 1,845 & 180.23 \\
\bottomrule
\end{tabular}
\end{table}

\subsection{Network Architecture Comparison}
\label{supp_sec:arch_comp}

To isolate the effect of architecture while controlling for capacity, we compare models with a budget of
approximately 1.4 million parameters. Hidden widths and layer counts are adjusted
to keep the MLP, GRU, LSTM, and Transformer variants within 25\% of the
Transformer parameter count (Tab.~\ref{supp_tab:arch_config}). All models use
the same training data, target definitions, loss weights, augmentation, and
evaluation protocol. Unless noted below, they also use a batch size of 16, AdamW with a weight decay of $10^{-4}$, and a dropout rate of 0.1. Each input contains eight
historical samples and the current sample, all sampled at approximately 58.8\,Hz.
The MLP uses a lower learning rate ($3\times10^{-5}$) because the larger rate was unstable during differentiable
rollouts. \name uses four self-attention layers, four heads, hidden
dimension 192, feed-forward dimension 384, mean pooling, and query-dependent
gated attention. Its encoder feed-forward blocks use GELU, while the output
heads use SiLU, matching the implementation used in our experiments.
\begin{table}
  \centering
  \caption{\textbf{Neural architecture configurations for ablation study.} All models are trained with identical loss weights, data augmentation, and early stopping criteria. Parameter counts are matched within 25\% of the Transformer baseline.}
  \label{supp_tab:arch_config}
  \vspace{-1mm}
  \setlength{\tabcolsep}{6pt}
  \resizebox{0.85\columnwidth}{!}{
  \begin{tabular}{lcccc}
    \toprule
    \textbf{Model} & \textbf{Hidden} & \textbf{Latent} & \textbf{Params} & \textbf{LR} \\
    \midrule
    MLP & 416 & 208 & 1.14M & $3\times10^{-5}$ \\
    GRU & 325 & 162 & 1.43M & $1\times10^{-4}$ \\
    LSTM & 275 & 137 & 1.44M & $1\times10^{-4}$ \\
    \midrule
    \textbf{Ours} & 192 & 96 & 1.44M & $1\times10^{-4}$ \\
    \bottomrule
  \end{tabular}
  }
  \vspace{-2mm}
\end{table}

\subsection{\texorpdfstring{Taxonomy of Robotic Force-Sensing Methods}{Taxonomy of Robotic Force-Sensing Methods}}
\label{supp_sec:force_sensing_taxonomy}

Tab.~\ref{supp_tab:force_sensing_taxonomy} summarizes representative approaches for robotic force sensing and contact inference, grouped by sensing signals and whether they explicitly estimate force magnitude.
Classic sensorless collision monitoring infers external torques from residuals/momentum observers and mainly supports contact detection~\cite{de2005sensorless,de2006collision,haddadin2008collision}.
Recent intrinsic and whole-body touch methods exploit joint-torque and position sensing, whereas learning-based force-aware policies span low-cost bilateral-control arms and force-sensor-free whole-body robots~\cite{iskandar2024intrinsic,fu2025unitac,kobayashi2025ilbit,zhi2025learning}.
For manipulators, virtual force sensing, contact localization, and observer/filter-based methods can estimate interaction forces or localize contacts, generally using torque feedback, commanded torque, or calibrated dynamics~\cite{magrini2014estimation,manuelli2016localizing,hu2017contact,liu2021sensorless,han2021toward,linderoth2013robotic,osburg2022using}.
Visual cues can estimate contact-force location and magnitude, whereas inertial-aided proprioceptive methods improve collision monitoring~\cite{wang2022visual,birjandi2020observer}.
In contrast, \name enables force-magnitude estimation on low-cost servos by learning a torque surrogate from actuation telemetry and jointly predicting external forces without dedicated torque sensors.

\subsection{Imitation Learning}
\label{supp_sec:imitation_learning}

\begin{figure}
    \centering
    \includegraphics[width=\linewidth]{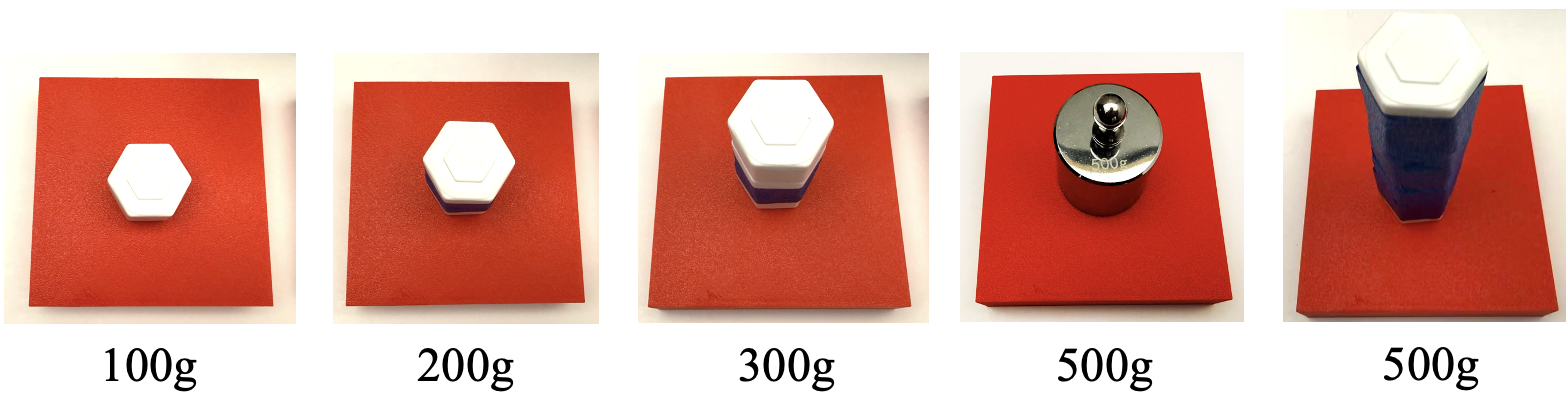}
    \vspace{-6mm}
    \caption{\textbf{Payloads used in the high-level control tasks.}}
    \label{fig:weights_for_highlevel}
    \vspace{-2mm}
\end{figure}

We compare two approaches for training the high-level controller via behavior cloning.
For the \textbf{position-only baseline}, the per-step observation is
$\mathbf{o}^{\mathrm{pos}}_t=[\mathbf{q}_{1:4,t},a_t]^\top\in\mathbb{R}^5$.
The policy receives the eight-step history
$\mathbf{O}^{\mathrm{pos}}_t=(\mathbf{o}^{\mathrm{pos}}_{t-7},\ldots,
\mathbf{o}^{\mathrm{pos}}_t)\in\mathbb{R}^{40}$ and minimizes
$\mathcal{L}_{\mathrm{BC}}=
\|\pi_\psi(\mathbf{O}^{\mathrm{pos}}_t)-\mathbf{g}^*_t\|_2^2$,
where $\mathbf{g}^*_t\in\mathbb{R}^5$ is the demonstrated command for the four
arm joints and gripper aperture.
\textbf{Our force-aware controller} augments the input with the estimated end-effector external force
$\hat{\mathbf{f}}_t$ predicted by a \emph{pretrained} \name, giving
$\mathbf{o}^{\mathrm{force}}_t=[\mathbf{q}_{1:4,t},a_t,
\hat{\mathbf{f}}_t]^\top\in\mathbb{R}^8$ and the corresponding eight-step
history $\mathbf{O}^{\mathrm{force}}_t\in\mathbb{R}^{64}$, which replaces
$\mathbf{O}^{\mathrm{pos}}_t$ in the same BC objective.
\name is trained once on the full payload set $\{100,200,300,400,500\}$\,g and is kept \emph{frozen} during policy learning.
The BC-policy demonstrations also include the 500\,g condition used in the pick-and-place evaluation.
The downstream protocol therefore evaluates task-level transfer within the payload range represented during pretraining.
For offline force analysis, we compute a trajectory-level statistic on
each demonstration-aligned rollout:
\begin{equation}
\qquad
\mathcal{R}_{f}
=\frac{1}{T_r}\sum_{k=0}^{T_r-1}
\|\hat{\mathbf{f}}_{t+k}\|_2^2.
\end{equation}

\begin{figure}
    \centering
    \includegraphics[width=0.85\linewidth]{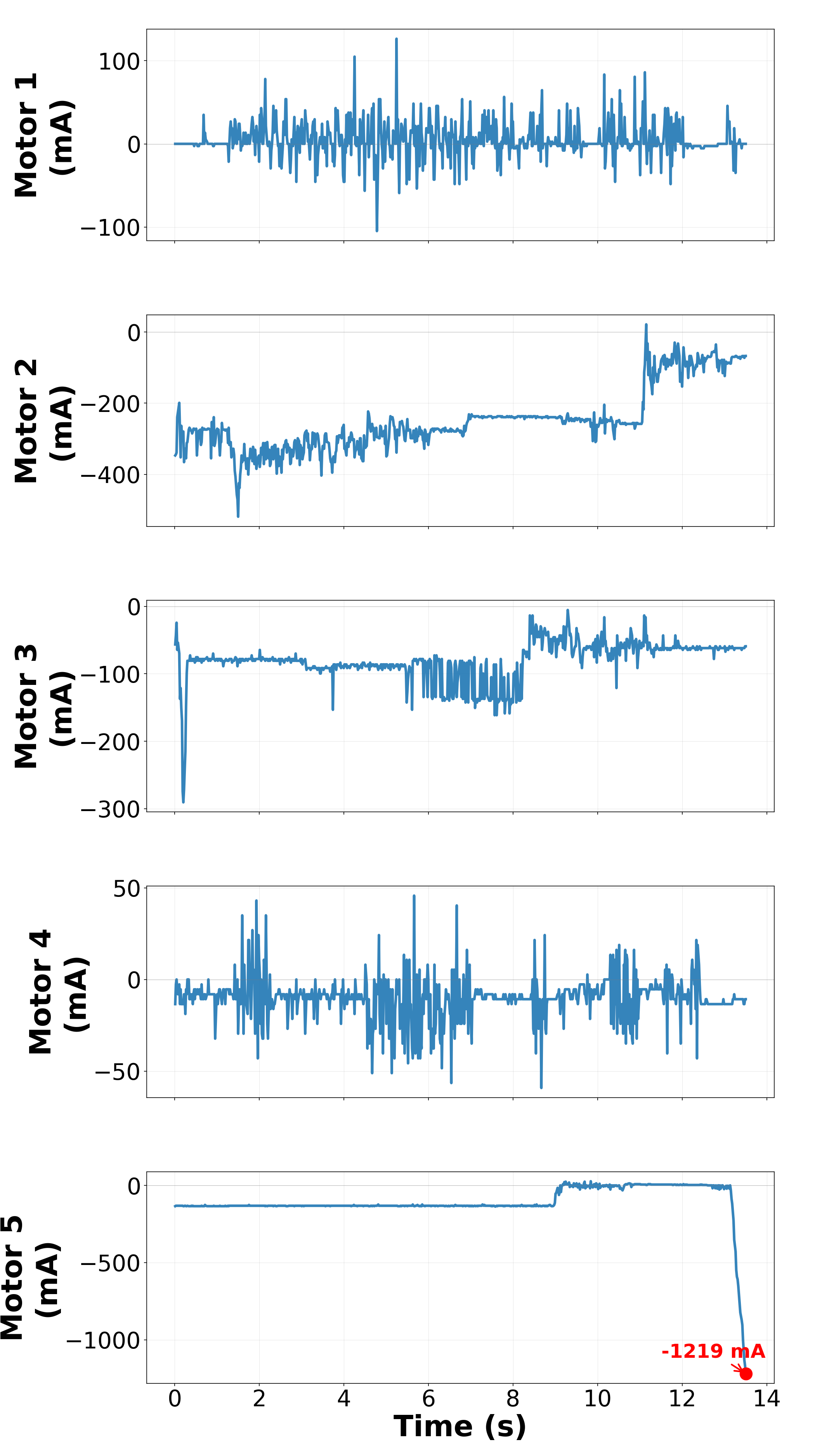}
    \vspace{-1mm}
    \caption{\textbf{Current overload in position-only behavior cloning.}
    Representative motor-current traces during high-level execution.
    The position-only baseline exhibits over-current spikes beyond the 1200\,mA safety limit, which triggers the onboard protection and terminates the rollout.}
    \label{fig:current_overload}
\end{figure}

\begin{table*}
\vspace{-5mm}
\centering
\small
\setlength{\tabcolsep}{4pt}
\renewcommand{\arraystretch}{1.12}

\begin{tabularx}{\textwidth}{@{}
p{6.1cm}
>{\raggedright\arraybackslash}p{2.5cm}
>{\centering\arraybackslash}p{2.0cm}
>{\centering\arraybackslash}p{2.0cm}
p{4.0cm}
@{}}
\toprule
\textbf{Direction} &
\textbf{Works} &
\textbf{Signals} &
\textbf{Force mag.} &
\textbf{Typical regime} \\
\midrule

Residual/momentum collision monitoring
& \cite{de2005sensorless,de2006collision,haddadin2008collision}
& $q,\dot q,\tau_c$ / TS
& Partial
& \makecell[l]{Simulation /\\ torque-sensed robots} \\
\midrule

Intrinsic/whole-body touch from proprioception
& \cite{iskandar2024intrinsic,fu2025unitac}
& TS $+\,q$
& Partial
& High-end (torque-capable) \\
\midrule

Learning-based force-aware policies
& \cite{kobayashi2025ilbit,zhi2025learning}
& ILBiT: $q,\dot q,\hat\tau$; Zhi: state history
& Partial / Yes
& Low-cost arm / legged robots \\
\midrule

Arm contact localization / virtual force sensing
& \cite{magrini2014estimation,manuelli2016localizing}
& Magrini: RGB-D $+\,q,\dot q,\tau$; Manuelli: $q,\dot q,\tau$
& Yes / Partial
& \makecell[l]{Torque-capable /\\ calibrated dynamics} \\
\midrule

Observer/filter for sensorless force estimation
& \cite{hu2017contact,liu2021sensorless,han2021toward,linderoth2013robotic,osburg2022using}
& $q,\dot q$ $+$ TS / I$\ \!\rightarrow\!\tau$
& Yes
& Industrial / Mixed \\
\midrule

Vision- and inertial-aided force/contact inference
& \makecell[l]{Vision:~\cite{wang2022visual}\\ IMU:~\cite{birjandi2020observer}}
& Wang: Vision $+\,q,\dot q$; Birjandi: IMU $+\,q$
& Yes / Partial
& \makecell[l]{Mixed\\(observability-dependent)} \\
\midrule

\addlinespace[2pt]
\name (ours)
& This paper
& Telemetry $+\,q,\dot q$
& Yes
& Low-cost (servo-driven) \\
\bottomrule
\end{tabularx}

\vspace{3mm}
\caption{\textbf{Taxonomy of robotic force sensing and contact inference.}
\textbf{Signals:} TS = torque sensing; I$ \ \!\rightarrow\!\tau$ = torque inferred from current/effort; $\tilde\tau$ = torque surrogate; $q,\dot q$ = kinematics; $\tau_c$ = commanded torque; $\hat\tau$ = estimated torque; Vision = exteroception; IMU = inertial sensing.
\textbf{Force mag.:} \emph{Yes} = explicit force/wrench magnitude; \emph{Partial} = mainly detection/localization or residual signals.
\name predicts $\tilde\tau$ from actuation telemetry and estimates external forces on low-cost servos without dedicated torque sensors.}
\label{supp_tab:force_sensing_taxonomy}
\vspace{-10mm}
\end{table*}

\begin{figure*}
  \vspace{-5mm}
  \centering
  \includegraphics[width=\textwidth]{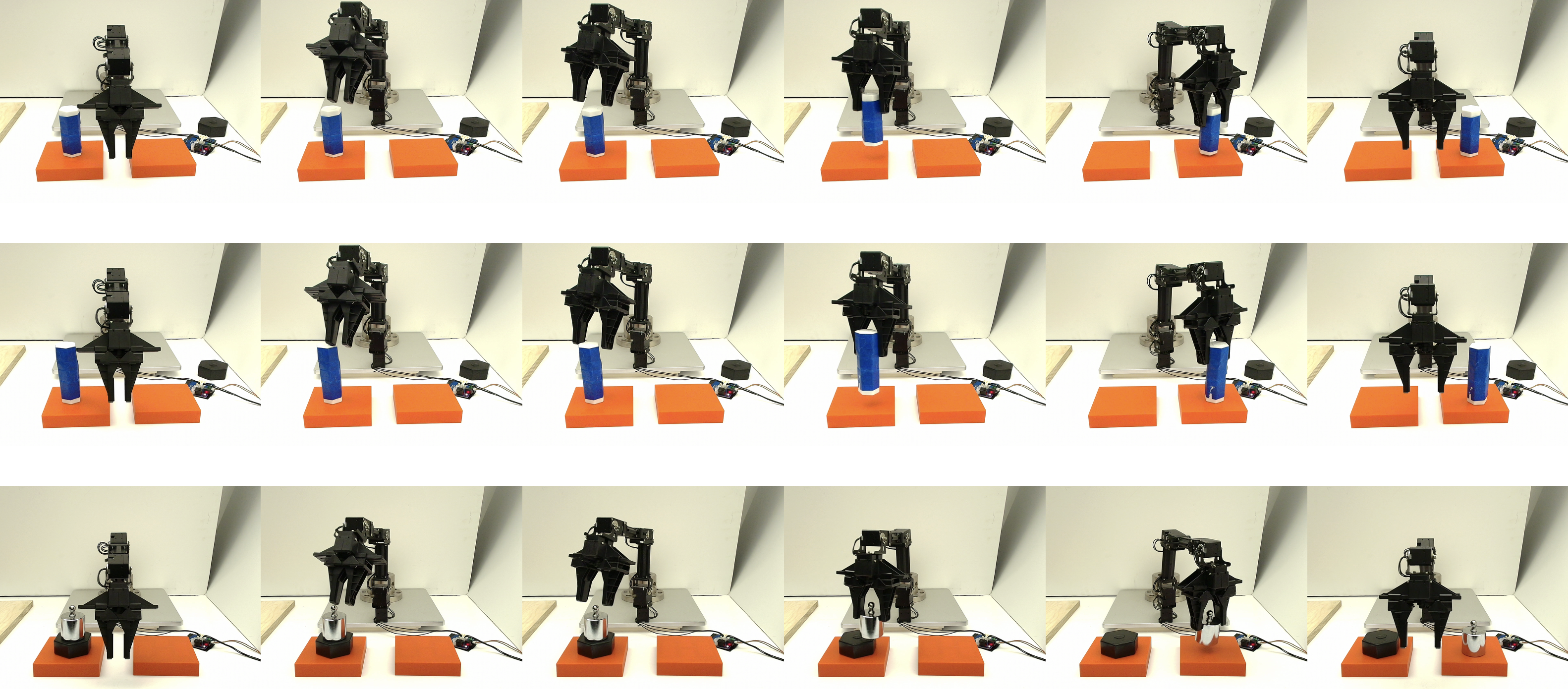}
\caption{\textbf{Qualitative results on high-level pick-and-place.}
Time-lapse snapshots (left to right) of the robot grasping, lifting, transporting, and placing objects under different payloads.
From top to bottom, we evaluate 400\,g, 500\,g, and a differently shaped 500\,g weight.}
  \label{fig:high_level_pick_place}
  \vspace{-5mm}
\end{figure*}

Here, $T_r$ is the rollout length. These offline rollouts replay the synchronized current measured in the
corresponding demonstration at every step. Because \name does not
predict the current that would result from a counterfactual policy action,
$\mathcal{R}_f$ is a diagnostic rather than an additional policy-training loss;
the policy is optimized with $\mathcal{L}_{\mathrm{BC}}$. During hardware
execution, $\hat{\mathbf f}_t$ is recomputed causally from telemetry available
through time $t$, including the live current $\mathbf i_t$.
Both methods use identical Transformer policies (4 layers, hidden dimension
128, and 4 heads) and are trained for 5,000 epochs on the same demonstrations.
For policy training, we use payloads $\{100,200,300,500\}$\,g with gripper
contact points at approximately matched heights, and evaluate on 400\,g and
500\,g loads. The position-only baseline can produce current spikes above the
1200\,mA safety threshold, triggering onboard protection and terminating the
execution early (Fig.~\ref{fig:current_overload}).

\end{document}